\documentclass[final,5pt,onecolumn]{elsarticle}

\usepackage{lineno} 
\modulolinenumbers[5]

\usepackage{hyperref}
\usepackage{microtype}
\usepackage{graphicx}
\usepackage{subfigure}

\graphicspath{{./figs/},{./figs2/}}

\usepackage{times,amsmath,epsfig}
\usepackage{multirow}
\usepackage{epstopdf}
\usepackage{amsmath}
\usepackage{amsfonts}
\usepackage{color}
\usepackage{fancyhdr}
\usepackage{algorithm}
\usepackage{algorithmic}
\usepackage{diagbox}

\def\ds3c{\textbf{D$\text{S}^3$C}}
\def\s3c{$\text{S}^3$C}

\def\x{\textbf{x}}
\def\y{\textbf{y}}
\def\z{\textbf{z}}
\def\0{\textbf{0}}
\def\1{\textbf{1}}

\def\and{\textrm{and}}

\newcommand{\RR}{I\!\!R} 

\newcommand{\myparagraph}[1]{\smallskip\noindent\textbf{#1.}}

\newcommand{\mcb}{\color{black}}

\def\x{\boldsymbol{x}}
\def\y{\boldsymbol{y}}
\def\z{\boldsymbol{z}}

\def\I{\mathcal{I}}
\def\J{\mathcal{J}}

\def\N{\mathcal{N}}
\def\R{\mathcal{R}}
\def\X{\mathcal{X}}
\def\Y{\mathcal{Y}}

\begin{document}

\begin{frontmatter}
%
\title{ Density-Adaptive Kernel based Efficient Reranking Approaches for Person Reidentification}

\author{Ruopei Guo}
\author{Chun-Guang Li} 
\author{Yonghua Li}
\author{Jiaru Lin}
\author{Jun Guo}
\address{School of Information and Communication Engineering, Beijing University of Posts and Telecommunications, Beijing 100876, P.R. China.}

\begin{abstract}
Person reidentification (ReID) refers to the task of verifying the identity of a pedestrian observed from nonoverlapping views in a surveillance camera network. It has recently been validated that reranking can achieve remarkable performance improvements in person ReID systems. However, current reranking approaches either require feedback from users or suffer from burdensome computational costs. In this paper, we propose to exploit a density-adaptive smooth kernel technique to achieve efficient and effective reranking. Specifically, we adopt a smooth kernel function to formulate the neighbor relationships among data samples with a density-adaptive parameter. Based on this new formulation, we present two simple yet effective reranking methods, termed \emph{inverse} density-adaptive kernel based reranking (inv-DAKR) and \emph{bidirectional} density-adaptive kernel based reranking (bi-DAKR), in which the local density information in the vicinity of each gallery sample is elegantly exploited. Moreover, we extend the proposed inv-DAKR and bi-DAKR methods to incorporate the available extra probe samples and demonstrate that when and why these extra probe samples are able to improve the local neighborhood and thus further refine the ranking results. Extensive experiments are conducted on six benchmark datasets, including: PRID450s, VIPeR, CUHK03, GRID, Market-1501 and Mars. The experimental results demonstrate that our proposals are effective and efficient.

\end{abstract}

%

\end{frontmatter}


\section{Introduction}
\label{sec:intro}

Person reidentification (ReID) refers to the task of verifying the identity of a pedestrian observed from nonoverlapping views of surveillance camera networks~\cite{Gong:Springer2014}. Due to its importance to public security, it has received extensive attention and has increasingly become one of the most critical tasks in video analysis. However, the task of ReID is quite challenging because the views captured by the surveillance cameras are recorded under unconstrained conditions, and thus, the obtained images may contain large variations due to changes in pose, viewpoint, and illumination, as well as occlusion, blurring, background variations, etc.

To address these challenges, the standard pipeline of a person ReID system usually consists of two components: a) robust and discriminative feature extraction and b) supervised metric learning. 
In previous works, the majority of efforts have focused on extracting robust and discriminative visual representations. It has been verified that local features, \textit{e.g.}, color and histograms of oriented gradients \cite{Dalal:CVPR2005,Dalal:ECCV2006,Liu:ECCV2012,Yang:ECCV2014,Si:TSMC17}, are effective for person ReID, and combining multiple types of features, \textit{e.g.}, color, texture, and spatial structure, is useful for finding more informative matches \cite{Dong:BMVC2014, Gheissari:CVPR2006, Hu:CVPR2013, Gray:ECCV2008, Wang:ICCV2007, Matsukawa:CVPR2016, Liao:CVPR2015, DAI:PR18, Franco:PR17}.
On the other hand, supervised metric learning methods, which learn a discriminative distance metric (or equivalently a low-dimensional subspace), 
in which the samples of the same person are closer, can facilitate the task of finding informative matches \cite{ Davis:ICML2007, Dikmen:ACCV2010, Kostinger:CVPR2012, Li:CVPR2013, WEINBERGER:NIPS2006, Zheng:CVPR2011,Si:ICIP15}. In addition, simultaneous feature extraction and metric learning has also been investigated under the framework of deep convolutional neural networks~\cite{Si:CVPR18,Wu:PR17, Wang:PR18, Zhicheng:PR18,Guo:ACCESS2020,Lin:ICPR2020,Hao:AAAI2019,Hao:MM2019,Tang:YWSG20}.
{\mcb For example, in \cite{Hao:AAAI2019}, an end-to-end dualstream hypersphere manifold embedding network is proposed to capture the correlation between classification and identification constraints, in which not only a hypersphere is learned to describe the intra-modality variations and cross-modality variations, but also a two-stage training scheme is designed to produce decorrelated features. And in \cite{Hao:MM2019}, a dual-alignment feature embedding model is designed for cross-modality person re-identification, in which the camera-invariant information is obtained by part-level spatial alignment, and the embedding features across visible and infrared modalities are aligned. Moreover, in \cite{Tang:YWSG20}, a novel model is proposed to fix the error back propagation problem of feature pyramids by feature pyramid optimization, and to reduce the background clutters of images by gradual background suppression.
}

Compared with methods based on feature representation and metric learning, which use content information or supervision information, remarkable performance improvements in person ReID have recently been achieved by using reranking methods, which consider \emph{context information} in ranking lists, \textit{e.g.}, \cite{Liu:ICCV2013, Wang:ECCV2016, Bai2017Scalable, Bai:TIP2016, Zhong:CVPR2017}.
Roughly, these reranking methods are based on exploiting the context information from the feedback of users (\textit{e.g.}, \cite{Liu:ICCV2013, Wang:ECCV2016}), from the manifold structure of the gallery samples (\textit{e.g.}, \cite{Bai2017Scalable}), or from the neighborhood structure of local neighbors (\textit{e.g.}, \cite{Bai:TIP2016, Zhong:CVPR2017}). However, gathering feedback information is a substantial burden on users, and approximating the manifold structure of the gallery samples by building an affinity graph is computationally expensive. While the neighborhood structure based methods lead to promising performance improvements, existing methods still suffer from either sensitivity to the tradeoff parameters or a heavy computational burden.


In this paper, we exploit a \textit{density-adaptive kernel} technique 
to perform efficient and effective reranking for person ReID. Specifically, we adopt a density-adaptive parameter to capture the local density information and then use it to formulate smooth kernel functions for finding the $k$-\emph{nearest neighbors} ($k$-NN), the $k$-\emph{inverse nearest neighbors} ($k$-INN), and the $k$-\emph{reciprocal nearest neighbors} ($k$-RNN) of a probe sample.
Such a density-adaptive smooth kernel function quantifies the neighborhood of a sample in the form of a continuous (nonnegative) value measured on \emph{an individual sample-specific scale}.
%
In this way, the inverse ranking list is quantified, and thus, merging the inverse ranking list with the direct ranking list to form the final reranking list becomes extremely easy. Concretely,
%
%
we present two simple yet effective reranking methods, termed \emph{inverse} density-adaptive kernel based reranking (inv-DAKR) and \emph{bidirectional} density-adaptive kernel based reranking (bi-DAKR). 
Depending on how the probe samples are used, we divide the proposed reranking approaches into the following two groups:
%
\begin{itemize}
\item inv-DAKR and bi-DAKR, which are used in the setting in which only a single probe sample and a set of gallery samples are available. The gallery samples are used to provide local density information and thus refine the ranking results to improve the accuracy of a person ReID system. This setting is studied in our preliminary work \cite{Guo:ICPR2018}.

\item inv-DAKR+ and bi-DAKR+, which are used in scenarios in which a set of probe samples and a set of gallery samples are both available. The extra probe samples can provide correct neighborhood information for the gallery samples, and thus provide more accurate local density information, leading to remarkable improvements in the final reranking results.

\end{itemize}
%

This paper is a substantial extension of our preliminary work \cite{Guo:ICPR2018}. Compared to our previous work, the extensions include the following three aspects:
\begin{itemize}
\item  We present a set of precise interpretations of the essential connections from the proposed inv-DAKR and bi-DAKR methods to $k$-INN and $k$-RNN methods, and we describe the fundamental mechanism of how the ambiguity is captured and preserved in inv-DAKR and bi-DAKR.

\item We extend the proposed inv-DAKR and bi-DAKR methods to inv-DAKR+ and bi-DAKR+, respectively, in which the available extra probe samples are used in an unsupervised manner to improve the reranking results.

\item We extend the previous $k$-INN and $k$-RNN reranking approaches to $k$-INN+ and $k$-RNN+, respectively, in which the extra probe samples are used to improve the local neighborhood in an unsupervised manner, thus leading to improvements in the reranking results. 

\item We conduct more extensive experimental evaluations on six benchmark datasets and achieve promising experimental results, 
     supplemented with thorough analysis and discussions, which demonstrate when and why the extra probe samples bring performance improvements 
     and how to set the 
     hyperparameter $k$.




\end{itemize}

\myparagraph{Paper Outline} The remainder of this paper is organized as follows. Section~\ref{sec:related-work} reviews the relevant work, and Section~\ref{sec:kNN-to-kRNN} describes the concepts of $k$-INN and $k$-RNN. Section~\ref{sec:our-proposal} presents our proposals. Section~\ref{sec:experiments} reports the experiments and presents corresponding discussions, and Section~\ref{sec:conclusion} concludes the paper.

\section{Related Work}
\label{sec:related-work}

This section will review the relevant works that have addressed the reranking task. 
Note that reranking in person ReID can be viewed as a postprocessing stage that exploits the context information among the ranking results. The context information used to perform reranking roughly comes from three sources: a) user feedback, b) manifold structure of gallery samples, and c) neighborhood structure. Accordingly, we divide the existing reranking methods for person ReID into three categories.
%

\subsection{User Feedback based Methods}

In person ReID, the content information alone is sometimes not reliable to produce accurate ranking results due to dramatic changes in pose, viewpoint, illumination, background, etc. Therefore, the context information in the ranking list is considered to refine the ranking list.

The context information can be explored from users' feedback. For example, in \cite{Liu:ICCV2013, Wang:ECCV2016}, the feedback knowledge from users is used to 
refine the ranking results.
Specifically, in \cite{Liu:ICCV2013}, negative samples are required from users, synthesized probes from pairs of cameras are generated, and Laplacian SVM is used to iteratively refine a post-ranking function. Unfortunately, these operations do not scale to multiple-camera configurations. In contrast to \cite{Liu:ICCV2013}, the reranking list is incrementally optimized for each new probe from feedback in \cite{Wang:ECCV2016}. 
However, feedback-based refinement approaches rely on continuous user feedback, which places a heavy burden on users. Without feedback from users, these methods cannot work.

\subsection{Manifold Structure based Methods}


The manifold structure of the gallery samples also represents useful contextual information for refining the ranking list.
In \cite{Chen:ICIP13}, each probe sample is embedded into the manifold structure, which is formed by the gallery samples. The reranking is performed by label propagation, which needs to build an affinity graph and compute the matrix inverse. Thus, it is time consuming and sensitive to the hyperparameter.
Similarly, an approach called supervised smoothed manifold (SSM) is proposed in \cite{Bai2017Scalable}, in which an affinity graph 
is built as context information to capture the manifold structure in the gallery set, and the pairwise supervision information in the training set is propagated across the affinity graph.
However, building the affinity graph of the probe and gallery samples is also computationally expensive because all the samples in the gallery and training set are involved.

\subsection{Neighborhood Structure based Methods} 

Context information for reranking can be explored from the neighborhood of the probe sample. This approach is mainly based on an interesting observation that reliable matches between a probe and the gallery samples are usually \emph{mutually sharing neighbor relationships}. 

Note that the nearest neighbor relationship is not symmetric, that is, the $k$-nearest neighbors ($k$-NN) of a query sample might not include the query sample as one of their $k$-NN samples \cite{Korn:ASR2000}. A subset of samples that include the query sample as one of their $k$-NN samples are called the $k$-inverse nearest neighbors ($k$-INN) of the query sample. The $k$-INN provide important context information for reranking. While the asymmetry of the $k$-INN provides useful information, finding a complete set of $k$-INN requires one to check whether the probe sample is among the $k$-NN list for each sample in the gallery set, which is computationally expensive. 
In \cite{Qin:CVPR2011}, the concept of $k$-reciprocal nearest neighbors ($k$-RNN) is proposed for image retrieval, in which the $k$-RNN of a probe sample are defined as the intersection of its $k$-NN and $k$-INN sets. 
Due to the heavy computational cost, it can only be applied in a distributed computation framework. To increase the computation speed, sparse contextual activation (SCA) is proposed in \cite{Bai:TIP2016} for visual retrieval task; in SCA, the local distribution of the data samples in the $k$-RNN set is encoded to form a representation vector such that the Jaccard distance between two data samples in the $k$-RNN set can be simply obtained by comparing two representation vectors. Furthermore, in \cite{Zhong:CVPR2017}, an approach based on $k$-RNN combined with query expansion and SCA is proposed. 
In addition, in \cite{Yu:BMVC2017}, the original high-dimensional feature vector is divided into several subfeature vectors, and the final reranking score \cite{Bai:TIP2016, Zhong:CVPR2017} is computed by fusing all contextual information from the subfeatures. In \cite{Leng:ICME2013, Leng:MTA2015}, the index information of the $k$-NN set and $k$-INN set is used as content information, and the index information of the $k$-RNN set is used as context information. The final similarity score for reranking is computed by combining the content information and the context information. However, the shortcomings of the $k$-INN and $k$-RNN approaches, \textit{i.e.}, a heavy computational burden and sensitivity to the hyperparameter $k$, still remain.
Finally, it should be noted that a different framework using the $k$-RNN approach is proposed in \cite{Garcia:2017TIP}, in which discriminative features for reranking are learned with the help of the ranking list.

\begin{figure}[phtb]
\centering
\subfigure[$k$-NN]{\includegraphics[clip=true,trim=0 0 0 0,width=0.45\columnwidth]{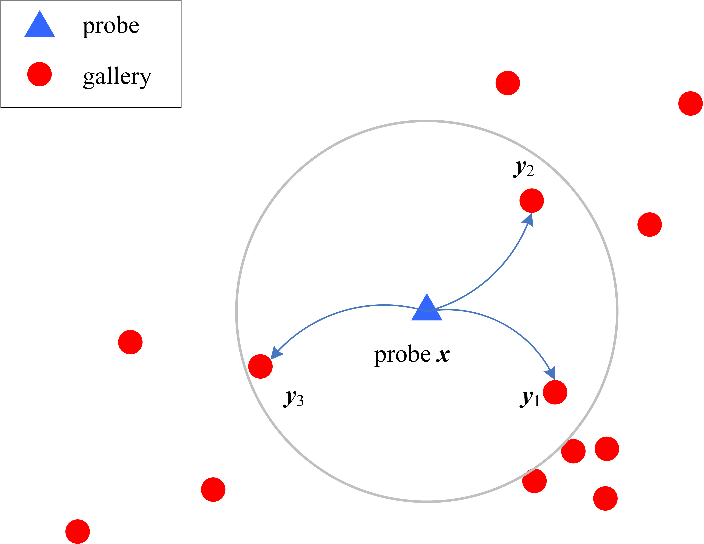} 
\label{fig:illustration-Re-ID-1}}
\hspace{18pt}
\\
\subfigure[$k$-INN]{\includegraphics[clip=true,trim=0 0 0 0,width=0.45\columnwidth]{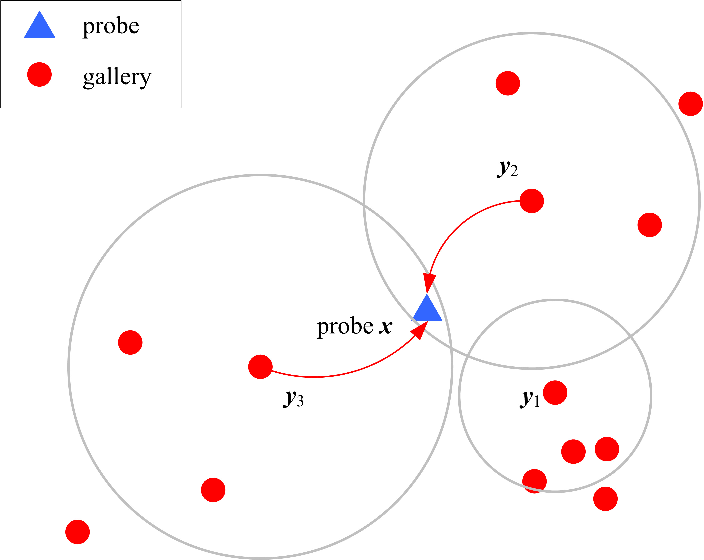} 
\label{fig:illustration-Re-ID-2}}
\hspace{18pt}
\subfigure[$k$-RNN]{\includegraphics[clip=true,trim=0 0 0 0,width=0.45\columnwidth]{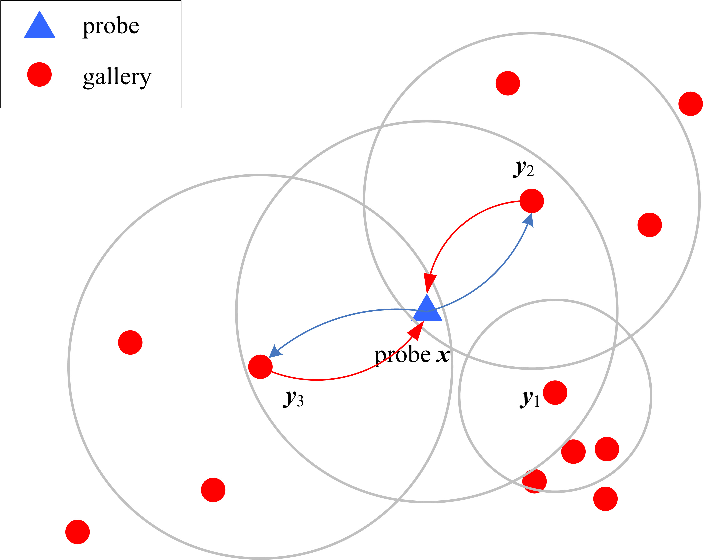} 
\label{fig:illustration-Re-ID-3}}
\caption{ Illustration of ranking and reranking in person ReID. (a) $k$-NN; (b) $k$-INN for inverse reidentification; (c) $k$-RNN, \emph{i.e.}, intersection of $k$-NN with $k$-INN, for bidirectional reidentification. The large circles indicate the boundaries of the local neighborhoods specified by the $k$-NN with $k=3$. In panel (a), the gallery samples $\y_1$, $\y_2$ and $\y_3$ are the three nearest neighbors of the probe $\x$. If we consider the $k$-NN in an inverse way, then only the gallery samples $\y_2$ and $\y_3$ are retained, as shown in panel (b). When combining the $k$-NN and $k$-INN results, we obtain the reranking results based on the $k$-RNN, as shown in panel (c). In this case, the $k$-INN and $k$-RNN results are the same.}
\label{fig:illustration-Re-ID}
\end{figure}

\section{From $k$-NN and $k$-INN to $k$-RNN: Exploring Secrets in Neighborhoods} 
\label{sec:kNN-to-kRNN}

Denote the probe set as $\X=\{\x_1,\cdots,\x_M\}$ and the gallery set as $\Y=\{\y_1,\cdots,\y_N\}$, where $\x_i$ and $\y_j \in \RR^d$. Since that our methods can be viewed as \textit{smoothed} versions of the $k$-INN and $k$-RNN based reranking methods, we formally describe the concepts of $k$-NN, $k$-INN, and $k$-RNN before presenting our proposals.

\subsection{$k$-Nearest Neighbors ($k$-NN) based Ranking} 
\label{sec:direct-unilateral-identification}

In a person ReID system, one calculates the distance between the probe $\x_i \in \X$ and each gallery sample $\y_j \in \Y$ and then determines a set of best matching candidates from the gallery set $\Y$. Precisely, finding the top-$k$ best matching candidates can be formulated as follows:
\begin{align}
\N(\x_i,k) =\arg\min ^{[1:k]}\limits_{\y_j \in \Y}~\| \y_j - \x_i \|,
\label{eq:re-id}
\end{align}
where $\|\cdot\|$ is a distance metric predefined or learned from the data, $\N(\x_i,k)$ is the set of $k$-NN of the probe $\x_i$, and $[1:k]$ indicates taking the first $k$ candidates from the sorted list.
%
%
%
The true matches are expected to be included in $\N(\x_i,k)$. In Fig.\ref{fig:illustration-Re-ID-1}, the blue links starting from the probe $\x_i$ and terminating at the gallery samples $\{\y_1, \y_2, \y_3\}$ illustrate the $k$-NN of $\x_i$ with $k=3$. 
Finding reliable matches for person ReID based on $k$-NN is simple and efficient to implement, however, 
the potentially useful information in the local neighborhoods of the gallery samples has not been exploited.

\subsection{$k$-Inverse Nearest Neighbors ($k$-INN)  based Reranking} 
\label{sec:inverse-unilateral-identification}


To exploit the useful information in local neighborhoods of gallery samples, an intuitive approach is to apply $k$-NN approach \emph{inversely} on the gallery samples to search for the probe sample. This is the so-called $k$-\emph{inverse nearest neighbors} ($k$-INN) method~\cite{Korn:ASR2000}.

Specifically, for each gallery sample $\y_j \in \Y$, we find the best matches for $\y_j$ from $\{\x_i\} \cup \Y_{-j}$ where $\Y_{-j}$ is the gallery set without the $j$-th sample $\y_j$. 
Denote the $k$-NN of $\y_j$ as $\N(\y_j, k)$, which is defined as follows:
\begin{align}
\N(\y_j, k) =\arg\min^{[1:k]}\limits_{\y \in \{\x_i\} \cup \Y_{-j}}~\|\y - \y_j\|.
\label{eq:re-id-inverse}
\end{align}
If the probe $\x_i$ is included in $\N(\y_j, k)$, then $\y_j$ is viewed as one of the $k$-INN samples of $\x_i$.

To determine all the $k$-INN of $\x_i$, one has to compute $\N(\y_j, k)$ for all $j=1,\cdots, N$, where $N$ is the size of the gallery set. Let $\I(\x_i, k)$ denote the set of all $k$-INN samples of $\x_i$. 
If $\y_j \in \I(\x_i,k)$, the probe $\x_i$ is accepted as a good match of $\y_j$; otherwise, $\x_i$ is rejected by $\y_j$. The true match is expected to be included in $\I(\x_i,k)$.
%
%
%
%
%
%
In Fig.~\ref{fig:illustration-Re-ID-2}, the red links starting from the gallery samples $\{\y_2, \y_3\}$ and terminating at the probe $\x_i$ illustrate the reranking based on the $k$-INN. 
In this case, although none of the samples $\y_2$ and $\y_3$ identify $\x_i$ as the best match if only the (inverse) nearest neighbor (\textit{i.e.}, $k=1$) is considered, both $\y_2$ and $\y_3$ include the probe $\x_i$ in their $k$-NN if $k = 3$, and thus $\y_2$ and $\y_3$ are the $k$-INN of $\x_i$.

When $N$ is large, finding $\I(\x_i, k)$ is quite time consuming. 
Thus, in previous work, $k$-INN is performed only with the $k$-NN samples of the probe sample, rather than with the whole gallery set due to the heavy computation burden.
Unfortunately, finding $k$-INN with only a small subset (\textit{i.e.}, the $k$-NN of the probe sample) of the gallery samples is incomplete. Moreover, while the selected samples in $\I(\x_i,k)$ can further be sorted according to their distances to $\x_i$, the ambiguity of the potential matching candidates and the local density information have been ignored. Note that in previous work~\cite{Qin:CVPR2011, Leng:MTA2015, Bai:TIP2016, Leng:ICME2013, Zhong:CVPR2017}, $k$-INN has been considered as an intermediate step rather than as a reranking method, and thus, there is no valid evaluation of the performance of using $k$-INN as reranking for person ReID.

\subsection{$k$-Reciprocal Nearest Neighbors ($k$-RNN)  based Reranking}
\label{sec:bilateral-identification}

Conceptually, ``reidentification'' (\textit{e.g.}, $a \leftrightarrow b$) consists of two single-directional implications (\textit{i.e.}, $a \rightarrow b$ and $a \leftarrow b$). In person ReID, the $k$-NN-based and $k$-INN-based methods for finding the best matches can be analogously viewed as two single-directional implications. Thus, it is natural to integrate $k$-NN and $k$-INN to perform reidentification with \textit{bidirectional implications}. This concept leads to 
the so-called $k$-reciprocal nearest neighbors ($k$-RNN) approach, as illustrated in Fig.~\ref{fig:illustration-Re-ID-3}.
%
%
%
Denote the $k$-RNN of $\x_i$ as $ \R(\x_i,k)$. Then, $ \R(\x_i,k)$ is the intersection of $\N(\x_i,k)$ and $\I(\x_i,k)$, \textit{i.e.},
\begin{align}
\R(\x_i,k)=\N(\x_i,k) \cap \I(\x_i,k).
\end{align}
A gallery sample $\y_j$ does not belong to $\R(\x_i, k)$ as long as $\y_j \not \in \N(\x_i,k)$ or $\y_j \not \in \I(\x_i,k)$. Note that judging a sample $\y_j$ to belong or not to belong to $\N(\x_i,k)$ or $\I(\x_i,k)$ is a binary decision with a hard boundary. The ambiguity of the gallery sample $\y_j$ is ignored, especially for those gallery samples lying near the decision boundary.
%
%
The $k$-RNN approach is illustrated in Fig.~\ref{fig:illustration-Re-ID-3}. 
In this case, by integrating $k$-NN and $k$-INN, 
both $\y_2$ and $\y_3$ become reliable candidate matches to the probe sample $\x_i$ when using $k=3$.

%
%

In $k$-RNN, the dissimilarity between the probe sample $\x_i$ and a gallery sample $\y_j \in \R(\x_i,k)$ is measured in different ways. In \cite{Qin:CVPR2011}, the dissimilarity is measured by a minimum cutoff. In \cite{Leng:ICME2013, Leng:MTA2015}, the index information included in $k$-NN and $k$-INN is used to define a similarity. In addition, in \cite{Garcia:2017TIP}, the $k$-RNN set is used to learn discriminant features for computing the dissimilarity. 
In~\cite{Bai:TIP2016,Zhong:CVPR2017,Yu:BMVC2017}, the dissimilarity is computed based on Jaccard distance, 
which is defined as follows:
\begin{align}
\J(\y_j,\x_i) =1-\frac{|\N(\y_j,k) \cap \N(\x_i,k)|}{|\N(\y_j,k) \cup \N(\x_i,k)|},
\label{eq:Jaccard}
\end{align}
where $| \cdot |$ is to calculate the cardinality of a set. Since that Jaccard distance as defined in \eqref{eq:Jaccard} is built upon the overlapping of two local neighborhoods, the ambiguity of the potential matching candidates in the returned $k$-INN is ignored. Thus the previous $k$-RNN-based methods are quite sensitive to the parameter $k$.

To address the aforementioned limitations, in this paper, we exploit \textit{density-adaptive kernels} to carry out the ideas of 
$k$-INN and $k$-RNN to perform efficient and effective reranking for person ReID. 
Specifically, 
%
rather than finding $k$-NN, which has a hard boundary, we apply a smooth kernel function with a local density-adaptive parameter to each sample and use the responses of the kernel function to define reidentification scores that are continuous, real-valued and of individual sample-specific scales.
Thus, our proposed methods can be viewed as a smoothed version of the $k$-INN and the $k$-RNN reranking method.
Unlike in $k$-INN and $k$-RNN, the ambiguities in the ranking lists and the local density information are properly accommodated, yielding improved performance.  

\section{Our Proposals: Density-Adaptive Kernel based Reranking}
\label{sec:our-proposal}

This section introduces a density-adaptive kernel function, describes our proposed density-adaptive kernel based reranking approaches, and then presents some analysis and discussion. 

\subsection{Density-Adaptive Kernel Function}
\label{sec:our-proposal-DAKR-direct}

To begin with, we introduce a density-adaptive kernel function, which serves as the core of our proposed reranking approaches.

Rather than simply keeping a $k$-NN list, we adopt a smooth kernel function to compute the neighborship of samples to accommodate the ambiguity in the ranking list.
Consider a probe sample $\x_i$ and a gallery set $\Y$.
Specifically, we define a smooth kernel function $\kappa(\y_j |\x_i,\sigma_i)$, where $\x_i$ is the location of the kernel function and $\sigma_i>0$ is a local parameter.
For convenience, we choose the \textit{radial basis function}\footnote{We use the radial basis function because of its explicit or implicit connection to the distance function, which will be interpreted later.}
to define 
$\kappa(\y_j |\x_i,\sigma_i)$, \textit{i.e.},
\begin{align}
\kappa(\y_j | \x_i, \sigma_i)=\phi(\frac{\|\y_j - \x_i\|}{\sigma_i}),
\label{eq:re-id-kernel-xi}
\end{align}
where $\phi(\cdot): \R \rightarrow \R^+$ is a monotonically decreasing function. By default, we use $\phi(t)=\exp(-t)$.

To make the kernel function $\kappa(\y_j | \x_i, \sigma_i)$ be adaptive to the density of the samples, the parameter $\sigma_i$ should be density-adaptive.
In a denser region, the parameter $\sigma_i$ should be smaller to make the kernel function more selective (or sensitive) to reject more samples, whereas in a sparser region, the parameter $\sigma_i$ should be larger to make the kernel function relatively inclusive (or less sensitive) to accept more samples.
In this sense, the expected parameter $\sigma_i$ should encode the density information in the local neighborhood of $\x_i$. Thus, as suggested in \cite{Zelnik:NIPS2005}, we define $\sigma_i$ as the distance of $\x_i$ to its $k$-th nearest neighbor $\x_i^{(k)}$ in $\Y$, \textit{i.e.},
\begin{align}
\sigma_i=\|\x_i - \x^{(k)}_i\|,
\label{eq:re-id-kernel-sigma-i}
\end{align}
where $k \ge 1$ is a preset integer. The $\sigma_i$ defined in Eq.~\eqref{eq:re-id-kernel-sigma-i} roughly encodes the density information in the local neighborhood in $\Y$ of $\x_i$.

The advantages of using a smooth kernel function with a density-adaptive parameter are at least twofold:
a) the smoothness of the kernel function preserves the ambiguity of the potential candidates in the ranking list, and b) the local density-adaptive parameter endows the kernel function with an individual sample-specific scale.
 To be more specific, we present the following interpretation. Using $\sigma_i$ as defined in Eq.~\eqref{eq:re-id-kernel-sigma-i}, the quantity $\frac{\|y_j - \x_i\|}{\sigma_i}$ in Eq.~\eqref{eq:re-id-kernel-xi} is density-adaptively re-scaled, such that:
\begin{itemize}
\item $\frac{\|\y_j - \x_i\|}{\sigma_i}=1$ if $\y_j$ is the $k$-th nearest neighbor of $\x_i$;
\item $\frac{\|\y_j - \x_i\|}{\sigma_i} > 1$ if $\y_j$ is farther away than the $k$-th nearest neighbor of $\x_i$;
\item $\frac{\|\y_j - \x_i\|}{\sigma_i}< 1$ if $\y_j$ is included within the $k$-NN set of $\x_i$.
\end{itemize}
Although having such a continuous real-valued quantity is not important for $k$-NN ranking, it will help to determinate $k$-INN and $k$-RNN by capturing and preserving the ambiguity in the ranking lists. In the next subsections, we will show that the proposed smooth kernel function with a density-adaptive local parameter is useful in formulating a \emph{smoothed} version of $k$-INN and a \emph{smoothed} version of $k$-RNN for reranking.

\subsection{Inverse Density-Adaptive Kernel based Reranking (inv-DAKR)}
\label{sec:our-proposal-DAKR-inverse}

Equipped with the density-adaptive kernel function, we are ready to present our inv-DAKR. The key ingredient of inv-DAKR is that, instead of finding the list of $k$-INN directly, we use a smooth kernel function with a density-adaptive parameter to score all gallery samples. 

Recall that the $k$-INN set of the probe sample $\x_i$ is defined by the list of gallery samples that inversely find the probe sample $\x_i$ as one of their $k$-NN. As an analogue, 
we put a smooth kernel function $\kappa(\x |\y_j,\sigma_j)$ at each gallery sample $\y_j$ with an adaptive local parameter $\sigma_j$, with $j=1,\cdots, N$. 
Specifically, we use the \textit{radial basis function} to define $\kappa(\x |\y_j,\sigma_j)$, \textit{i.e.}, 
\begin{align}
\kappa(\x|\y_j,\sigma_j)=\phi(\frac{\|\x - \y_j\|}{\sigma_j}),
\label{eq:re-id-kernel}
\end{align}
where $\sigma_j$ is an adaptive local parameter defined as the distance from $\y_j$ to its $k$-th nearest neighbor $\y_j^{(k)}$, \textit{i.e.},
\begin{align}
\sigma_j=\|\y_j - \y^{(k)}_j\|.
\label{eq:re-id-kernel-sigma-j}
\end{align}
%
In inv-DAKR, rather than finding the list of $k$-INN, we use the kernel function in Eq.~\eqref{eq:re-id-kernel}, which is located at each gallery sample $\y_j$, to \emph{inversely} calculate a score for the probe sample $\x_i$. Then, reranking by inv-DAKR is conducted by descendingly sorting the $N$ scores $\{ \kappa(\x_i| \y_j, \sigma_j) \}_{j=1}^N$ for $\x_i$.

Compared to $k$-INN, inv-DAKR has the following merits:
a) the ambiguity of the potential matching candidates is preserved by 
the adaptively scaled kernel function;
b) the computation cost in the test stage is significantly reduced, because inv-DAKR first computes $N$ scores $\{\kappa(\x_i|\y_j,\sigma_j)\}_{j=1}^N$ and then sort them only once, rather than sorting them for each sample $N$ times as in $k$-INN\footnote{Note that the $N$ density-adaptive parameters $\{\sigma_j\}_{j=1}^N$ can be computed in advance.}.

Note that the density-adaptive scale parameter $\sigma_j = \|\y_j - \y_j^{(k)} \|$, which takes into account the adjacency of the other samples in the gallery set with respect to $\y_j$, is individually defined for each $\y_j \in \Y$. With the help of parameter $\sigma_j$, the \textit{scaled} distance $\frac{\| \x - \y_j \|}{\sigma_j} $ for $\y_j \in \Y$ will yield an \textit{adaptively scaled distance} based \textit{ranking} for the gallery sample $\y_j \in \Y$---this eventually leads to a reranking for $\y_j \in \Y$.
To be more precise, we present the following interpretation: a) if $\x$ lies at the $k$-th nearest neighbor of $\y_j$, then $\frac{\| \x - \y_j \|}{\sigma_j} = 1$; b) if $\x$ lies farther away than the $k$-th nearest neighbor of $\y_j$, then $\frac{\| \x - \y_j \|}{\sigma_j} > 1$; and c) if $\x$ lies nearer than the $k$-th nearest neighbor of $\y_j$ (\textit{i.e.}, $\x$ lies within the list of $k$-NN of $\y_j$), then $\frac{\| \x - \y_j \|}{\sigma_j} < 1$.
Thus, to find all samples $\y_j \in \Y$ belonging to $k$-INN set of $\x$, it is necessary to retain the samples $\y_j \in \Y$ such that $\frac{\| \x - \y_j \|}{\sigma_j} \le 1$, \textit{i.e.}, the $k$-INN set of $\x$ is turned out to be: $\{ \y_j \in \Y: ~ \frac{\| \x - \y_j \|}{\sigma_j} \le 1 \}$. To compare two of the $k$-INN samples of $\x$, we can directly compare the values $\frac{\| \x - \y_j \|}{\sigma_j} $. Therefore, we refer to our inv-DAKR as a smoothed version of the $k$-INN method, and as discussed above, not only the ambiguity in the ranking lists is preserved but also the calculation process is significantly  simplified.

Now, we give an interpretation of \textit{the ambiguity} of the potential matching candidates in inv-DAKR. Note that the continuous real-valued quantity $\frac{\| \x - \y_j \|}{\sigma_j}$ is able to capture \textit{how much} (the ambiguity) for each of the gallery samples $\y_j \in \Y$ \textit{likely} to belonging to the $k$-INN set of $\x$. That is, for each $\y_j \in \Y$, whenever $\frac{\| \x - \y_j \|}{\sigma_j} \rightarrow 1-\epsilon$ where $\epsilon$ is a small value, we confirm that $\y_j$ must lie nearby the boundary of the $k$-INN list; otherwise, if $\frac{\| \x - \y_j \|}{\sigma_j} \ll 1$, we confirm that $\y_j$ is more reliable in the $k$-INN list. The significance of capturing the ambiguity will become more important when we need to merge the ranking list of $k$-NN with the ranking list of $k$-INN to form the reranking list of $k$-RNN. Contrarily, in the $k$-INN approach, the returned results for a probe $\x$ are merely a list of gallery samples $\y_j \in \Y$ where each of them includes $\x$ as its $k$-NN sample. For any sample $\y_j \in \Y$ that does not include $\x$ as one of its $k$-NN samples, it will clearly not be considered in the final sorted list of the $k$-INN.

\subsection{Bidirectional Density-Adaptive Kernel based Reranking (bi-DAKR)}
\label{sec:our-proposal-DAKR-bidirectional}

Recall that $k$-RNN is defined as the intersection of the $k$-NN and $k$-INN of the probe sample $\x_i$. Specifically, in $k$-RNN-based reranking, one first determines the lists of the $k$-NN and $k$-INN, and then take the intersection of the two lists. In bi-DAKR, we first compute the scores via the density-adaptive kernel functions, and then combine and sort those scores to find the final list.  In this sense, bi-DAKR can be viewed as a smoothed version of the $k$-RNN reranking approach.

To implement bidirectional reidentification, we gather the scores from both the direct path 
and the inverse path. 
The density-adaptive kernel functions are well prepared to define the bidirectional reidentification. Note that:

\begin{itemize}
\item The kernel function located at the probe sample $\x_i$, \textit{i.e.}, $\kappa(\y_j|\x_i,\sigma_i) =\phi(\frac{\|\y_j - \x_i\|}{\sigma_i})$,
yields a score for the gallery sample $\y_j$, which can be viewed as \emph{a belief from the perspective of the probe sample $\x_i$ seeking the gallery sample $\y_j$}, where $j=1,\cdots,N$.

\item Similarly, the kernel function located at the gallery sample $\y_j$, \textit{i.e.}, $\kappa(\x_i|\y_j,\sigma_j) =\phi(\frac{\|\x_i - \y_j\|}{\sigma_j})$,
yields a score for the probe sample $\x_i$, which can be viewed as \emph{a belief from the perspective of the gallery sample $\y_j$ seeking the probe sample $\x_i$}.

\end{itemize}

Having computed the $2N$ scores from these bidirectional paths, it is straightforward to combine them. While there are several ways to define bidirectional scores, as investigated in our preliminary work \cite{Guo:ICPR2018}, we prefer to use the following form:
\begin{align}
\chi(\x_i, \y_j)= \phi ( \frac{\|\y_j - \x_i\| \cdot \|\x_i - \y_j\|}{\sigma_i \sigma_j}).
\label{eq:re-id-bilateral-meltout}
\end{align}
Since 
$\|\y_j - \x_i\| \cdot \|\x_i - \y_j\| = \|\y_j - \x_i\|^2$, we have $\chi(\x_i, \y_j)= \phi ( \frac{\|\y_j - \x_i\|^2}{\sigma_i \sigma_j})$. The radial symmetry in the functional form of $\phi(\frac{\| \cdot \|}{\sigma})$ can reduce the calculation of the belief scores $\{ \chi(\x_i, \y_j) \}_{j=1}^N$. Another reason to choose the radial basis function is that the functional form of the radial basis function $\phi(\frac{\| \cdot \|}{\sigma})$ has an explicit connection to the distance function, leading to clear interpretations of DAKR, inv-DAKR and bi-DAKR as smoothed versions of the $k$-NN, $k$-INN and $k$-RNN methods, respectively.

In practice, 
we compute in advance $N$ parameters $\{\sigma_j\}_{j=1}^N$. Then, for a probe sample $\x_i$, we calculate $N$ belief scores $\{ \chi(\x_i, \y_j) \}_{j=1}^N$ and produce the final result by sorting the $N$ scores in descending order.

Compared to $k$-RNN, our bi-DAKR has the following advantages: a) the ambiguity in the ranking list is preserved due to the scores being continuous and real-valued; b) the belief scores are scaled individually and sample-specifically due to the density-adaptive parameter, and c) the computation is convenient due to the symmetry of the specific functional form.

 Now, we will give a precise interpretation of how \textit{the ambiguity} of the potential matching candidates is preserved in bi-DAKR.
%
%
%
In bi-DAKR, as in the case of inv-DAKR, the continuous real-valued quantity $\frac{\| \x - \y_j \|}{\sigma_j}$ is able to preserve the ambiguity during merging the lists of the $k$-NN with $k$-INN to form the $k$-RNN. We consider a probe $\x$ and a gallery sample $\y_j$.
\begin{itemize}
\item In the $k$-NN list, suppose that $\y_j$ does not belong to the $k$-NN samples of $\x$, but lies only slightly farther away than the $k$-th nearest neighbor (\textit{i.e.}, $\x^{(k)})$ of $\x$. That is, $\frac{\| \y_j - \x \|}{\sigma} = 1 + \epsilon > 1$ where $\sigma = \|\x - \x^{(k)} \|$. In this case, whenever in the $k$-INN list, $\x$ lies in the $k$-NN list of $\y_j$ and lies slightly closer to $\y_j$, \textit{i.e.}, $\frac{\|\x - \y_j \|}{\sigma_j} < \frac{1}{1 + \epsilon} < 1$, where $\sigma_j = \|\y_j - \y_j^{(k)} \|$, we will have
        \begin{align}
        \frac{\| \y_j - \x \|}{\sigma} \cdot \frac{\|\x - \y_j \|}{\sigma_j} < (1 + \epsilon) \cdot \frac{1}{1 + \epsilon} =1.
        \end{align}
        This quantity $\frac{\| \y_j - \x \|}{\sigma} \cdot \frac{\|\x - \y_j \|}{\sigma_j} < 1$ suggests $\x$ and $\y_j$ are still $k$-RNN.

\item In the $k$-INN list, suppose that $\y_j$ does not include $\x$ as its $k$-NN sample but that $\x$ lies only slightly farther away than the $k$-th nearest neighbor (\textit{i.e.}, $\y_j^{(k)})$ of $\y_j$. That is, $\frac{\|\x - \y_j \|}{\sigma_j} = 1 + \epsilon > 1$, where $\sigma_j = \|\y_j - \y_j^{(k)} \|$. 
    In this case, whenever in the $k$-NN list of $\x$, $\y_j$ lies in the $k$-NN list of $\x$ and lies slightly closer to $\x$, \textit{i.e.}, $\frac{\|\y_j - \x \|}{\sigma} < \frac{1}{1 + \epsilon} < 1$, where $\sigma = \|\x - \x^{(k)} \|$, we will have $\frac{\|\x - \y_j \|}{\sigma_j} \cdot \frac{\| \y_j - \x \|}{\sigma} < 1$, which again suggests that $\x$ and $\y_j$ are still $k$-RNN.

\end{itemize}
Contrarily, in the two cases listed above, $\y_j$ will not be selected as $k$-RNN of the probe $\x$.

\begin{figure*}[phtb]
\centering
\subfigure[$k$-NN+]{\includegraphics[clip=true,trim=0 0 0 0,width=0.45\columnwidth]{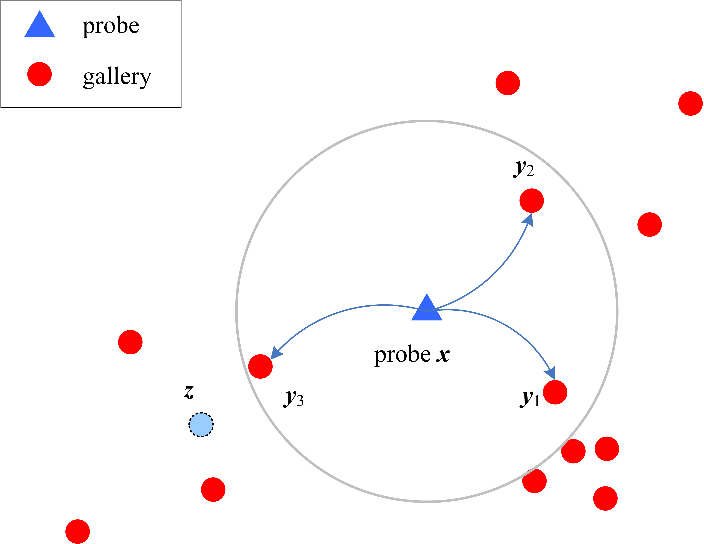} 
\label{fig:illustration-Re-ID-1+}}
\hspace{18pt}
\subfigure[$k$-INN+]{\includegraphics[clip=true,trim=0 0 0 0,width=0.45\columnwidth]{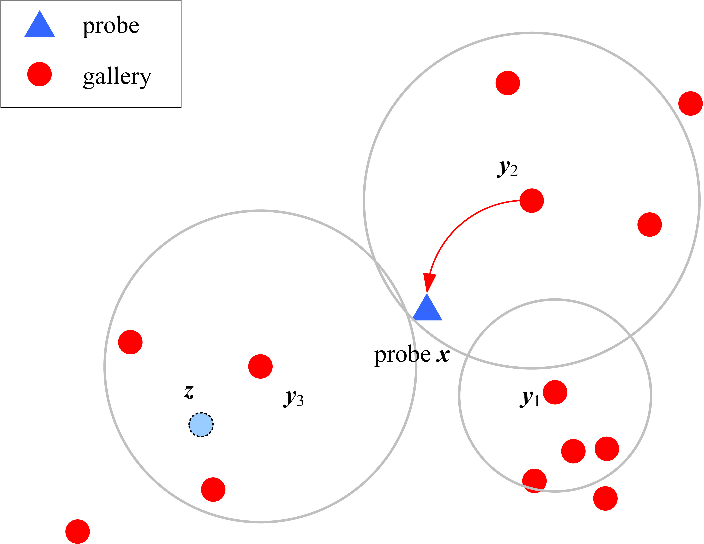} 
\label{fig:illustration-Re-ID-2+}}
\hspace{18pt}
\subfigure[$k$-RNN+]{\includegraphics[clip=true,trim=0 0 0 0,width=0.45\columnwidth]{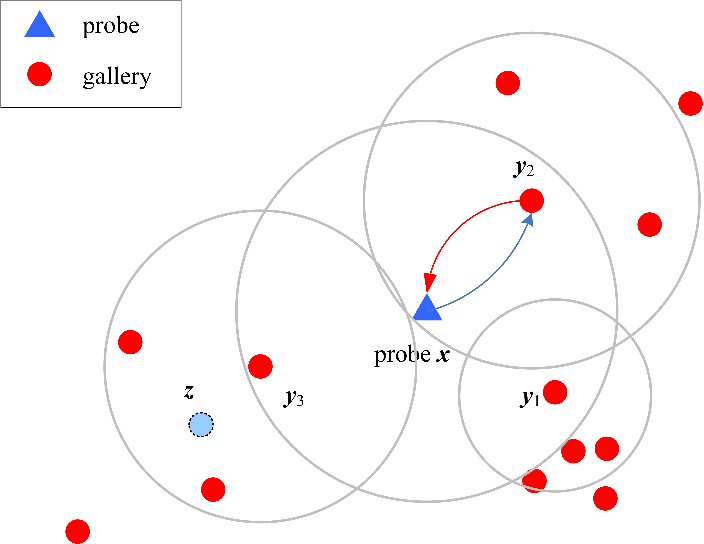} 
\label{fig:illustration-Re-ID-3+}}
\caption{Illustration of ranking and reranking when using an extra probe sample $\z$. (a) $k$-NN+; (b) $k$-INN+ for inverse reidentification; (c) $k$-RNN+, \emph{i.e.}, $k$-NN+ integrated with $k$-INN+ for bidirectional reidentification, where $k=3$. The large circles indicate the boundaries of the local neighborhoods specified by three nearest neighbors. The small blue-filled ball $\z$ indicates an extra probe sample. Compared to Fig.~\ref{fig:illustration-Re-ID}, while the result of $k$-NN in panel (a) does not change, the $k$-INN reranking results in panel (b) are changed because the added ``dummy'' probe sample $\z$ changes the local neighborhood of the gallery sample $\y_3$ such that $\y_3$ no longer takes the probe sample $\x$ as one of its $k$-NN samples. Thus, when combining the results of $k$-NN and $k$-INN, the obtained $k$-RNN reranking list is also changed, as illustrated in panel (c).}
\label{fig:illustration-Re-ID+}
\vspace{-0pt}
\end{figure*}

\subsection{Improving the Reranking Performance with Extra Probe Samples} 

\subsubsection{$k$-NN+, $k$-INN+ and $k$-RNN+}

If a set of probe samples $\X$ are available, we can use the probe samples, except for $\x_i$, to further improve the reranking performance. 
Specifically, we find the top-$k$ best matching candidates to the probe $\x_i$ with the augmented sample set $\X_{-i} \cup \Y$, \textit{i.e.},
\begin{align}
\N(\x_i,k) =\arg\min ^{[1:k]}\limits_{\y_j \in \X_{-i} \cup \Y }~\| \y_j - \x_i \|,
\label{eq:re-id-plus}
\end{align}
where $\X_{-i}$ refers to the probe samples except for the $i$-th sample $\x_i$. We refer to the $k$-NN method with augmented samples as $k$-NN+. Note that the extra probe samples are viewed as ``dummy samples'' because they \emph{occupy} positions in the $k$-NN list but \textit{do not provide any supervision information (\textit{i.e.}, identity labels).} 

Similarly, to identify the $k$-INN, we can also find the best matches of $\y_j \in \Y$ to the probe $\x_i$ with the augmented samples $\X \cup \Y_{-j}$, \textit{i.e.},
\begin{align}
\N(\y_j, k) =\arg\min^{[1:k]}\limits_{\y \in \X \cup \Y_{-j}}~\|\y - \y_j\|,
\label{eq:re-id-inverse-bilateral-plus}
\end{align}
where $\Y_{-j}$ denotes to the gallery set except for $\y_j$. We refer to the $k$-INN reranking method with augmented samples as $k$-INN+.

The assumption behind using such \textit{dummy samples} is that the \textit{dummy samples} lie nearby the correct samples and attract the gallery sample to find it as one of its $k$-NN. 
In $k$-INN+, while the extra probe samples are treated as \textit{dummy samples} without any supervision information, they can improve the local neighborhood of the relevant samples by occupying the correct positions in the $k$-NN list and thus ``pushing away'' some incorrect samples in the local neighborhood. When incorrect samples are pushed away by dummy samples, the reranking list will be refined, leading to an improved performance.

As expected, by combining $k$-NN+ and $k$-INN+, we can obtain a $k$-RNN reranking method, termed as $k$-RNN+. More specifically, we compute $k$-NN and $k$-INN with all samples (\textit{i.e.}, the \textit{dummy} samples) as usual and then ignore the \textit{dummy} samples from the lists of $k$-NN and $k$-INN when finding the reranking result. To demonstrate the effect of using the extra probe samples, we illustrate $k$-NN+, $k$-INN+, and $k$-RNN+ in Fig.~\ref{fig:illustration-Re-ID+}. As can be observed, with one extra probe sample, the local neighborhood of $\y_2$ is \textit{discriminately} improved, 
leading to different results of $k$-INN+ and $k$-RNN+.

\subsubsection{inv-DAKR+ and bi-DAKR+}

In inv-DAKR and bi-DAKR, the local density-adaptive parameter $\sigma_i$ is very important. If some extra probe samples $\X$ are also available, we can define $\sigma_i$ with the assistance of the extra probe samples. To be more specific, we set $\sigma_i$ as the distance of $\x_i$ to its $k$-th nearest neighbor $\x^{(k)}_i$ 
with the samples in the augmented set $\X \cup \Y$ rather than with the samples only in $\Y$.
For inv-DAKR+, we use Eq.\eqref{eq:re-id-kernel} to score each gallery sample and then sort the samples to find the top $k$ results, excluding the \textit{dummy} samples. Similarly, for bi-DAKR+, we use Eq.\eqref{eq:re-id-bilateral-meltout} to score each gallery sample, and then sort them to find the top $k$ results, excluding the \textit{dummy} samples.

We refer to the reranking methods based on inv-DAKR and bi-DAKR with the help of extra probe samples as inv-DAKR+ and bi-DAKR+, respectively.

\myparagraph{Remark}
In $k$-INN+ and $k$-RNN+,  
if the added extra probe samples can \textit{discriminately} 
improve the local neighborhoods of the gallery samples, then performance improvements will be achieved, whereas in inv-DAKR+ and bi-DAKR+, the addition of extra probe samples not only \textit{discriminately} improves the local neighborhood but also helps to yield a more accurate density-adaptive parameter, leading to further performance improvements. Nevertheless, if the extra probe samples cannot provide correct information to \textit{discriminately} improve the local neighborhoods of the gallery samples, the improvement cannot be observed.

\subsection{Analysis and Discussion}

\subsubsection{Effectiveness of inv-DAKR and bi-DAKR}
Compared to previous work \cite{Korn:ASR2000, Qin:CVPR2011, Leng:ICME2013, Leng:MTA2015, Garcia:2017TIP, Bai:TIP2016, Zhong:CVPR2017,Yu:BMVC2017}, the effectiveness of inv-DAKR and bi-DAKR comes from three aspects.
First, rather than \emph{using a binary decision, i.e., within or not within the list of the $k$-NN}, the radial basis functions used in inv-DAKR and bi-DAKR are able to model the ambiguity in the list of potential candidates.
Second, the used local parameter encodes the density information in each local neighborhood and thus makes the kernel functions in inv-DAKR and bi-DAKR individually and sample-specifically scaled. 
Third, in inv-DAKR and bi-DAKR, all gallery samples are used to find the $k$-INN and $k$-RNN of the probe sample, rather than using only a small subset of the gallery samples (\textit{e.g.}, the $k$-NN samples of the probe sample). In particular, if the true matches are not in the $k$-NN list of the probe sample, then they have no chance to be found if the $k$-INN and $k$-RNN are computed based only on the $k$-NN of the probe sample.

\begin{table*}
  \small
  \caption{Computational complexity comparison with unilateral information.} 
  \label{tab:Time-Complexity}
  \vspace{4pt}
  \centering
  \begin{tabular}{c|c|c}
    \hline
    \textbf{Methods} &\multicolumn{2}{c}{\textbf{Complexity}}  \\
    \hline
    $k$-NN & \multicolumn{2}{c}{$O(N + N\log_2N)$} \\
    $k$-INN\cite{Korn:ASR2000} & \multicolumn{2}{c}{$O(N(N+1)[1+\log_2(N+1)]+N\log_2N)$} \\
    $k$-INN\cite{Korn:ASR2000}+ & \multicolumn{2}{c}{$O(N(N+M)[1+\log_2(N+M)]+N\log_2N)$} \\
    $k$-RNN & \multicolumn{2}{c}{$O(N(N+1)[1+\log_2(N+1)]+N(1+\log_2N))$} \\
    $k$-RNN+ & \multicolumn{2}{c}{$O(N(N+M)[1+\log_2(N+M)]+(N+M)(1+\log_2(N+M)))$} \\
    \hline
    inv-DAKR &\multirow{4}{*}{offline} &\multirow{2}{*}{$O(N^2 + N^2\log_2N)$}\\
    bi-DAKR  &  &\\
    \cline{3-3}
    inv-DAKR+ & &\multirow{2}{*}{$O(N(N+M) + N(N+M)\log_2(N+M))$}\\
    bi-DAKR+ &  &\\
    \hline
    inv-DAKR &\multirow{4}{*}{online} &$O(N+N\log_2N)$\\
    inv-DAKR+ & &\\
    \cline{3-3}
    bi-DAKR  &  &$O(2N+2N\log_2N)$\\
    \cline{3-3}
    bi-DAKR+ &  &$O((2N+M)+N\log_2N+(N+M)\log_2(N+M))$\\
    \hline
  \end{tabular}
\end{table*}

\subsubsection{Efficiency of inv-DAKR and bi-DAKR}

\begin{figure*}[ht]
\vspace{-0pt}
\centering
\small
\subfigure[rank-1]{\includegraphics[clip=true,trim=12 5 15 0,width=0.245\columnwidth]{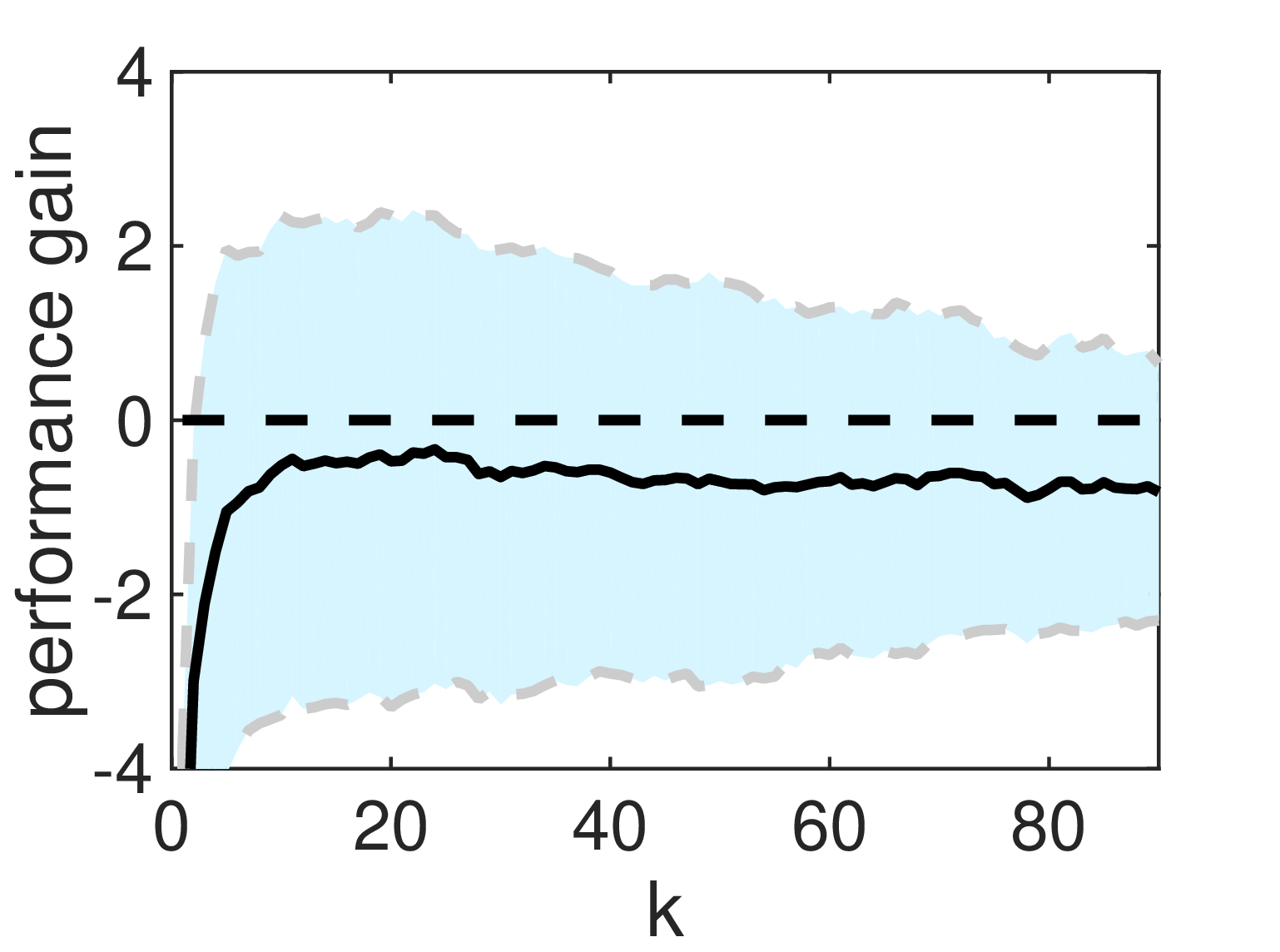}}
\subfigure[rank-5]{\includegraphics[clip=true,trim=12 5 15 0,width=0.245\columnwidth]{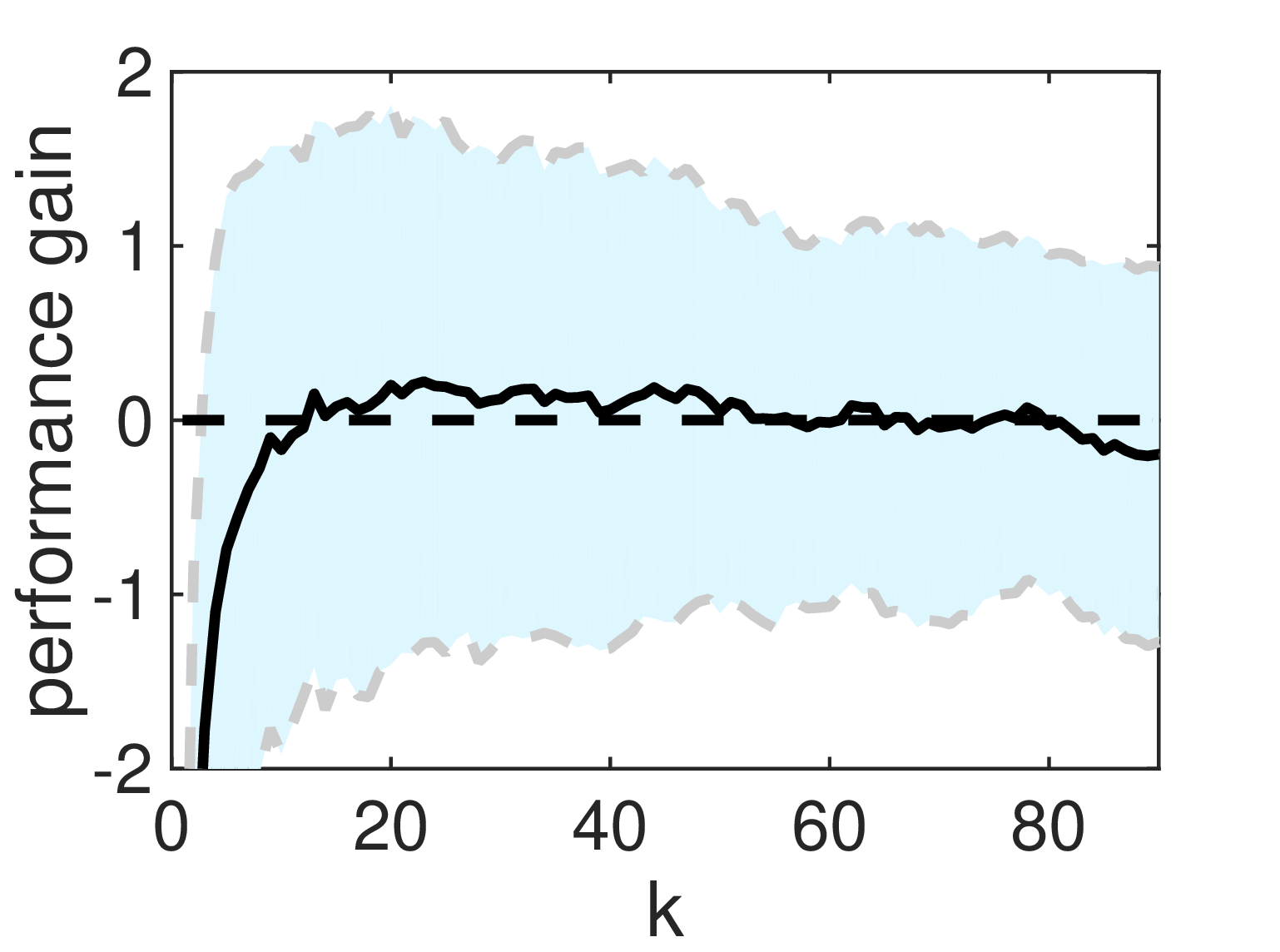}}
\subfigure[rank-10]{\includegraphics[clip=true,trim=12 5 15 0,width=0.245\columnwidth]{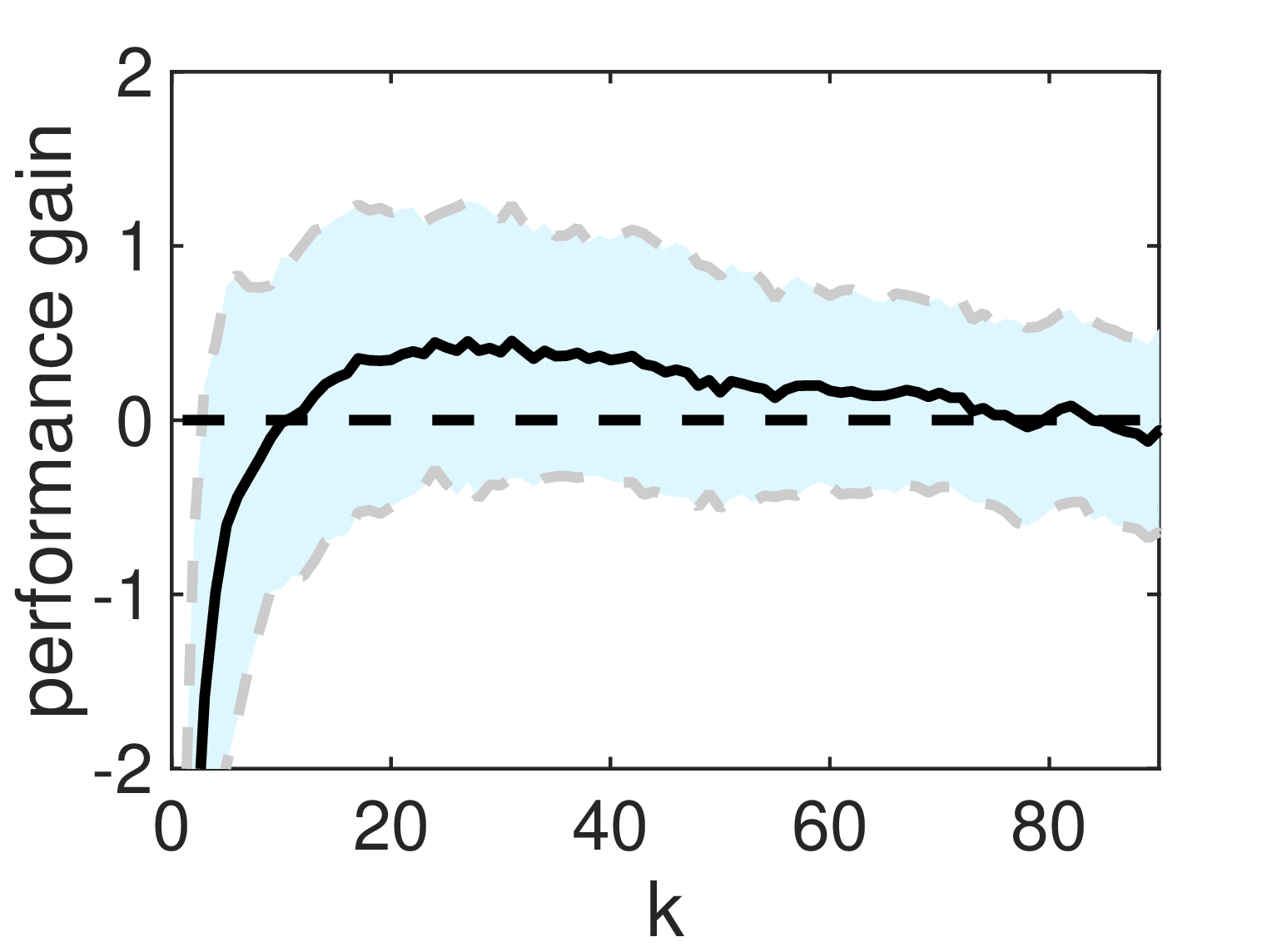}}
\subfigure[rank-20]{\includegraphics[clip=true,trim=12 5 15 0,width=0.245\columnwidth]{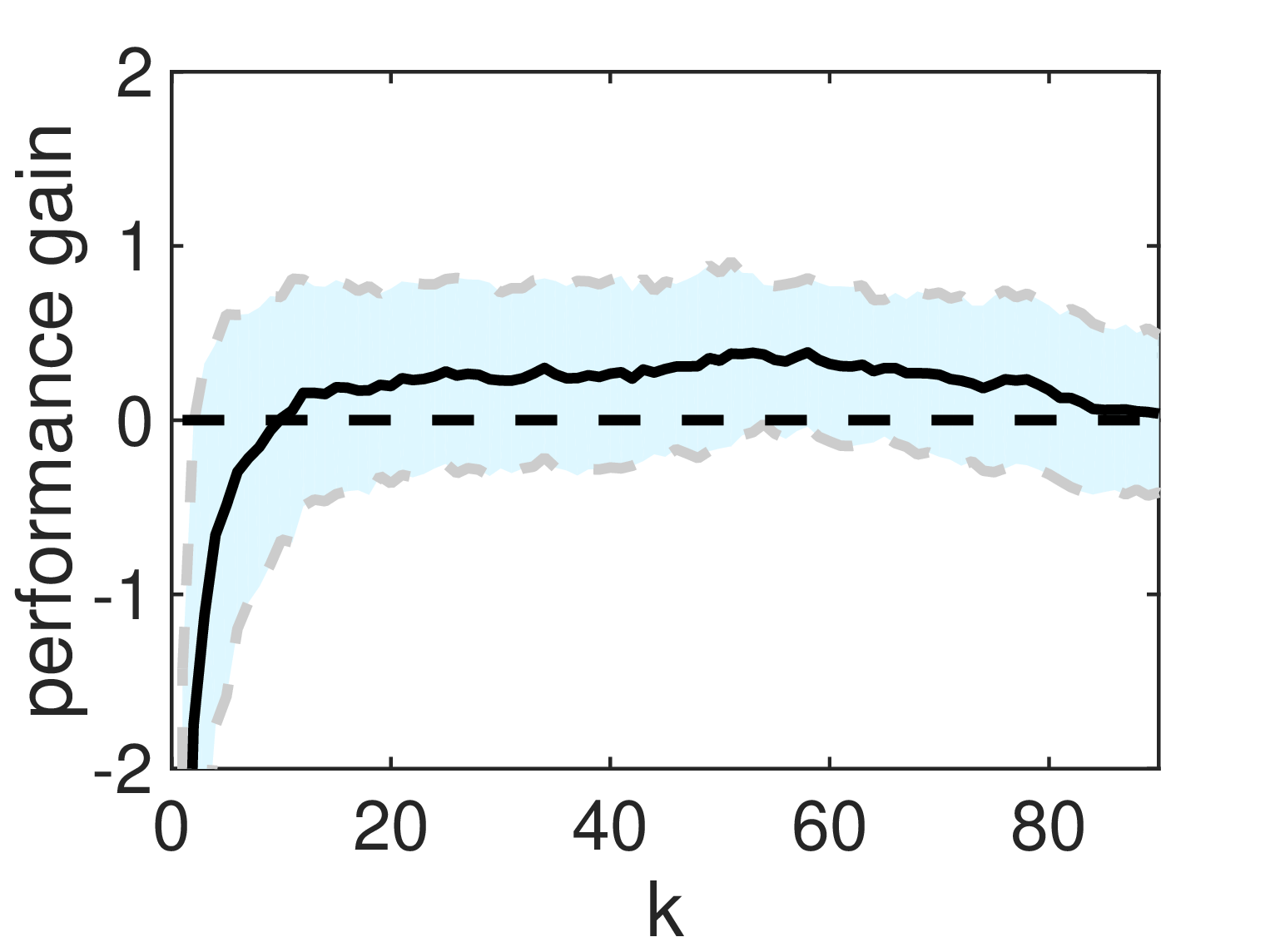}}
\caption{
 Average performance gain of inv-DAKR as a function of $k$ in a perfect one-to-one matching scenario.}
\label{fig:increment_inv_one_2}
\end{figure*}

\begin{figure*}[ht]
\vspace{-2pt}
\centering
\small
\subfigure[rank-1]{\includegraphics[clip=true,trim=12 5 15 0,width=0.245\columnwidth]{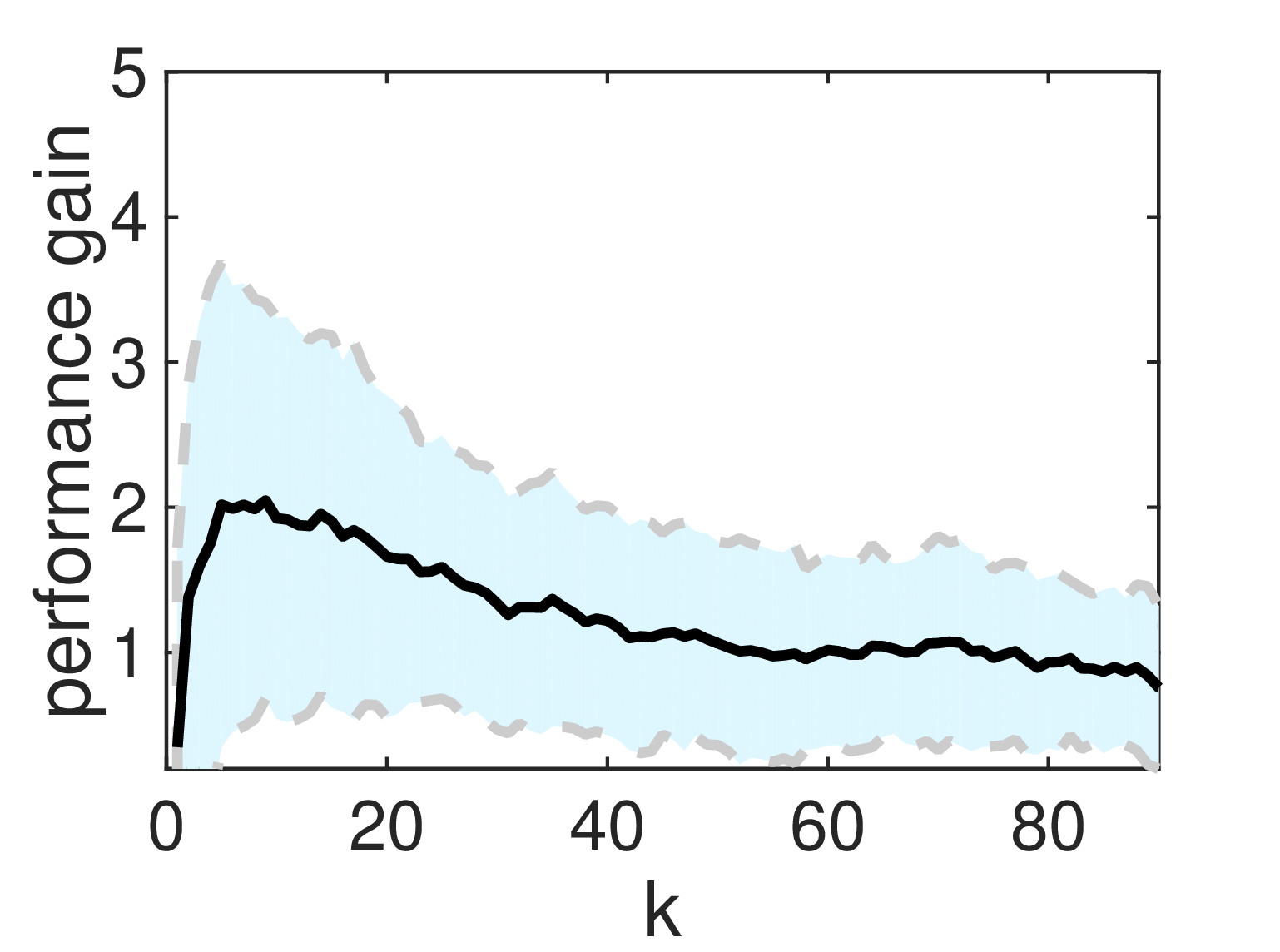}}
\subfigure[rank-5]{\includegraphics[clip=true,trim=12 5 15 0,width=0.245\columnwidth]{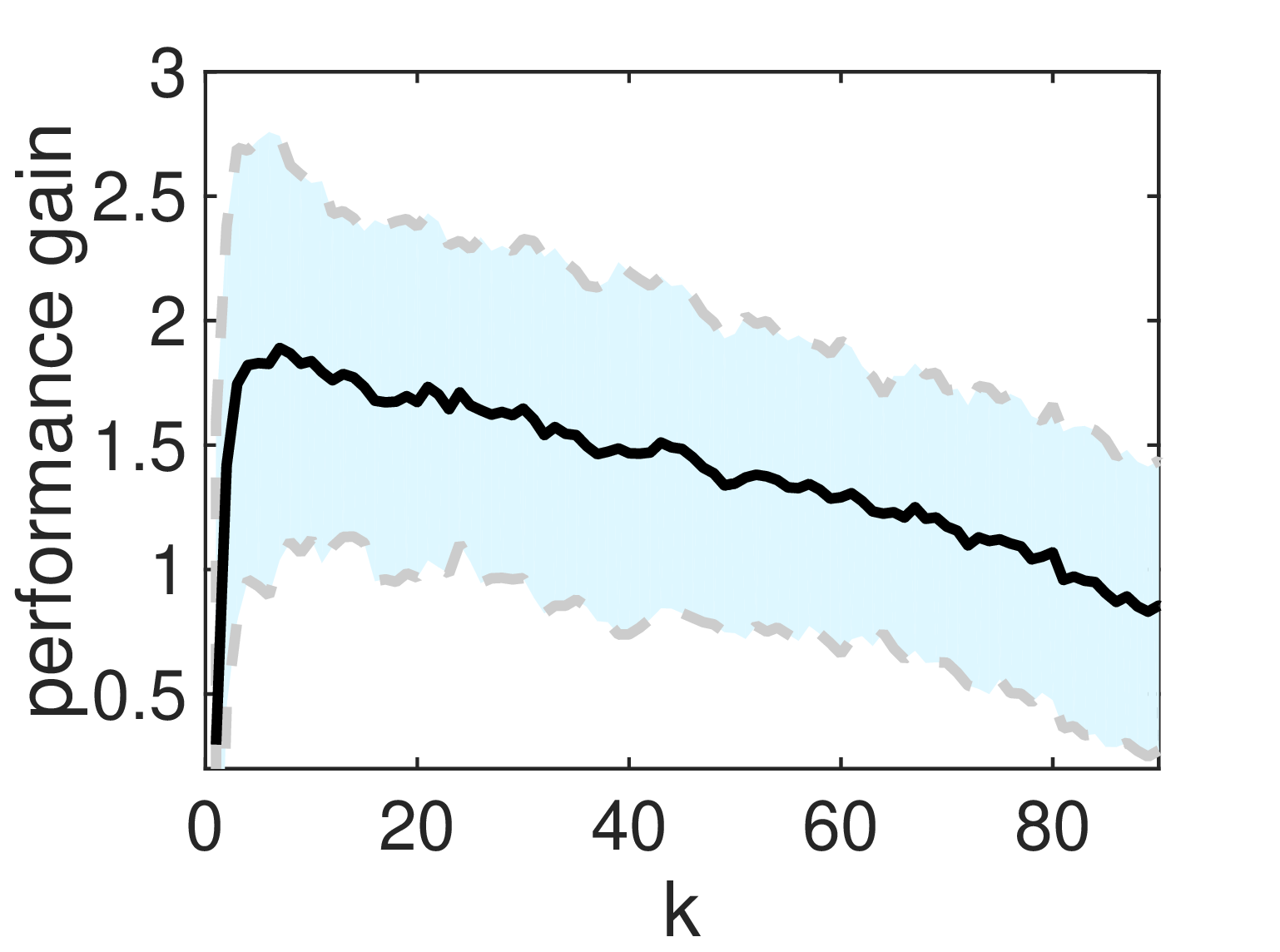}}
\subfigure[rank-10]{\includegraphics[clip=true,trim=12 5 15 0,width=0.245\columnwidth]{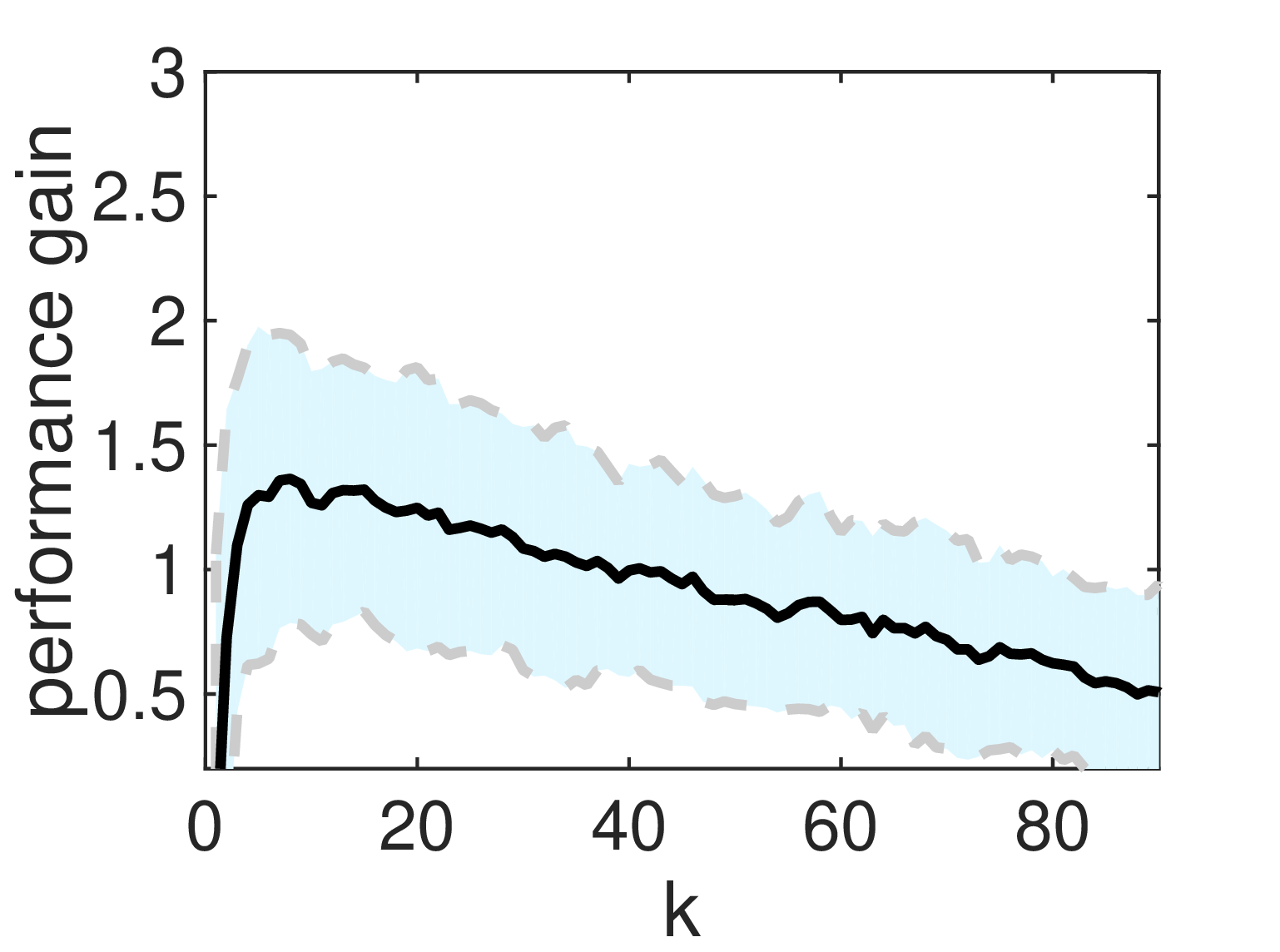}}
\subfigure[rank-20]{\includegraphics[clip=true,trim=12 5 15 0,width=0.245\columnwidth]{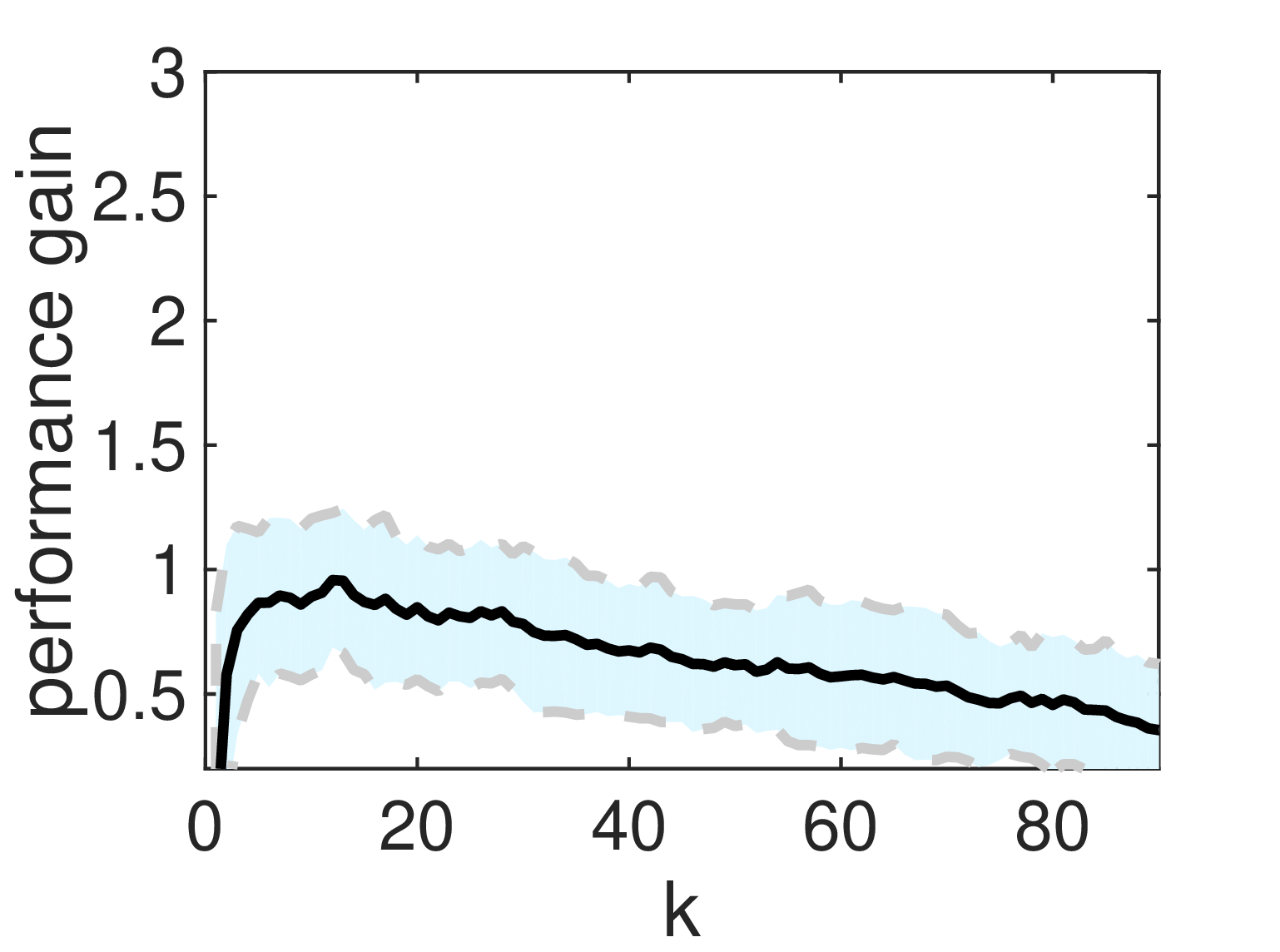}}
\caption{ Average performance gain 
of bi-DAKR as a function of $k$ in a perfect one-to-one matching scenario.} 
\label{fig:increment_bi_one_2}
\vspace{-1mm}
\end{figure*}

\begin{figure*}[ht]
\vspace{-2pt}
\centering
\small
\subfigure[rank-1]{\includegraphics[clip=true,trim=12 5 15 0,width=0.245\columnwidth]{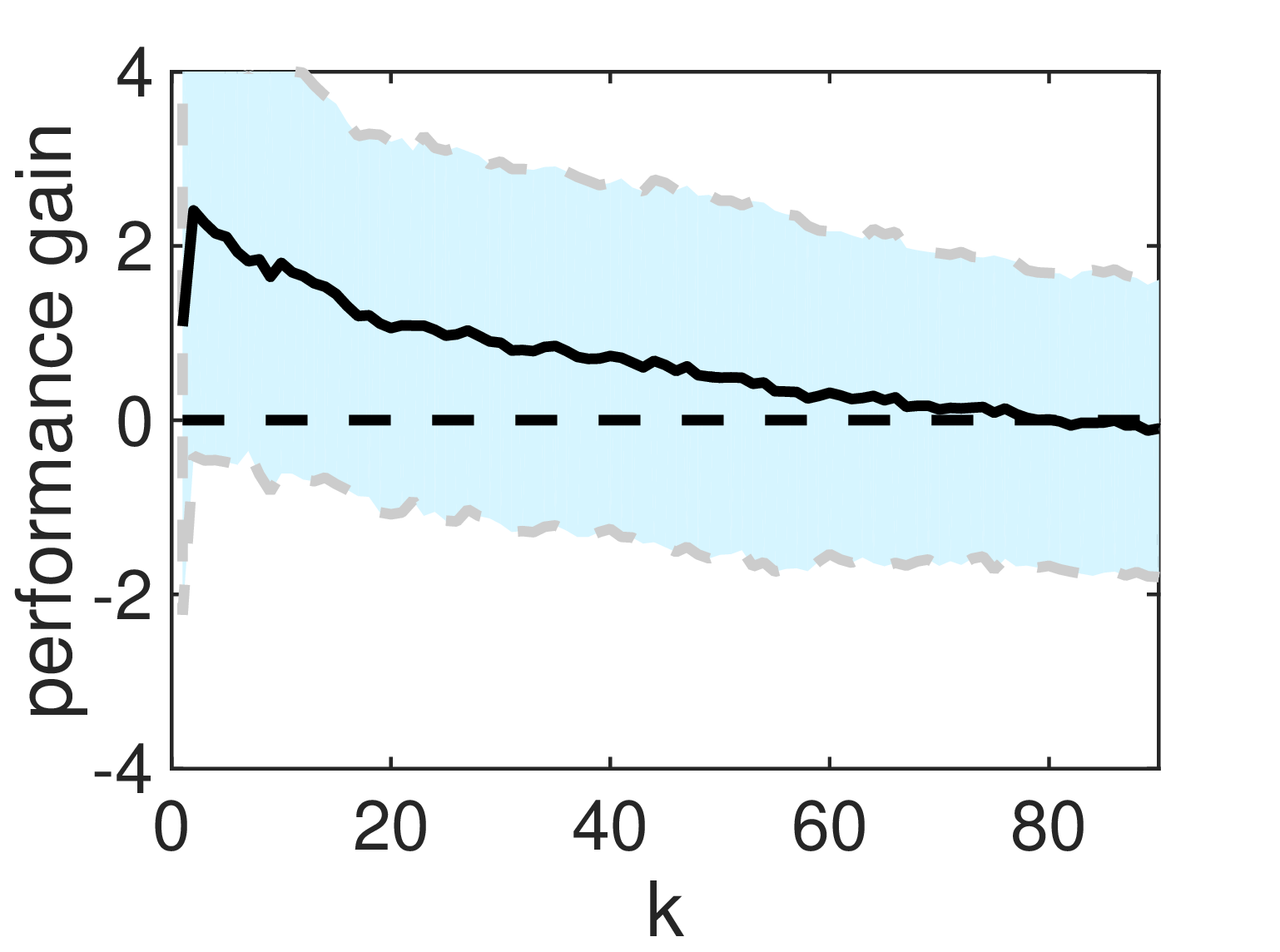}}
\subfigure[rank-5]{\includegraphics[clip=true,trim=12 5 15 0,width=0.245\columnwidth]{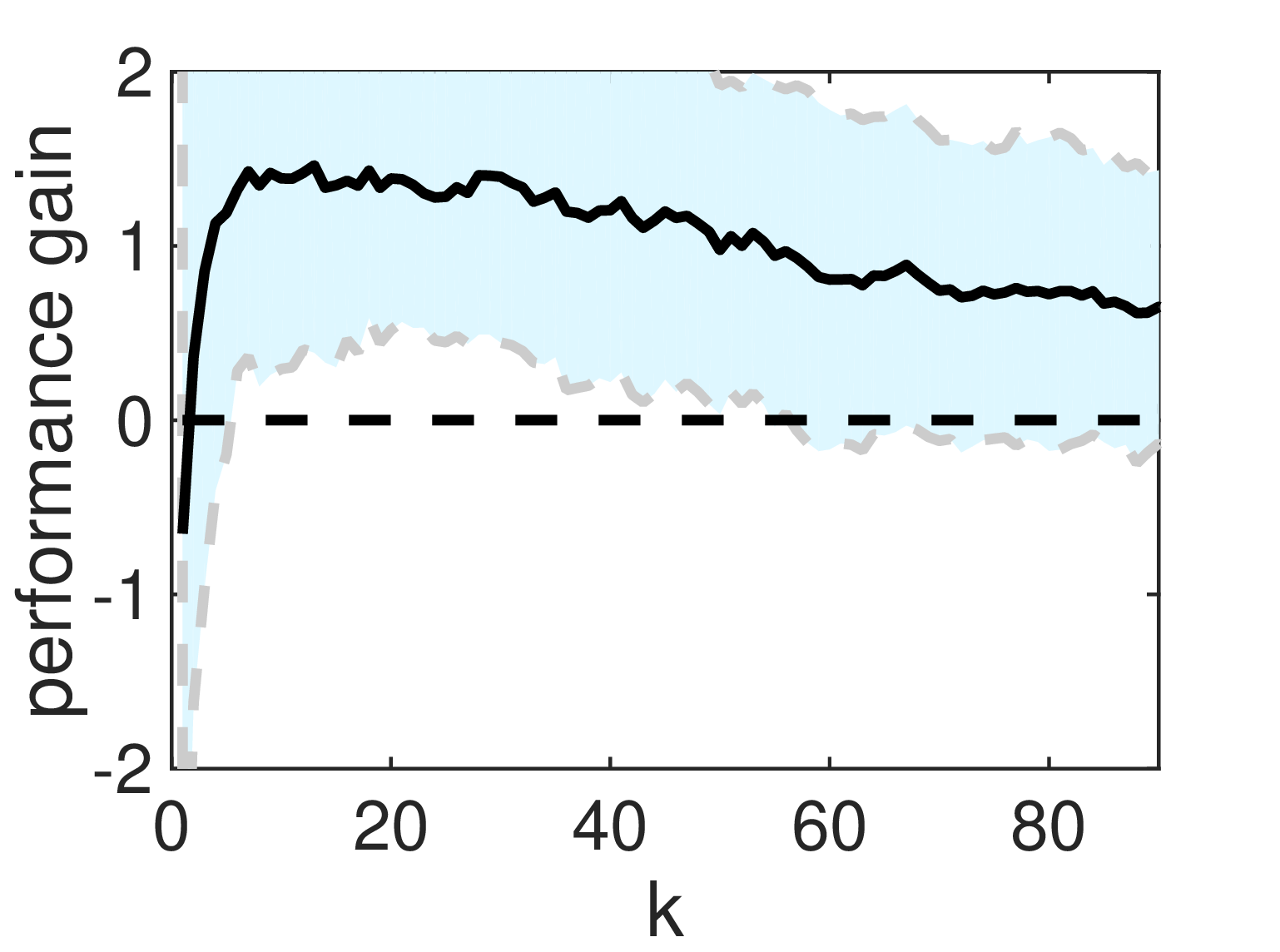}}
\subfigure[rank-10]{\includegraphics[clip=true,trim=12 5 15 0,width=0.245\columnwidth]{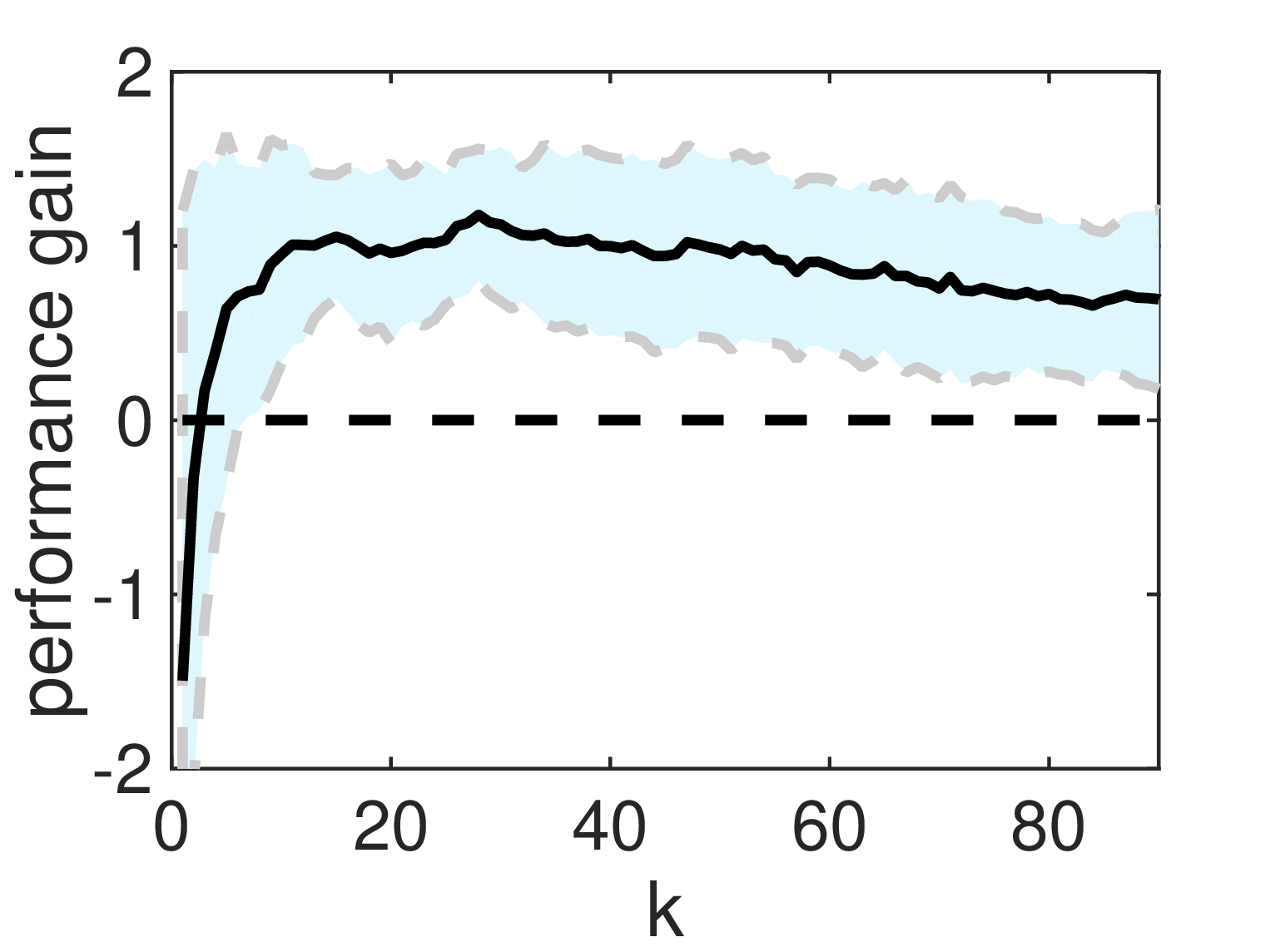}}
\subfigure[rank-20]{\includegraphics[clip=true,trim=12 5 15 0,width=0.245\columnwidth]{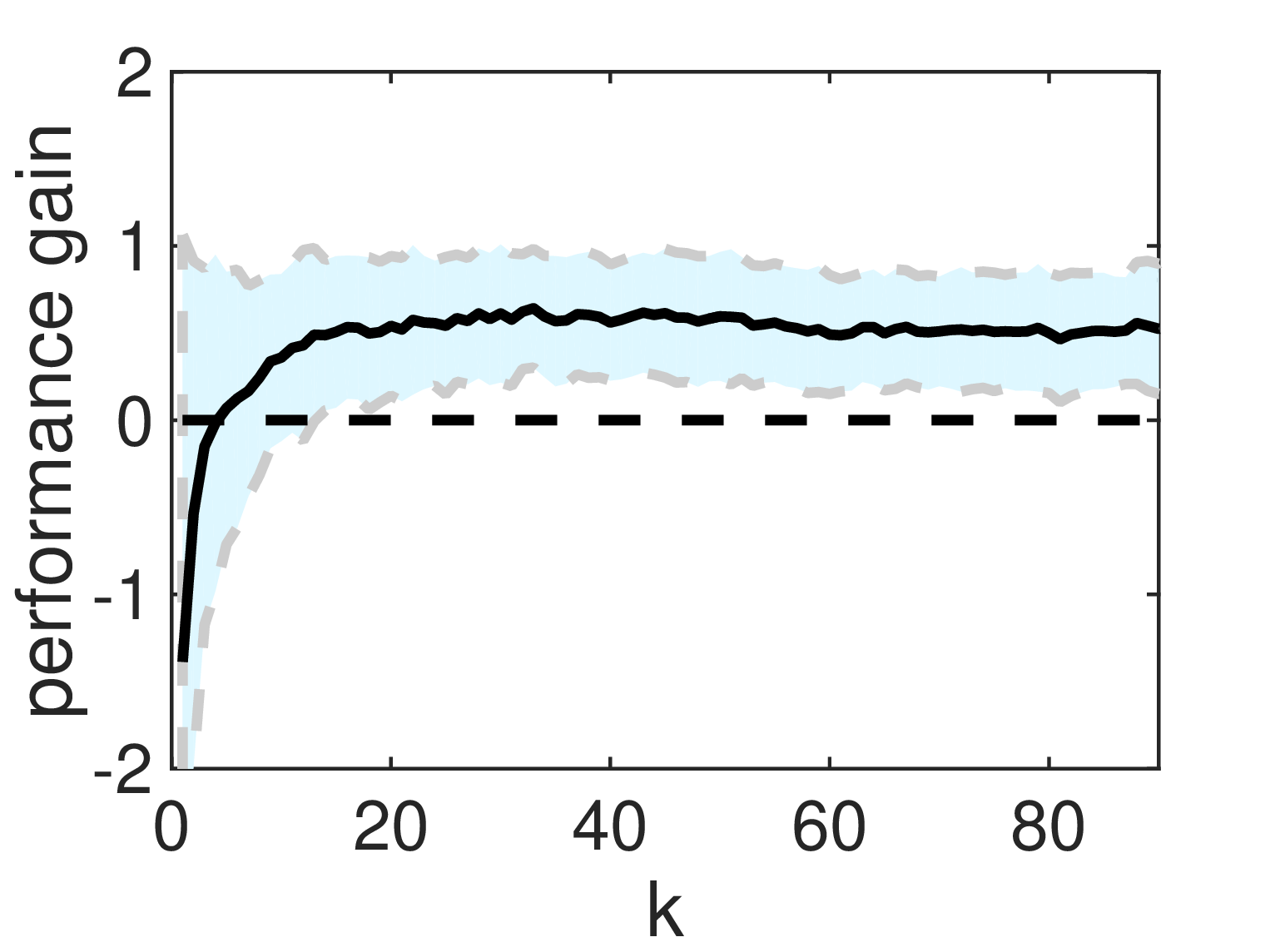}}
\caption{ Average performance gain 
of inv-DAKR+ as a function of $k$ in a perfect one-to-one matching scenario.} 
\label{fig:increment_inv_set_2}
\vspace{-1mm}
\end{figure*}

\begin{figure*}[ht]
\vspace{-2pt}
\centering
\small
\subfigure[rank-1]{\includegraphics[clip=true,trim=12 5 15 0,width=0.245\columnwidth]{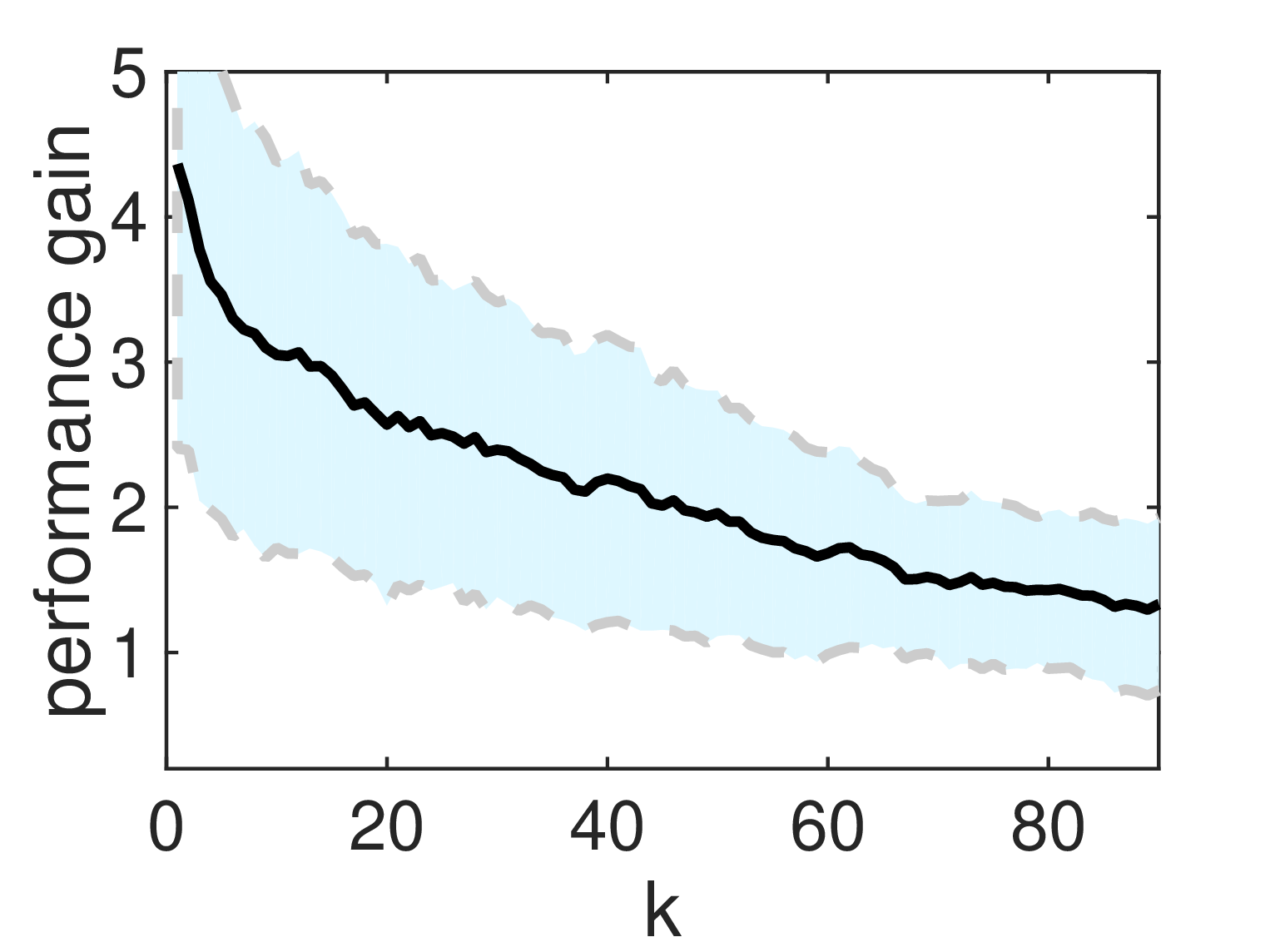}}
\subfigure[rank-5]{\includegraphics[clip=true,trim=12 5 15 0,width=0.245\columnwidth]{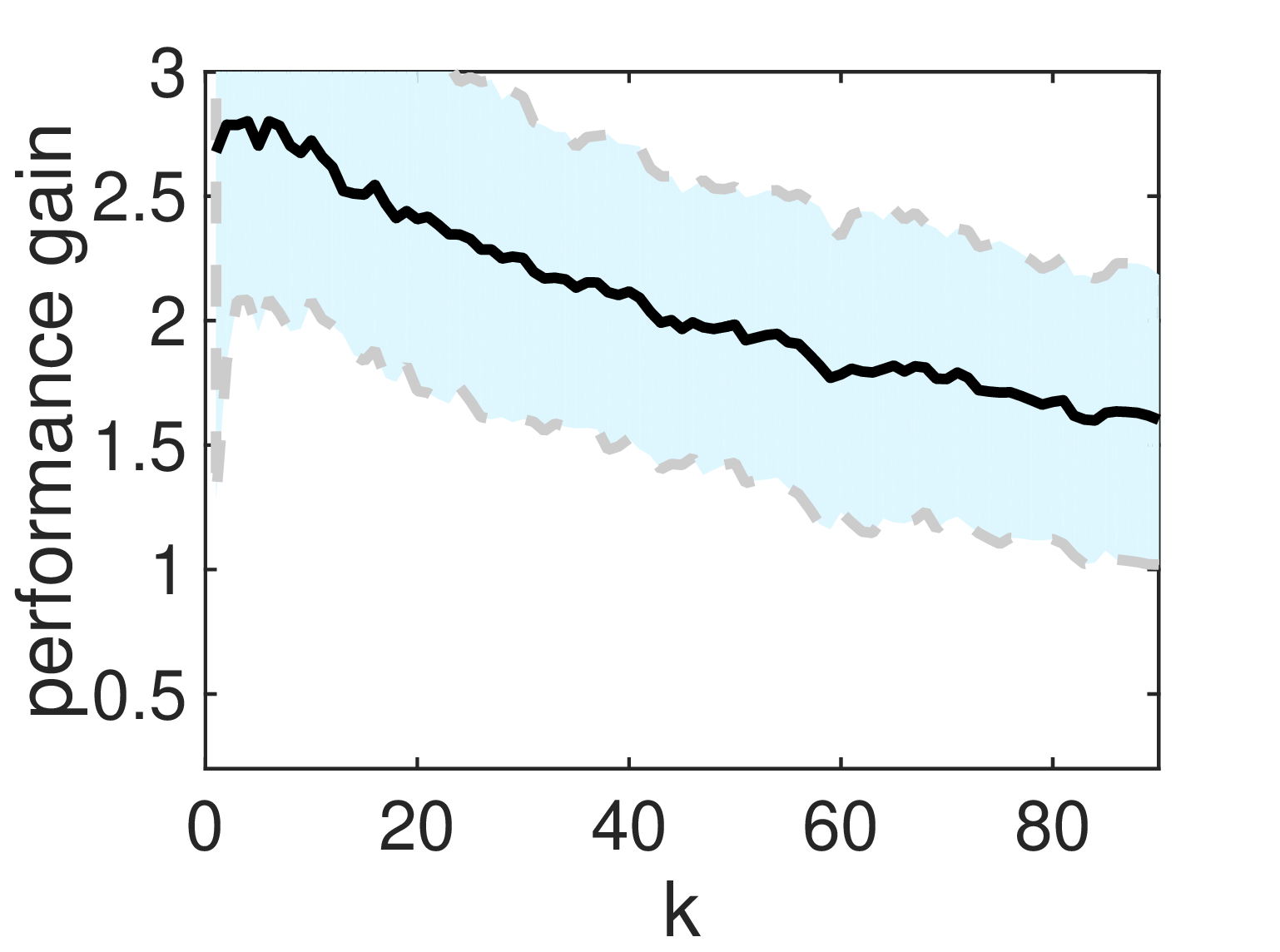}}
\subfigure[rank-10]{\includegraphics[clip=true,trim=12 5 15 0,width=0.245\columnwidth]{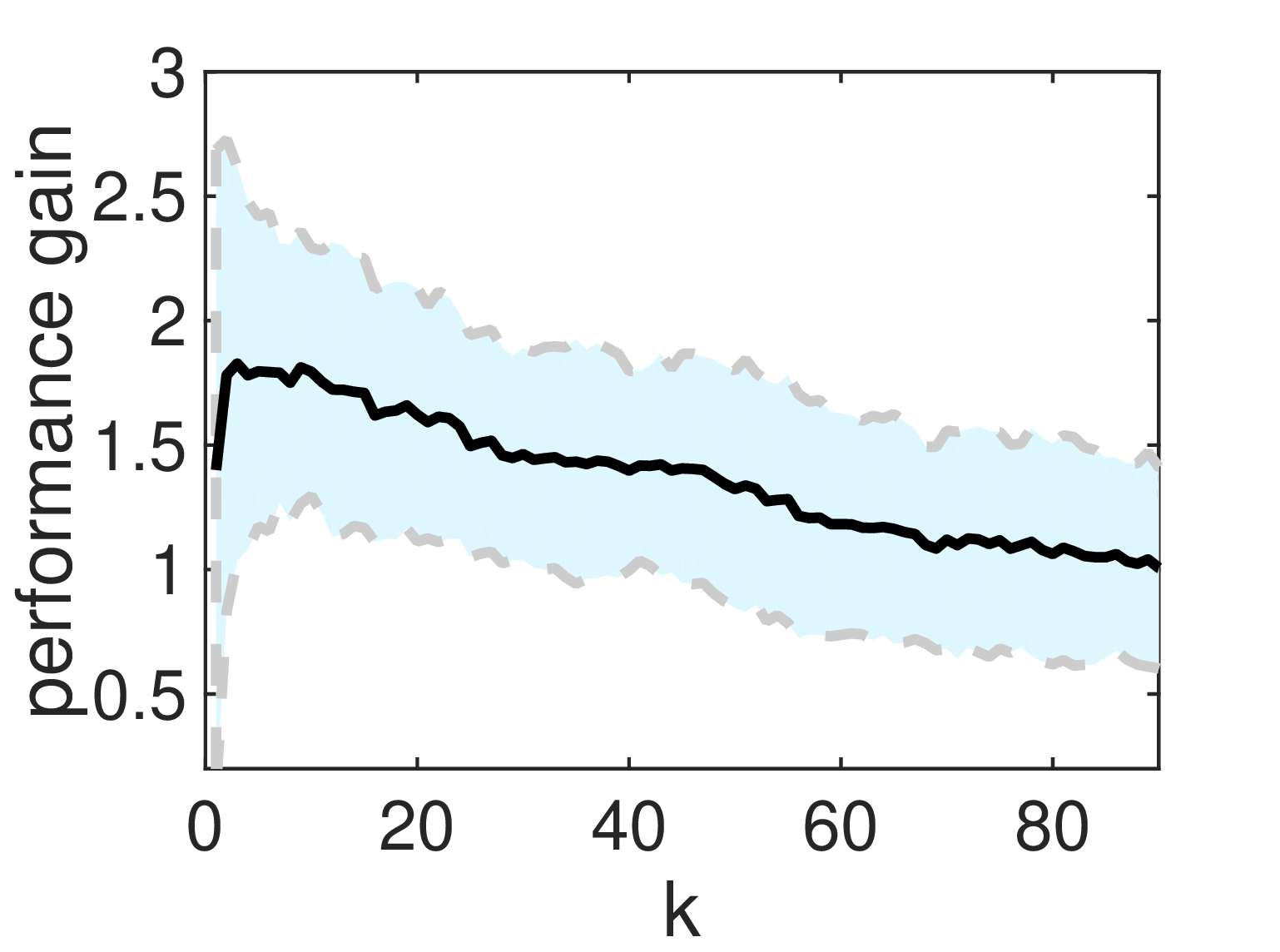}}
\subfigure[rank-20]{\includegraphics[clip=true,trim=12 5 15 0,width=0.245\columnwidth]{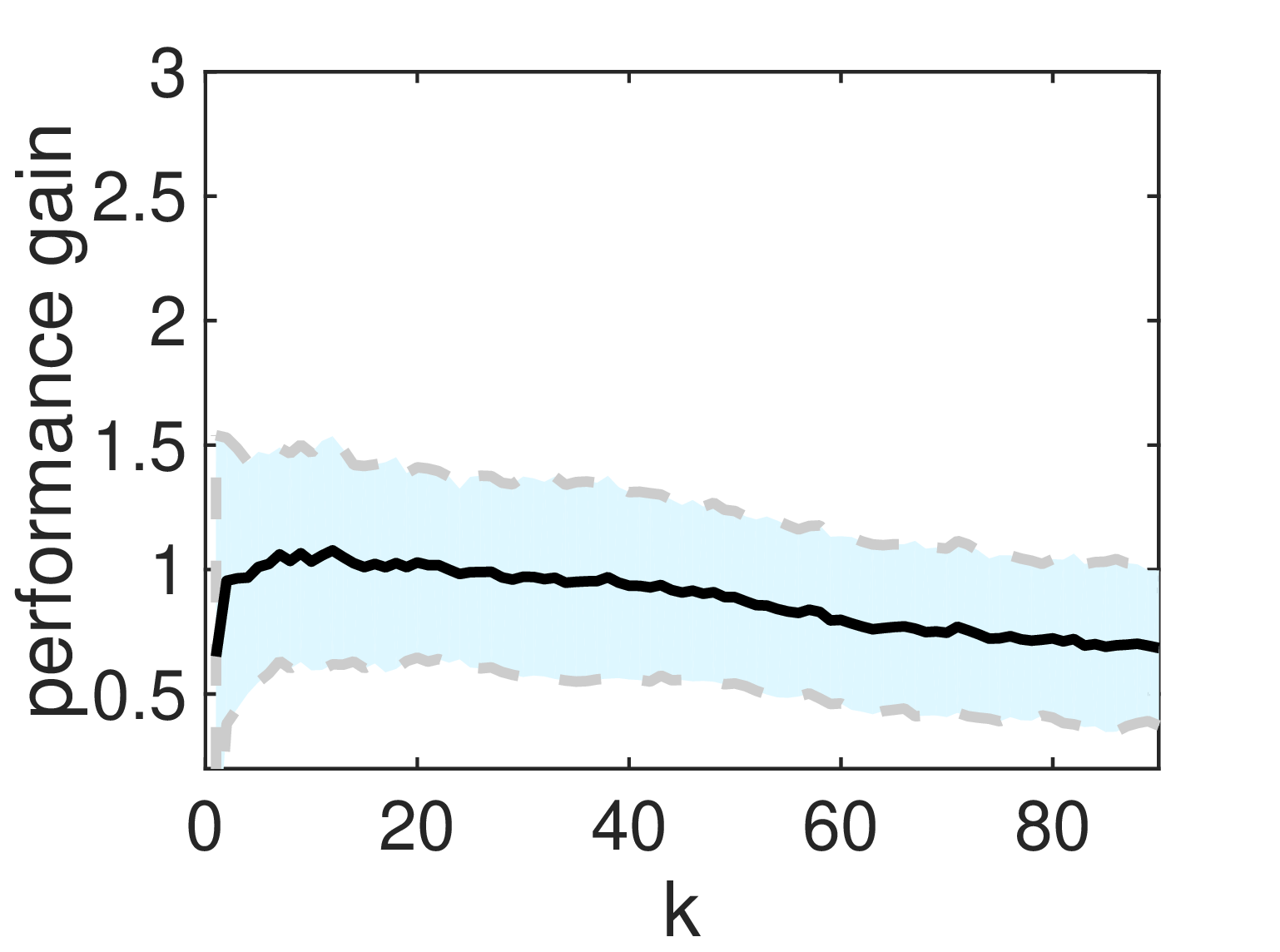}}
\caption{ Average performance gain 
of bi-DAKR+ as a function of $k$ in a perfect one-to-one matching scenario.} 
\label{fig:increment_bi_set_2}
\vspace{-1mm}
\end{figure*}

The efficiency of inv-DAKR and bi-DAKR for reranking comes from the fact that the radial symmetry of the kernel function formulation with the help of a density-adaptive parameter $\sigma_j$ (which can be computed in advance) reduces 
the redundant computation cost during sorting.
Note that in bi-DAKR, only the density-adaptive parameter $\sigma_i$ needs to be computed during the testing phase.
Therefore, the calculation process of our proposals can be separated into two parts: an \textit{offline} part and an \textit{online} part. Specifically, for inv-DAKR and bi-DAKR, all of the sample-specific scales $\sigma_j$ for each sample $\y_j$ in the gallery set $\Y$ can be calculated in advance at offline phase (Eq.\eqref{eq:re-id-kernel-sigma-j}). Then, for each new probe $\x$, it only needs to deal with the operations related to $\x$, such as Eq.\eqref{eq:re-id-kernel} for inv-DAKR, or Eq.\eqref{eq:re-id-kernel-sigma-i} and Eq.\eqref{eq:re-id-bilateral-meltout} for bi-DAKR. Therefore, the calculational complexity is also separated into an \textit{offline} part and an \textit{online} part.
For clarity, we list the detailed computational complexity of inv-DAKR and bi-DAKR compared to the baselines ($k$-NN, $k$-INN and $k$-RNN) in Table \ref{tab:Time-Complexity}. The comparison of the time costs will also be provided from experiments in Section~\ref{sec:time cost}.

\begin{table*}[!ht]
  \scriptsize
  \caption{Comparison on GRID, PRID450S, VIPeR, and CUHK03 with different features.}
  \label{tab:PRID450s-VIPeR-CUHK03}
  \vspace{4pt}
  \centering
  \setlength{\tabcolsep}{1.5mm}{
  \resizebox{\textwidth}{!}{
  \begin{tabular}{c|c|c c c|c c c|c c c|c c c|c c c}
     \hline
     \multirow{2}{*}{\textbf{Feature}} & \multirow{2}{*}{\textbf{Methods}} & \multicolumn{3}{c|}{\textbf{PRID450S}} & \multicolumn{3}{c|}{\textbf{VIPeR}} & \multicolumn{3}{c|}{\textbf{CUHK03 (labeled)}} & \multicolumn{3}{c}{\textbf{CUHK03 (detected)}}  & \multicolumn{3}{c}{\textbf{GRID}} \\
     \cline{3-17}
      & &\textbf{r=1} & \textbf{r=10} & \textbf{r=20} & \textbf{r=1} & \textbf{r=10} & \textbf{r=20} & \textbf{r=1} & \textbf{r=5} & \textbf{r=10} & \textbf{r=1} & \textbf{r=5} & \textbf{r=10}  &\textbf{r=1} & \textbf{r=10} & \textbf{r=20}\\\hline
     \hline
     \multirow{5}{*}{LOMO} & $k$-NN & 59.78 & 90.09 & 95.29 & 41.08 & 82.34 & 91.27 & 50.85 & 81.38 & 91.14 & 44.45 & 78.70 & 87.65 &16.56 &41.84 &52.40 \\
     \cline{2-2}
     &$k$-INN\cite{Korn:ASR2000}  &51.38 &90.04 &94.58 &35.32 &82.25 &90.85 &40.14 &80.73 &90.74 &36.65 &78.15 &88.75 &21.52  &44.88 &55.68\\
     \cline{2-2}
     &$k$-RNN  &45.51 &84.80 &91.69 &29.40 &77.47 &90.28 &40.74 &78.03 &89.09 &36.70 &73.35 &84.65 &19.44 &42.96  &54.56\\
     \cline{2-2}
     &inv-DAKR & 59.24 & 90.44 & 95.29 & 41.61 & 83.10 & 91.84 & 52.56 & 83.08 & 91.74 & 47.35 & 79.90 & 89.75 & 19.84 &\textbf{45.44} &56.24\\
     \cline{2-2}
     &bi-DAKR  & 61.42 & 92.40 & 96.93 & 42.97 & 83.86 & 92.41 & 53.45 & 84.74 & 92.84 &48.10 & 80.80 & 90.05 &19.60 &44.48 &\textbf{56.40} \\
     \cline{2-2}
     &$k$-INN\cite{Korn:ASR2000}+                    &58.09 &90.53 &94.80 &41.36 &83.64 &91.71 &52.05 &83.63 &91.79 &46.75 &81.05 &89.30&\textbf{21.60} &45.12 &56.00 \\
     \cline{2-2}
     &$k$-RNN+                    &60.31 &90.04 &94.76 &41.11 &82.56 &91.36 &51.00 &81.58 &91.09 &44.55 &78.70 &87.40 &17.36 &42.88  &54.48 \\
     \cline{2-2}
     &inv-DAKR+                   &60.67 &91.47 &96.00 &43.07 &83.83 &92.12 &53.00 &82.48 &89.19 &48.90 &77.95 &87.50 &19.68 &44.56 &55.68\\
     \cline{2-2}
     &bi-DAKR+                    &\textbf{63.29} &\textbf{93.02} &\textbf{97.29} &\textbf{43.83} &\textbf{84.27} &\textbf{92.66} &\textbf{55.11} &\textbf{85.58} &\textbf{92.99} &\textbf{50.00} &\textbf{82.30} &\textbf{90.70}&19.36 &44.48 &56.24 \\\hline
     \hline
     \multirow{5}{*}{GOG} & $k$-NN & 68.00 & 94.36 & 97.64 & 49.68 & 88.67 & 94.53 & 68.47 & 90.69 & 95.84 & 64.10 & 88.40 & 94.30 &24.80 &58.40 &68.88 \\
     \cline{2-2}
     &$k$-INN\cite{Korn:ASR2000}  &53.11 &94.22 &97.60 &41.61 &87.41 &94.59 &52.50 &90.49 &96.95 &49.10 &87.55 &94.35&\textbf{27.44}  &57.84 &68.56 \\
     \cline{2-2}
     &$k$-RNN  &47.38 &90.22 &96.09 &34.18 &83.16 &92.88 &56.26 &87.28 &94.54 &52.20 &85.80 &92.30&24.40 &56.40  &67.20\\
     \cline{2-2}
     &inv-DAKR & 65.02 & 94.98 & 98.00 & 48.73 & 89.18 & 94.97 & 70.32 & 92.54 & 97.20 & 67.20 & 90.30 & 95.60 &26.00 &58.00 &68.72\\
     \cline{2-2}
     &bi-DAKR & 68.98 & 95.82 & 98.62 & 50.66 & 90.19 & 95.51 & 71.87 & 93.24 & 97.70 & 68.80 & 90.50 & 95.80  &27.12 &\textbf{60.16} &\textbf{70.96} \\
     \cline{2-2}
     &$k$-INN\cite{Korn:ASR2000}+                    &64.71 &95.42 &97.96 &50.22 &89.34 &95.00 &70.87 &93.14 &97.45 &67.05 &90.65 &95.60&28.08 &58.64 &69.04\\
     \cline{2-2}
     &$k$-RNN+                    &68.67 &94.53 &97.60 &49.75 &88.32 &93.89 &68.82 &90.99 &95.94 &64.95 &88.35 &94.30&25.12 &58.80 &69.68\\
     \cline{2-2}
     &inv-DAKR+                   &68.13 &95.87 &98.53 &51.14 &89.78 &95.22 &73.23 &92.54 &96.45 &68.55 &90.95 &95.40&26.32 &57.52 &67.60\\
     \cline{2-2}
     &bi-DAKR+                    &\textbf{71.73} &\textbf{96.36} &\textbf{98.89} &\textbf{52.44} &\textbf{90.44} &\textbf{95.82} &\textbf{74.98} &\textbf{94.44} &\textbf{97.90} &\textbf{70.20} &\textbf{92.15} &\textbf{96.55}&26.96 &59.76 &70.32 \\\hline
     \hline
     \multirow{3}{*}{Fusion} & $k$-NN & 72.04  &95.96  &98.53 &53.26 & 90.95 & 95.73  & 71.87 & 92.64 & 96.80 & 68.05 & 90.15 & 94.95 &27.04 &59.36 &70.00 \\
     \cline{2-2}
     &$k$-INN\cite{Korn:ASR2000}  &55.38 &95.33 &97.96 &43.80 &89.78 &95.25 &53.55 &91.99 &97.50 &50.30 &89.65 &95.60 & 28.00  &58.96 &68.56 \\
     \cline{2-2}
     &$k$-RNN  &49.07 &91.38 &96.44 &37.44 &85.73 &93.73 &59.81 &90.34 &95.64 &56.10 &87.25 &93.35& 25.60 &57.12  &67.60 \\
     \cline{2-2}
     &inv-DAKR & 68.58  &96.00  &98.44 &52.53 & 90.57 & 95.89  & 73.53 & 94.24 & 98.15 & 70.65 & 92.10 & 96.25 &\textbf{28.16} &59.60 &69.84 \\
     \cline{2-2}
     &bi-DAKR &73.16  &97.02  &\textbf{99.11} &54.34 & 91.58 & 96.33  & 75.08 & 95.14 & 98.35 & 72.85 & 92.05 & 96.45 &28.00 &\textbf{61.52} &\textbf{71.36} \\
     \cline{2-2}
     &$k$-INN\cite{Korn:ASR2000}+                    &70.27 &96.22 &98.53 &52.94 &91.27 &95.73 &73.62 &94.35 &97.95 &72.40 &92.30 &96.40 &28.64  &59.44 &69.36 \\
     \cline{2-2}
     &$k$-RNN+                    &73.20 &95.82 &98.49 &53.32 &90.70 &95.41 &72.47 &92.84 &96.70 &68.85 &90.15 &94.90 &27.04 &60.48  &70.56 \\
     \cline{2-2}
     &inv-DAKR+                   &71.51 &96.76 &98.93 &53.70 &91.36 &96.01 &76.73 &94.44 &97.30 &72.35 &92.60 &96.35 &27.12 &59.36 &68.96\\
     \cline{2-2}
     &bi-DAKR+                    &\textbf{75.29} &\textbf{97.38} &99.07 &\textbf{55.89} &\textbf{91.93} &\textbf{96.87} &\textbf{78.48} &\textbf{95.79} &\textbf{98.40} &\textbf{75.30} &\textbf{93.40} &\textbf{97.00}&27.60 &61.44 &70.72  \\
     \cline{2-17}
     &SCA\cite{Bai:TIP2016}  &64.98 &91.91 &96.62 &44.94 &85.51 &93.96 &64.80 &83.35 &88.10 &67.40 &89.35 &94.75 & \textbf{*86.22} &*\textbf{87.27}  &\textbf{*88.36} \\
     \cline{2-17}
     &MRank$-L_{n}$\cite{Chen:ICIP13}  & 69.73 & 96.13 & 98.13 & 52.12 &91.58  &95.70 &73.47 &94.29 &97.60 &71.30 &91.50 &95.70 &27.76 &59.60 &69.04 \\
     \cline{2-17}
     &SSM\cite{Bai2017Scalable}  & 72.98 & 96.76 & \textbf{99.11} & 53.73 &91.49  &96.08 &76.63 &94.59 &97.95 &72.70 &92.40 &96.05&27.20 &61.12 &70.56 \\
     \cline{2-17}
     &Zhong's \cite{Zhong:CVPR2017} & 72.36 & 96.27 &98.71 &53.70  &91.65  &96.65 &73.42 &93.74 &97.29 &69.60 &91.50 &95.55&28.00 &60.40 &70.64 \\
     \hline
   \end{tabular}
   }
   }
\end{table*}

\section{Experimental Evaluations}
\label{sec:experiments}

To validate the efficiency and effectiveness of our proposals, we conduct experiments on six benchmark datasets: GRID~\cite{Chen:IJCV2010}, PRID450S~\cite{Roth:CVPR2014}, VIPeR~\cite{Gray:ECCV2008}, CUHK03~\cite{Li:CVPR2014DeepReID}, Market-1501~\cite{Zheng:ICCV2016} and Mars~\cite{Zheng:ECCV2016MARS}. {\mcb All the experiments are performed with MATLAB2015 on a server equipped with an Intel Xeon CPU E5-2630 and 64GB memory.}


\subsection{Experimental Protocol}
In our experiments, we use the standard protocol to split the data. The matching accuracy at 
different ranks 
on the datasets GRID, PRID450S, VIPeR, and CUHK03 is averaged over 10 trials. 
 As baselines, we consider the $k$-NN, 
$k$-INN \cite{Korn:ASR2000}, $k$-RNN, SSM \cite{Bai2017Scalable}, MRank-$L_{n}$~\cite{Chen:ICIP13}, SCA~\cite{Bai:TIP2016}, Zhong et al. \cite{Zhong:CVPR2017} and the reranking method proposed in \cite{Yu:BMVC2017}.
In addition, we use the symbol ``$+$''  to indicate the approaches that use extra probe samples, including $k$-INN+, 
$k$-RNN+, inv-DAKR+ and bi-DAKR+. In detail, we 
implement $k$-INN~\cite{Korn:ASR2000}, $k$-RNN and SCA~\cite{Bai:TIP2016} by ourselves with the optimal $k$ for each result, and we directly cite 
the results of SSM \cite{Bai2017Scalable} from its paper. For Zhong et al.~\cite{Zhong:CVPR2017}, we obtain each result using the optimal hyperparameter from the released code, and for \cite{Chen:ICIP13}, we implement 
MRank-$L_{n}$ with the optimal hyperparameter setting reported by \cite{Chen:ICIP13} and \cite{Zelnik:NIPS2005}.

Note that reranking can be viewed as a postprocessing step for person ReID. Our proposed reranking approaches inv-DAKR and bi-DAKR can be inserted into any person ReID pipeline. Therefore, we evaluate the proposed inv-DAKR and bi-DAKR on the six datasets with different combinations of feature extraction and metric learning methods.

For GRID, PRID450S, VIPeR, and CUHK03, we use LOMO~\cite{Liao:CVPR2015} and GOG~\cite{Matsukawa:CVPR2016} features, whereas for Market-1501 and Mars, we use IDE features~\cite{Zheng:ICCV2016, Zheng:ECCV2016MARS}. For GRID, we also use ELF6 features \cite{Liu:ECCV2012}. Moreover, we conduct experiments based on concatenation of LOMO with GOG features, which we call \emph{Fusion}. Specifically, we obtain the LOMO features of each dataset with 26,960 dimensions by the code provided by~\cite{Liao:CVPR2015} with the default settings and use the code provided by~\cite{Matsukawa:CVPR2016} to extract 4 types of GOG features (RGB, Lab, HSV and nRnG), and we concatenate them into one feature with 27,622 dimensions for use in our experiments. Another fusion features is introduced for GRID by equally concatenating \emph{Fusion} features with the unit $\ell_2$-norm ELF6 features, which we call \emph{FusionAll}. On the other hand, for metric learning, XQDA is implemented for PRID450S, VIPeR, and CUHK03, as shown in Table~\ref{tab:PRID450s-VIPeR-CUHK03}, and the Euclidean distance ($\ell_2$-norm) is added for GRID, whereas both the Mahalanobis distance, and KISSME \cite{Kostinger:CVPR2012} are added for Market-1501 and Mars.

In inv-DAKR and bi-DAKR, the density-adaptive parameter $\sigma_j$ depends on a preset parameter $k$, which changes from dataset to dataset. For each dataset, we report the results with an optimal $k$. In later subsections, we show the curves of the average performance gain achieved when using the proposed reranking methods with respect to the parameter $k$ and report an empirical rule to set a proper $k$.

To make the evaluation more systematic and clearer, we divide the six benchmark datasets into three groups:
\begin{itemize}
\item Datasets with perfect single-shot matching: PRID450s, VIPeR, and CUHK03.

\item A dataset with imperfect single-shot matching: GRID.

\item Datasets with multiple-shot matching: Market-1501 and Mars.

\end{itemize}

\subsection{Experiments on Datasets with Perfect Single-Shot Matching: PRID450s, VIPeR, and CUHK03}
\label{sec:experiments-perfect-single-shot}

\subsubsection{Dataset Descriptions}

Both the PRID450s and VIPeR datasets contain images of people walking captured by two disjoint cameras and 
mainly suffer from variations in viewpoints, illuminations and poses. PRID450s includes a total of 450 images of 225 pedestrians, and VIPeR includes 316 pairs of images of 316 pedestrians. In the experiments, half of the image pairs in both datasets are used for training, and half of the image pairs are used for testing.

In contrast to PRID450s and VIPeR, CUHK03 includes 13,164 images of 1,360 people walking captured by six disjoint cameras.
CUHK03 suffers from misalignments, occlusions and  missing body parts; thus, it is closer to real surveillance scenarios. 
Except for the manually cropped pedestrian images, it also includes pedestrian images obtained from state-of-the-art pedestrian detectors. In the experiments, 200 images of 100 people are used for testing, and approximately 13,000 images of 1260 people are used for training.

We refer to these three datasets as \emph{perfect single-shot matching} because in the testing phase, 
the samples in the gallery set 
have one-to-one correspondences to the samples in the probe set.

\subsubsection{Experimental Results}

\begin{table}
  \scriptsize
  \caption{Comparison on GRID with different features. }
  \label{tab:GRID}
  \vspace{4pt}
  \centering
  \begin{tabular}{c|c|c|c|c|c}
  \hline
  \textbf{Feature}  &\textbf{Metric} &\textbf{Methods} &\textbf{r=1} &\textbf{r=10} &\textbf{r=20}\\\hline
  \hline
  \multirow{5}{*}{ELF6} &\multirow{5}{*}{Euc} &$k$-NN &4.88 &20.32 &26.24\\
   & &$k$-INN\cite{Korn:ASR2000}  &7.20 &21.28 &31.60\\
   & &$k$-RNN  &6.72 &19.68  &26.20 \\
   & &inv-DAKR &8.32 &\textbf{22.80} &31.20\\
   & &bi-DAKR &6.32 &22.32 &28.80\\
   & &$k$-INN\cite{Korn:ASR2000}+  &6.96 &22.56 &\textbf{31.28}\\
   & &$k$-RNN+  &5.28 &21.36  &30.24 \\
   & &inv-DAKR+ &\textbf{8.80} &22.64 &30.4\\
   & &bi-DAKR+ &6.88 &\textbf{22.80} &28.88\\\hline
  \multirow{5}{*}{ELF6} &\multirow{5}{*}{XQDA} &$k$-NN &8.64 &30.48 &44.32\\
  &  &$k$-INN\cite{Korn:ASR2000}  &\textbf{13.60} &38.96 &50.88\\
  &  &$k$-RNN  &10.96  &37.44  &48.72 \\
  &  &inv-DAKR &13.36 &40.16 &52.08\\
  &  &bi-DAKR &11.28 &39.52 &\textbf{52.40}\\
   &  &$k$-INN\cite{Korn:ASR2000}+  &13.52 &39.12 &50.48\\
   &  &$k$-RNN+  &9.92  &34.16  &46.96 \\
   & &inv-DAKR+ &13.28 &\textbf{40.48} &52.16\\
   & &bi-DAKR+ &11.36 &39.44 &52.24\\\hline
  \multirow{5}{*}{LOMO} &\multirow{5}{*}{Euc} &$k$-NN &15.20 &30.80 &36.40\\
   & &$k$-INN\cite{Korn:ASR2000}  &14.64  &35.44 &42.64\\
   & &$k$-RNN  &13.36  &30.80  &38.16 \\
   & &inv-DAKR &16.00 &35.44 &43.76\\
   & &bi-DAKR &\textbf{17.84} &34.16 &41.44\\
   & &$k$-INN\cite{Korn:ASR2000}+  &14.64  &\textbf{35.76} &43.76\\
   & &$k$-RNN+  &16.56  &32.16  &38.56 \\
   & &inv-DAKR+ &14.80 &35.68 &\textbf{44.08}\\
   & &bi-DAKR+ &17.04 &34.24 &41.28\\\hline
  \multirow{5}{*}{GOG} &\multirow{5}{*}{Euc} &$k$-NN &13.28 &33.76 &44.40\\
   & &$k$-INN\cite{Korn:ASR2000}  &15.44  &34.88 &42.40\\
   & &$k$-RNN  &12.80 &30.72  &39.36 \\
   & &inv-DAKR &\textbf{16.56} &34.96 &43.20\\
   & &bi-DAKR &16.00 &\textbf{36.40} &\textbf{44.96}\\
   & &$k$-INN\cite{Korn:ASR2000}+  &15.60  &35.28 &42.96\\
   & &$k$-RNN+  &14.72 &34.40  &44.08 \\
   & &inv-DAKR+ &14.88 &34.96 &41.28\\
   & &bi-DAKR+ &15.44 &36.32 &44.72\\\hline
  \multirow{5}{*}{Fusion} &\multirow{5}{*}{Euc} &$k$-NN &14.72 &35.44 &45.84\\
   & &$k$-INN\cite{Korn:ASR2000}  &16.00  &35.76 &43.92\\
   & &$k$-RNN  &14.64 &32.56  &41.02 \\
   & &inv-DAKR &18.16 &37.04 &\textbf{46.00}\\
   & &bi-DAKR &\textbf{18.64} &37.68 &\textbf{46.00}\\
   & &$k$-INN\cite{Korn:ASR2000}+  &16.88  &35.52 &45.36\\
   & &$k$-RNN+  &15.92 &36.48  &45.60 \\
   & &inv-DAKR+ &17.44 &37.44 &43.44\\
   & &bi-DAKR+ &18.56 &\textbf{37.92} &45.52\\\hline
  \multirow{5}{*}{FusionAll} &\multirow{5}{*}{Euc} &$k$-NN &14.80 &35.60 &\textbf{46.24}\\
   & &$k$-INN\cite{Korn:ASR2000}  &15.04  &35.76 &43.44\\
   & &$k$-RNN  &13.76 &31.92  &39.68 \\
   & &inv-DAKR &17.68 &36.08 &45.12\\
   & &bi-DAKR &\textbf{17.76} &37.20 &46.08\\
   & &$k$-INN\cite{Korn:ASR2000}+  &16.40  &36.32 &44.32\\
   & &$k$-RNN+  &16.24 &36.64  &45.12 \\
   & &inv-DAKR+ &18.40 &37.28 &45.60\\
   & &bi-DAKR+ &17.68 &\textbf{37.68} &45.76\\\hline
  \multirow{5}{*}{FusionAll} &\multirow{5}{*}{XQDA} &$k$-NN &27.20 &61.12 &71.20\\
   & &$k$-INN\cite{Korn:ASR2000}  &28.56  &59.92 &70.00\\
   & &$k$-RNN  &26.08  &57.84 &69.20 \\
   & &inv-DAKR&\textbf{28.88} &60.40 &70.88\\
   & &bi-DAKR &28.08 &62.40 &72.08\\
   & &$k$-INN\cite{Korn:ASR2000}+  &29.52  &60.64 &70.48\\
   & &$k$-RNN+  &27.44  &61.84 &71.84 \\
   & &inv-DAKR+ &28.80 &60.96 &70.88\\
   & &bi-DAKR+ &28.24 &\textbf{62.56} &\textbf{72.24}\\
   \cline{3-6}
   & &MRank$-L_{n}$\cite{Chen:ICIP13} & 27.36 & 59.76 &69.92 \\
   & &SSM\cite{Bai2017Scalable}  & 27.60 & \textbf{62.56} & 71.60 \\
   & &Zhong's\cite{Zhong:CVPR2017} & 28.24 & 61.60 &71.92 \\
   \hline

  \end{tabular}
\end{table}

We show the experimental results in Table~\ref{tab:PRID450s-VIPeR-CUHK03}. The proposed inv-DAKR and bi-DAKR show notable improvements compared to the baseline results of $k$-NN.

If probe samples are available, we can use them to further improve the reranking results. Both inv-DAKR+ and bi-DAKR+ yield further improvements over inv-DAKR and bi-DAKR. Compared to inv-DAKR, the improvements of bi-DAKR+ are more notable. On average, inv-DAKR+ improved over inv-DAKR around $3\%$ and bi-DAKR+ improved over bi-DAKR around $5\%$.

Note that context information from probe samples is also used in the state-of-the-art reranking approaches used in SCA~\cite{Bai:TIP2016} and Zhong's method~\cite{Zhong:CVPR2017} to find reciprocal samples. 
While more sophisticated approaches are involved in~Zhong's method~\cite{Zhong:CVPR2017}, SSM~\cite{Bai2017Scalable} and MRank-$L_{n}$~\cite{Chen:ICIP13}, our simple proposals still yield superior or comparable results. In a later subsection, we will show that our proposed approaches are much less computationally expensive.

\subsubsection{Analysis and Discussion} To give a comprehensive understanding of the experimental results, we calculate the average performance gain with respect to the $k$-NN baseline as a function of the parameter $k$ 
in Fig.~\ref{fig:increment_inv_one_2}, \ref{fig:increment_bi_one_2}, \ref{fig:increment_inv_set_2}, and \ref{fig:increment_bi_set_2}.

On these three datasets, the performance of bi-DAKR is more promising than inv-DAKR. This suggests that integrating the reidentification information in both a direct manner and an inverse manner is quite useful. In contrast, we observe that both $k$-INN~\cite{Korn:ASR2000} and $k$-RNN yield inferior results, which are even worse than the $k$-NN baselines. This is because the feature space of these experiments is of very high dimensions (approximately 27,000 to 54,000) and because the size of the testing sets is only approximately 1,000 samples; thus the distribution of samples is highly sparse. As shown in \cite{Qin:CVPR2011}, this sparsity accounts for the fast degradation of the similarity function and leads to the degenerated performance of the $k$-INN~\cite{Korn:ASR2000} and $k$-RNN methods. At the same time, these results confirm that using a smooth kernel function with a local density-adaptive parameter to accommodate the ambiguity into the reranking list can fix this problem and achieve performance improvements.

If the probe samples are used, we observed that both $k$-INN+\cite{Korn:ASR2000} and $k$-RNN+ perform better than the $k$-NN method. In addition, as their smoothed versions,  inv-DAKR+ and bi-DAKR+ also show improved performance;  not only do the average performance gains increase on average, but the variances are also reduced, as shown in Fig.~\ref{fig:increment_inv_one_2}, \ref{fig:increment_bi_one_2}, \ref{fig:increment_inv_set_2}, and \ref{fig:increment_bi_set_2}. The performance improvements are particularly significant when using a smaller $k$, as seen at the  beginnings of the curves.
The reason is that each identity only has one sample in the original gallery set, which means that the local neighborhoods are not well constructed, while after adding probe samples without any supervision information, there are two samples of each identity,  allowing the local neighborhoods to be more suitably described and significantly enhancing performance. These results suggest that in this \emph{perfect single-shot matching scenario}, adding the available probe samples to the gallery set helps to improve the reranking results.

\begin{figure*}[ht]
\vspace{-2pt}
\centering
\small
\subfigure[rank-1]{\includegraphics[clip=true,trim=12 5 15 0,width=0.245\columnwidth]{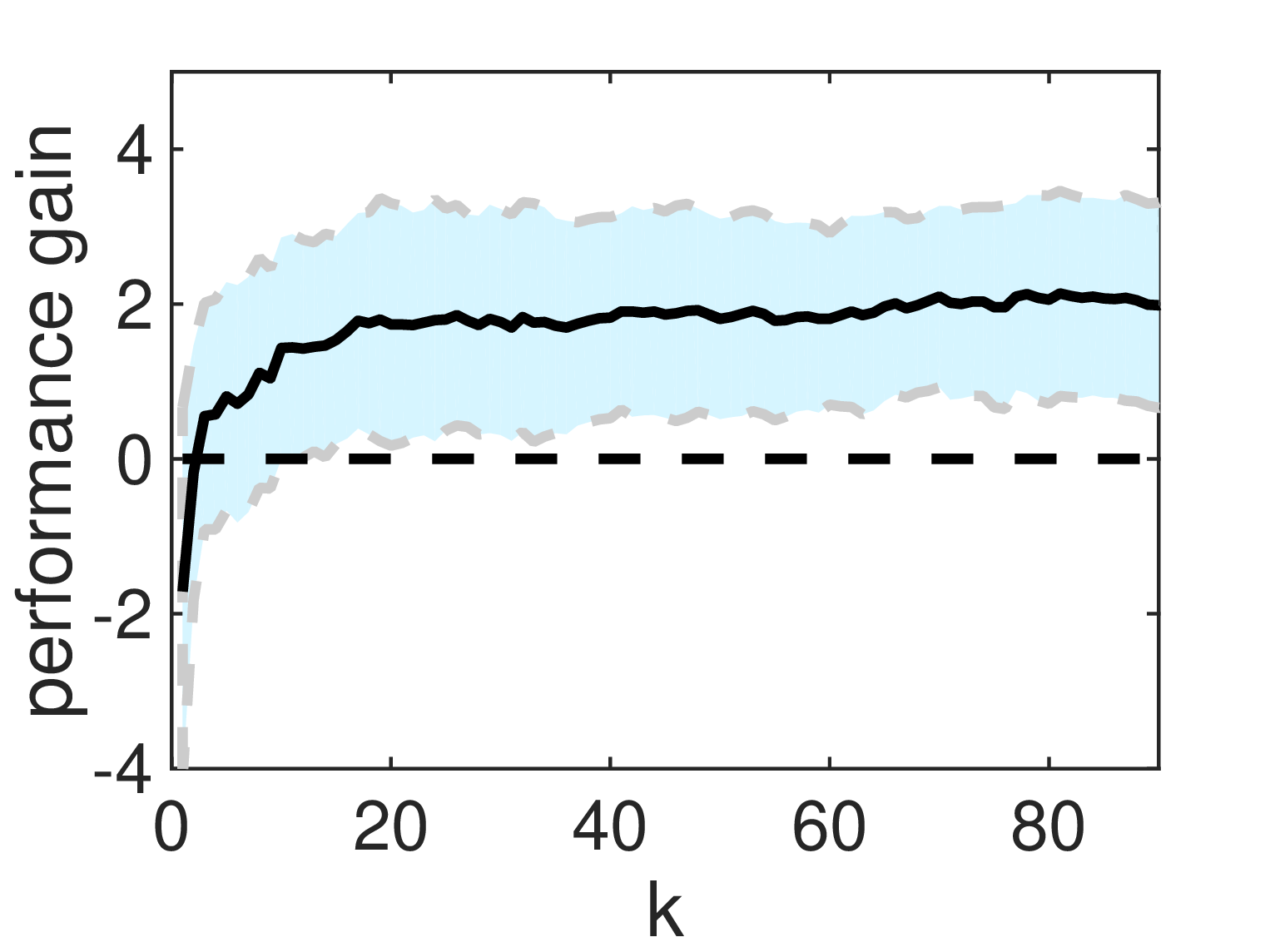}}
\subfigure[rank-5]{\includegraphics[clip=true,trim=12 5 15 0,width=0.245\columnwidth]{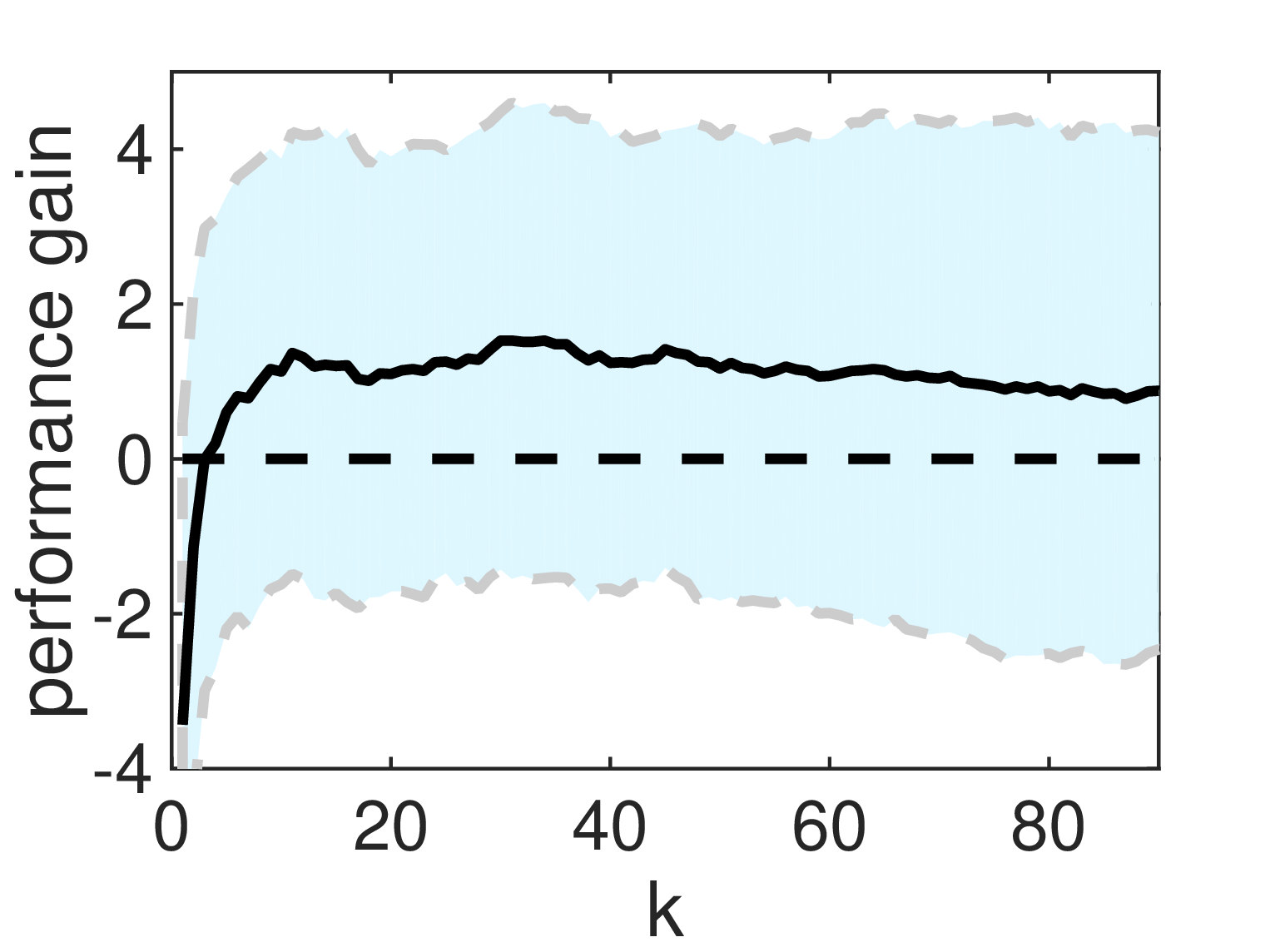}}
\subfigure[rank-10]{\includegraphics[clip=true,trim=12 5 15 0,width=0.245\columnwidth]{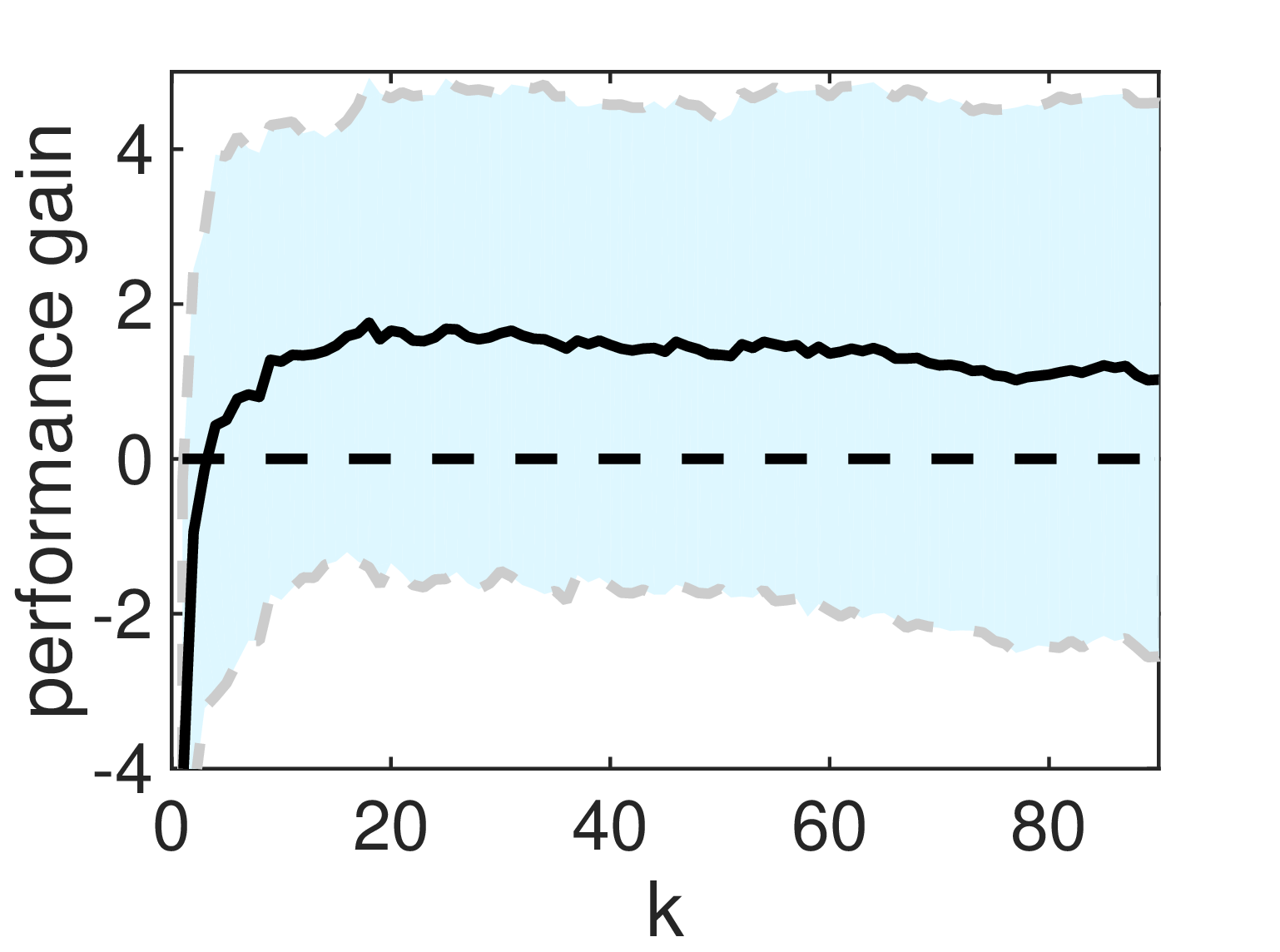}}
\subfigure[rank-20]{\includegraphics[clip=true,trim=12 5 15 0,width=0.245\columnwidth]{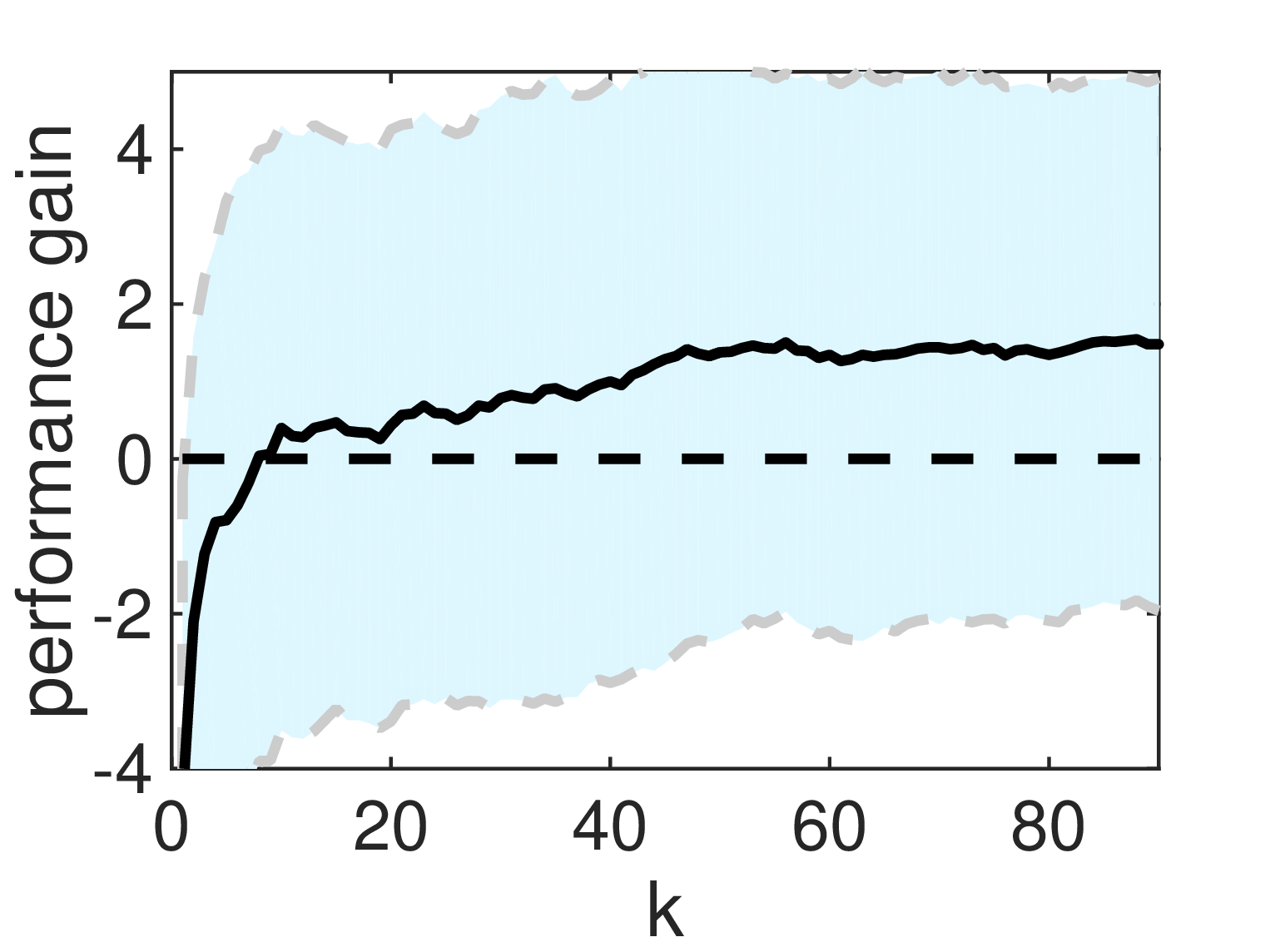}}
\caption{ Average performance gain of inv-DAKR as a function of $k$ in an imperfect one-to-one matching scenario.} 
\label{fig:increment_inv_one_1}
\vspace{-1mm}
\end{figure*}

\begin{figure*}[ht]
\vspace{-2pt}
\centering
\small
\subfigure[rank-1]{\includegraphics[clip=true,trim=12 5 15 0,width=0.245\columnwidth]{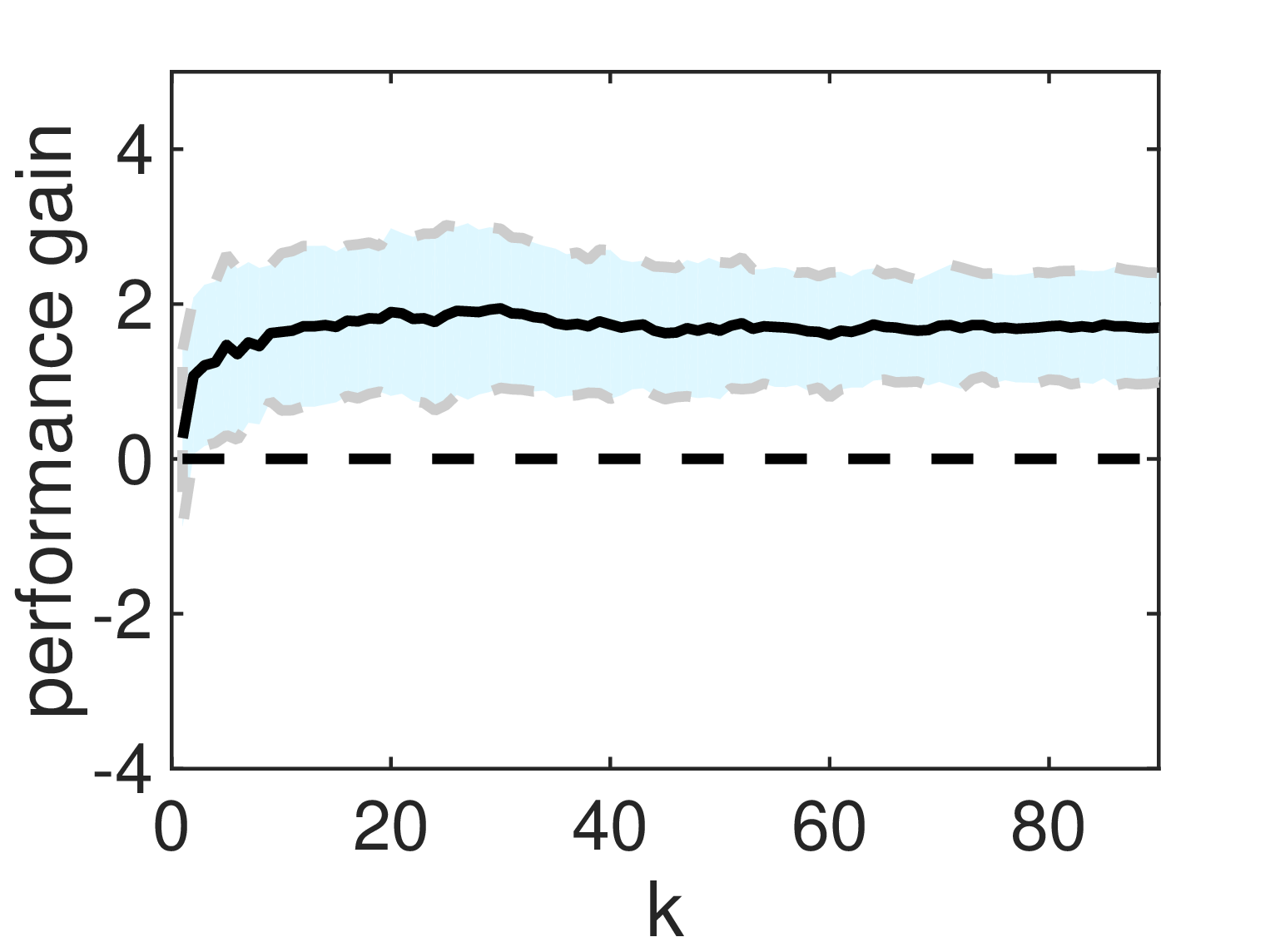}}
\subfigure[rank-5]{\includegraphics[clip=true,trim=12 5 15 0,width=0.245\columnwidth]{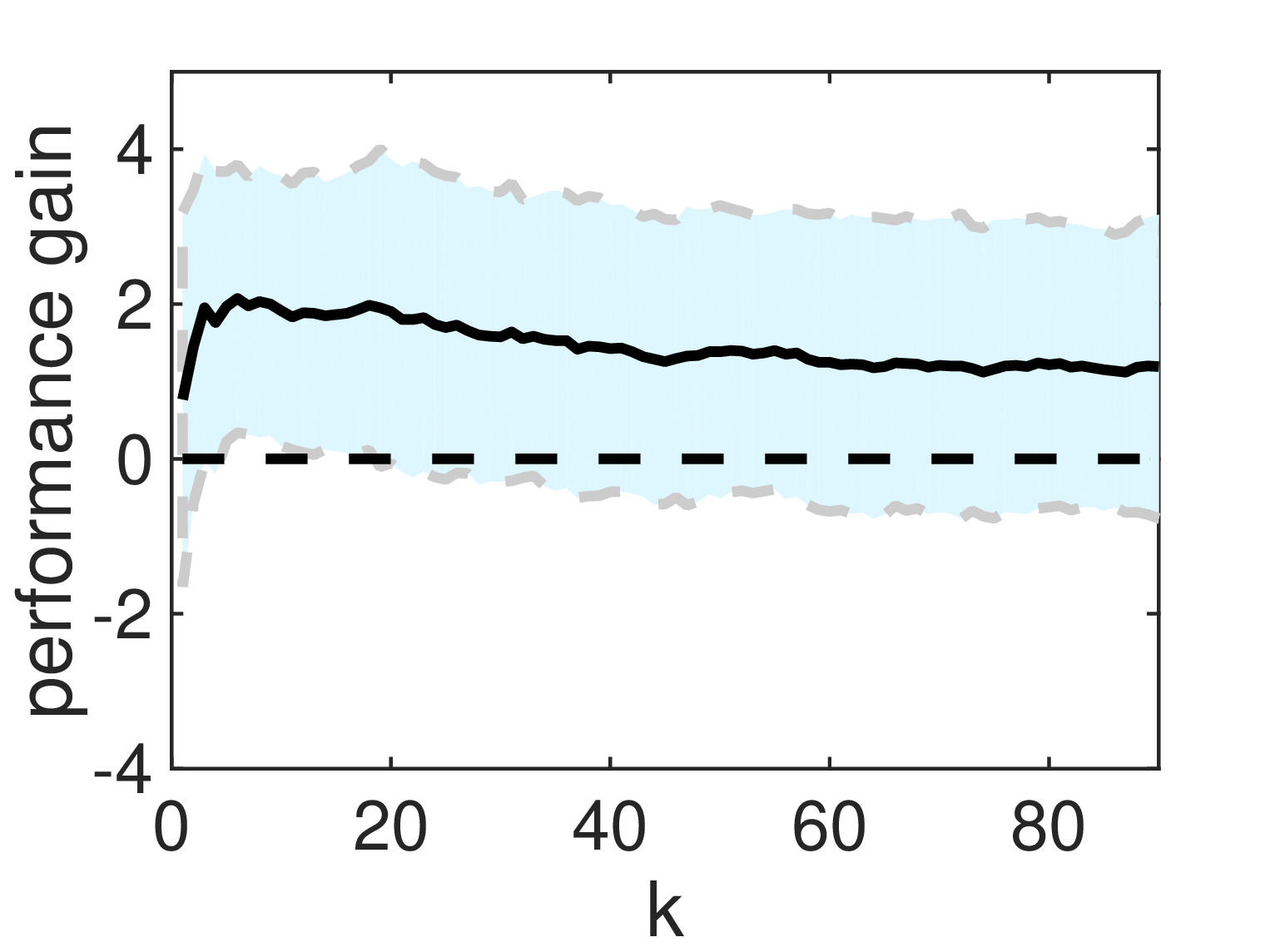}}
\subfigure[rank-10]{\includegraphics[clip=true,trim=12 5 15 0,width=0.245\columnwidth]{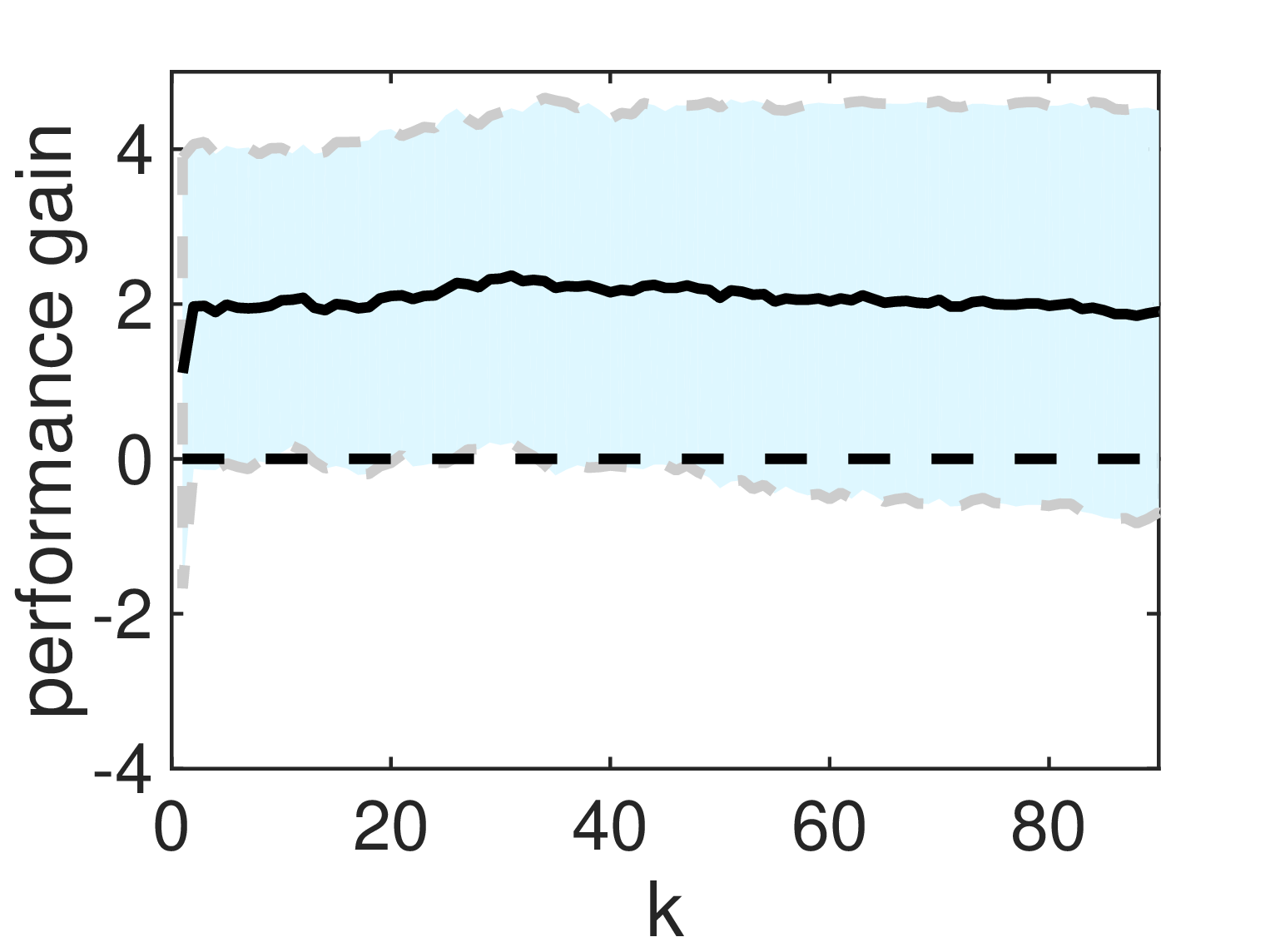}}
\subfigure[rank-20]{\includegraphics[clip=true,trim=12 5 15 0,width=0.245\columnwidth]{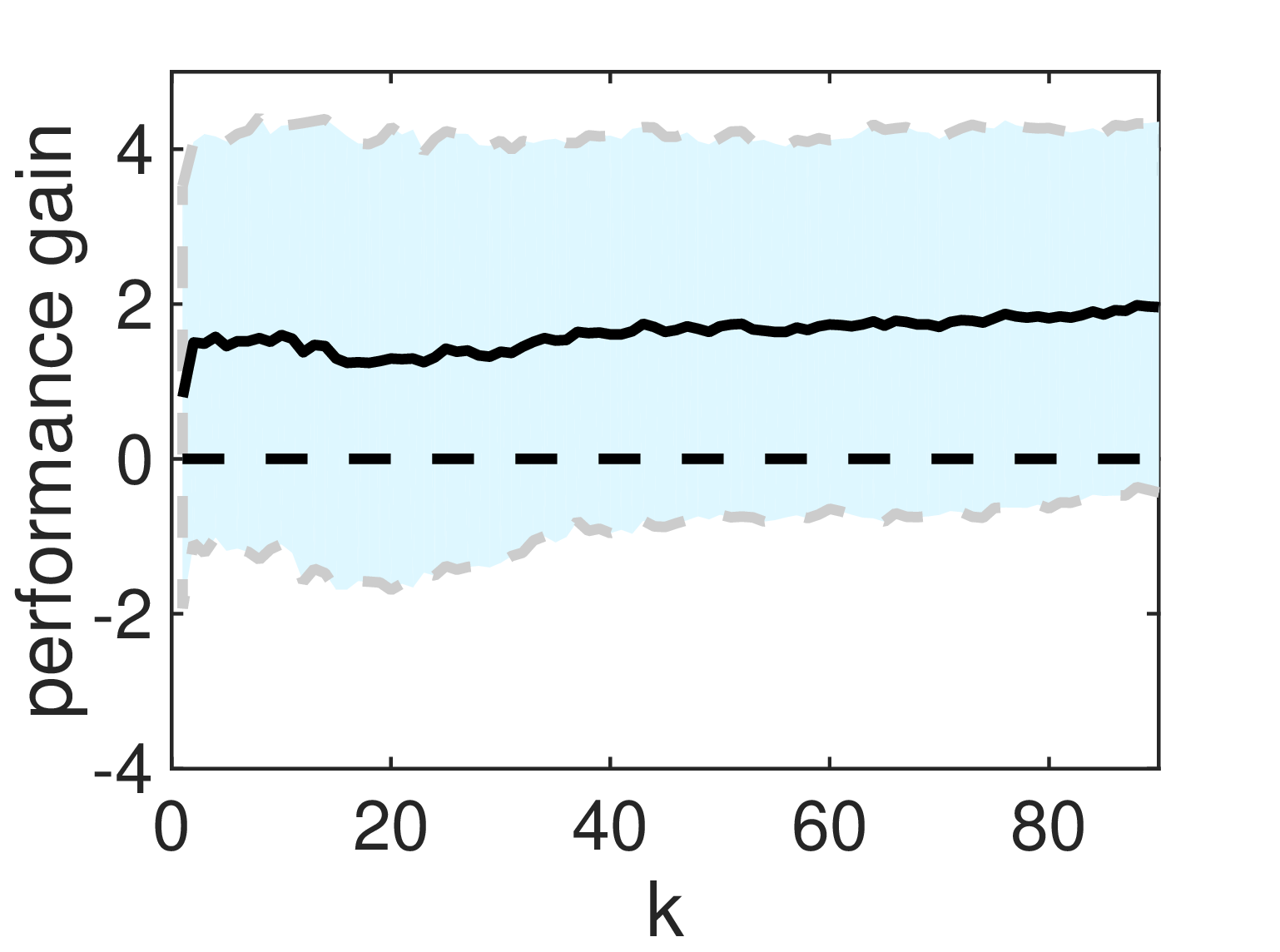}}
\caption{ Average performance gain 
of bi-DAKR as a function of $k$ in an imperfect one-to-one matching scenario.}
\label{fig:increment_bi_one_1}
\vspace{-1mm}
\end{figure*}

\begin{figure*}[htbp]
\vspace{-2pt}
\centering
\small
\subfigure[rank-1]{\includegraphics[clip=true,trim=12 5 15 0,width=0.245\columnwidth]{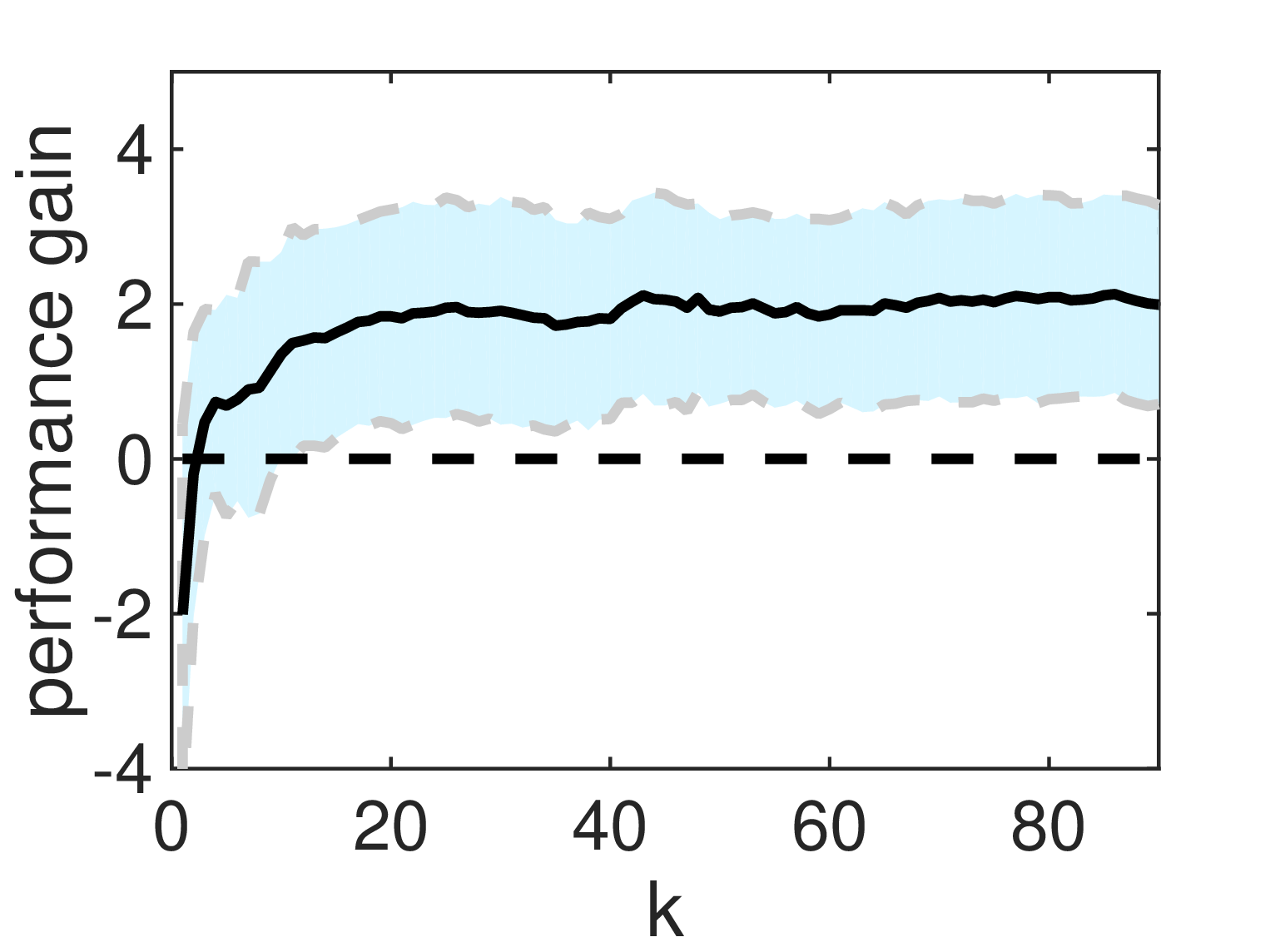}}
\subfigure[rank-5]{\includegraphics[clip=true,trim=12 5 15 0,width=0.245\columnwidth]{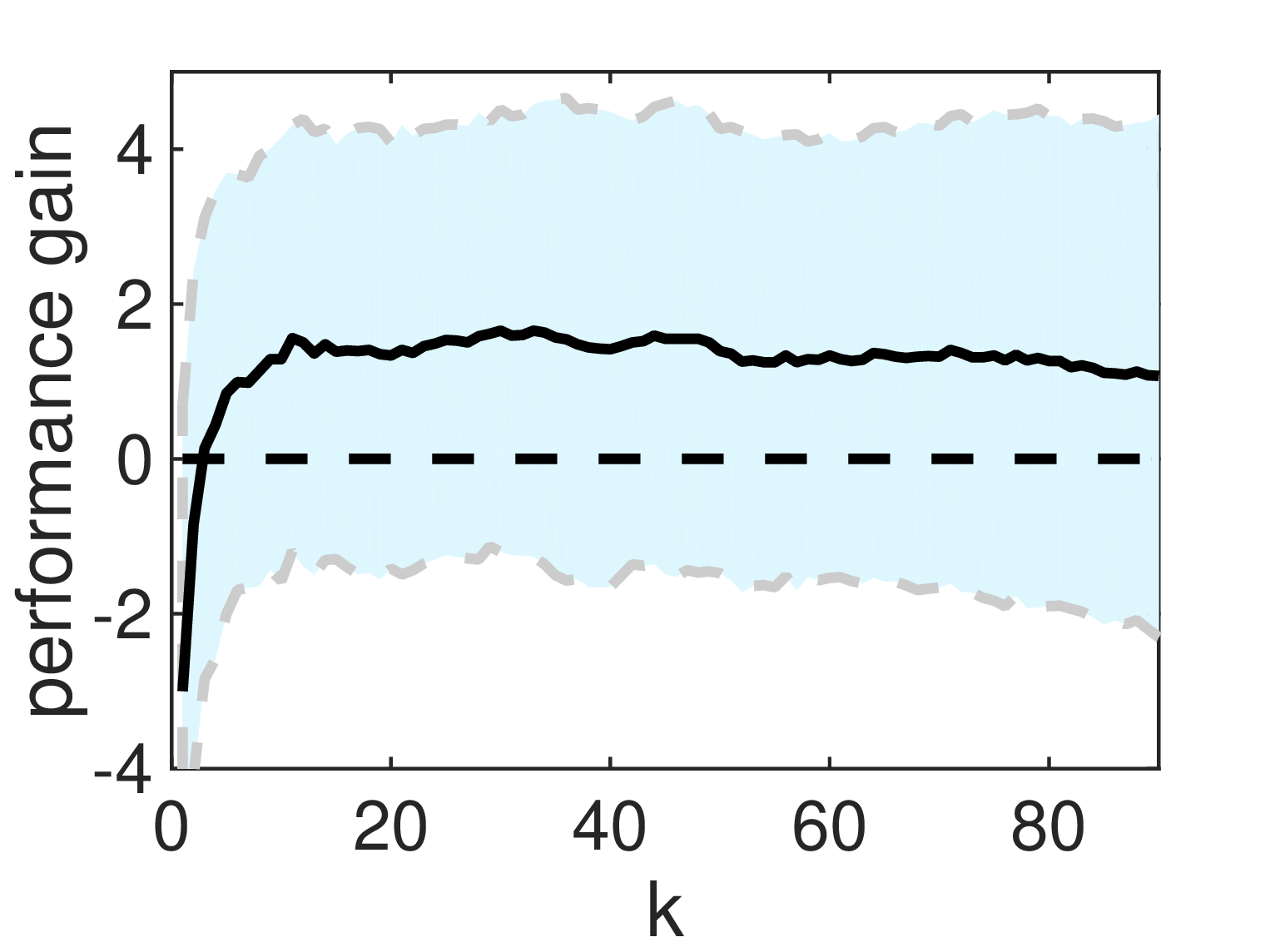}}
\subfigure[rank-10]{\includegraphics[clip=true,trim=12 5 15 0,width=0.245\columnwidth]{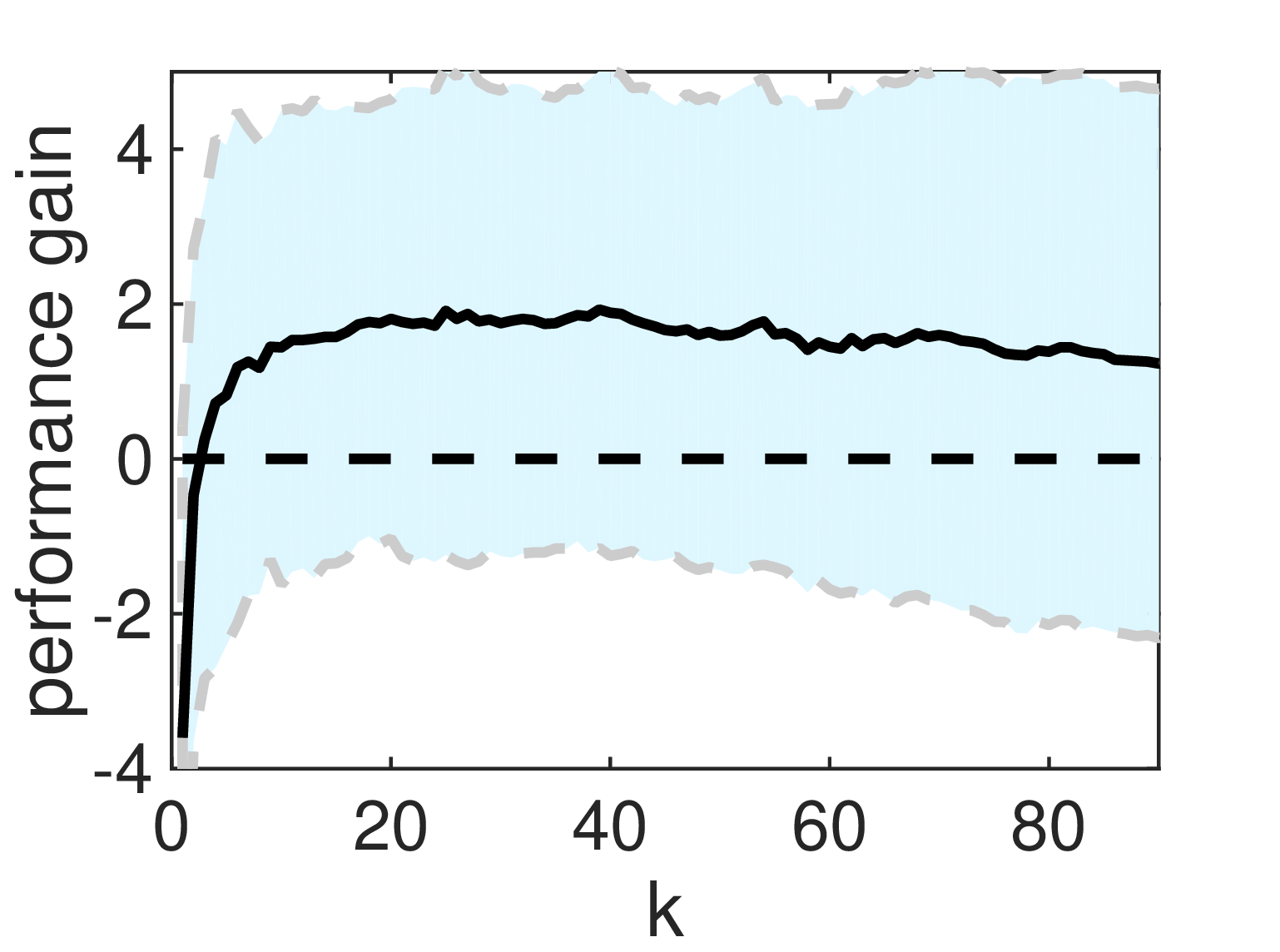}}
\subfigure[rank-20]{\includegraphics[clip=true,trim=12 5 15 0,width=0.245\columnwidth]{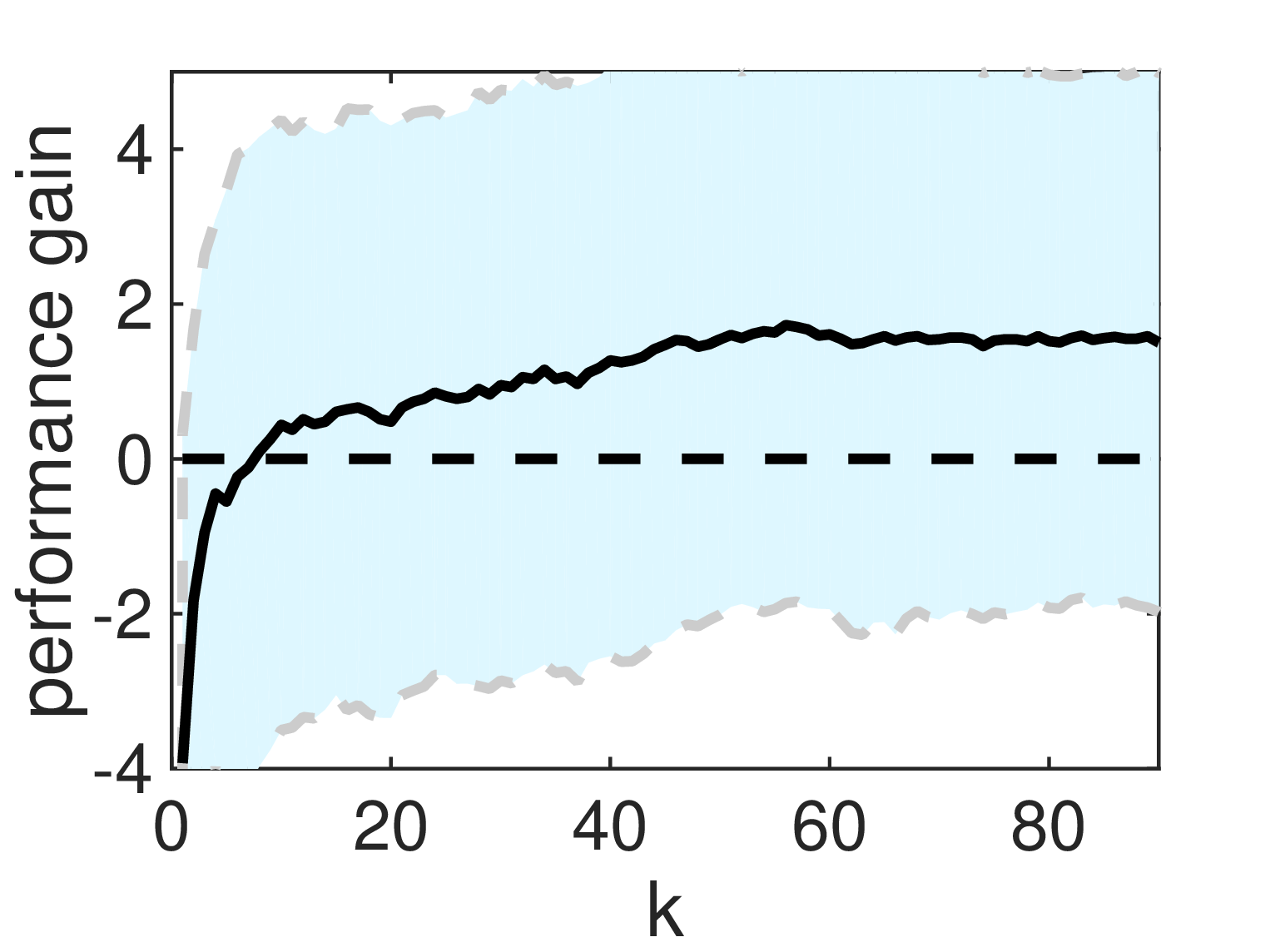}}
\caption{ Average performance gain 
of inv-DAKR+ as a function of $k$ in an imperfect one-to-one matching scenario.}
\label{fig:increment_inv_set_1}
\vspace{-1mm}
\end{figure*}

\begin{figure*}[htbp]
\vspace{-2pt}
\centering
\small
\subfigure[rank-1]{\includegraphics[clip=true,trim=12 5 15 0,width=0.245\columnwidth]{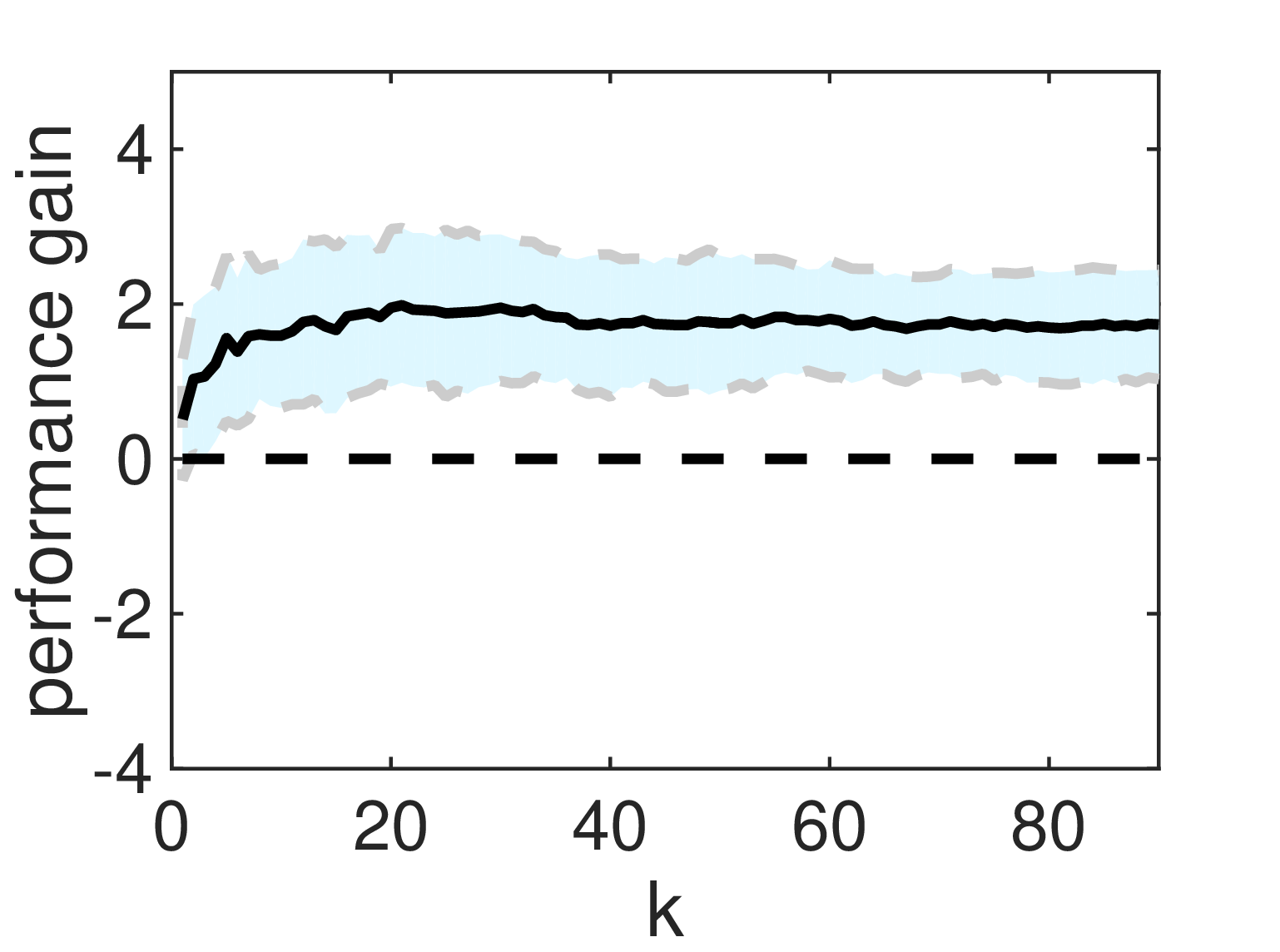}}
\subfigure[rank-5]{\includegraphics[clip=true,trim=12 5 15 0,width=0.245\columnwidth]{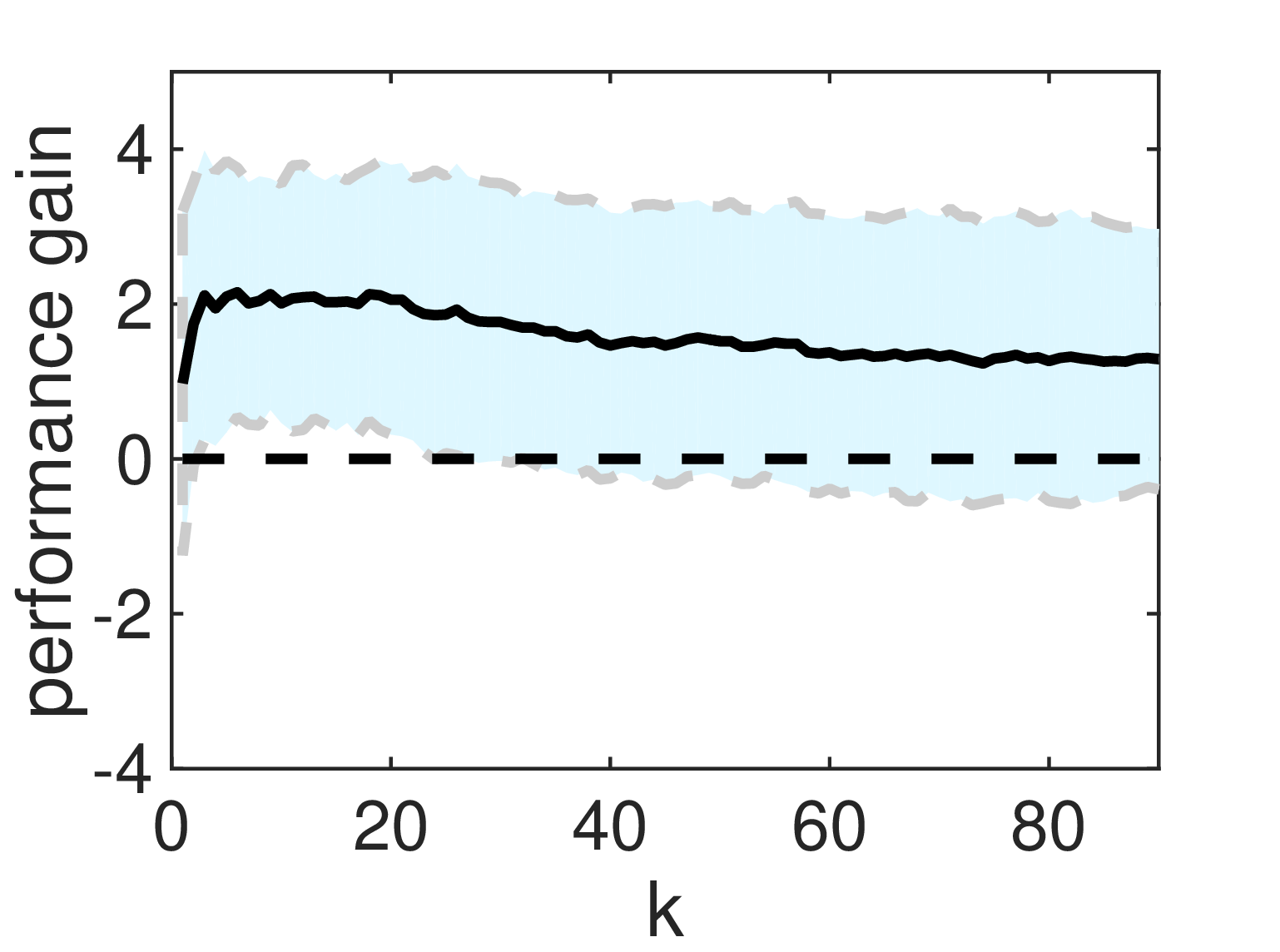}}
\subfigure[rank-10]{\includegraphics[clip=true,trim=12 5 15 0,width=0.245\columnwidth]{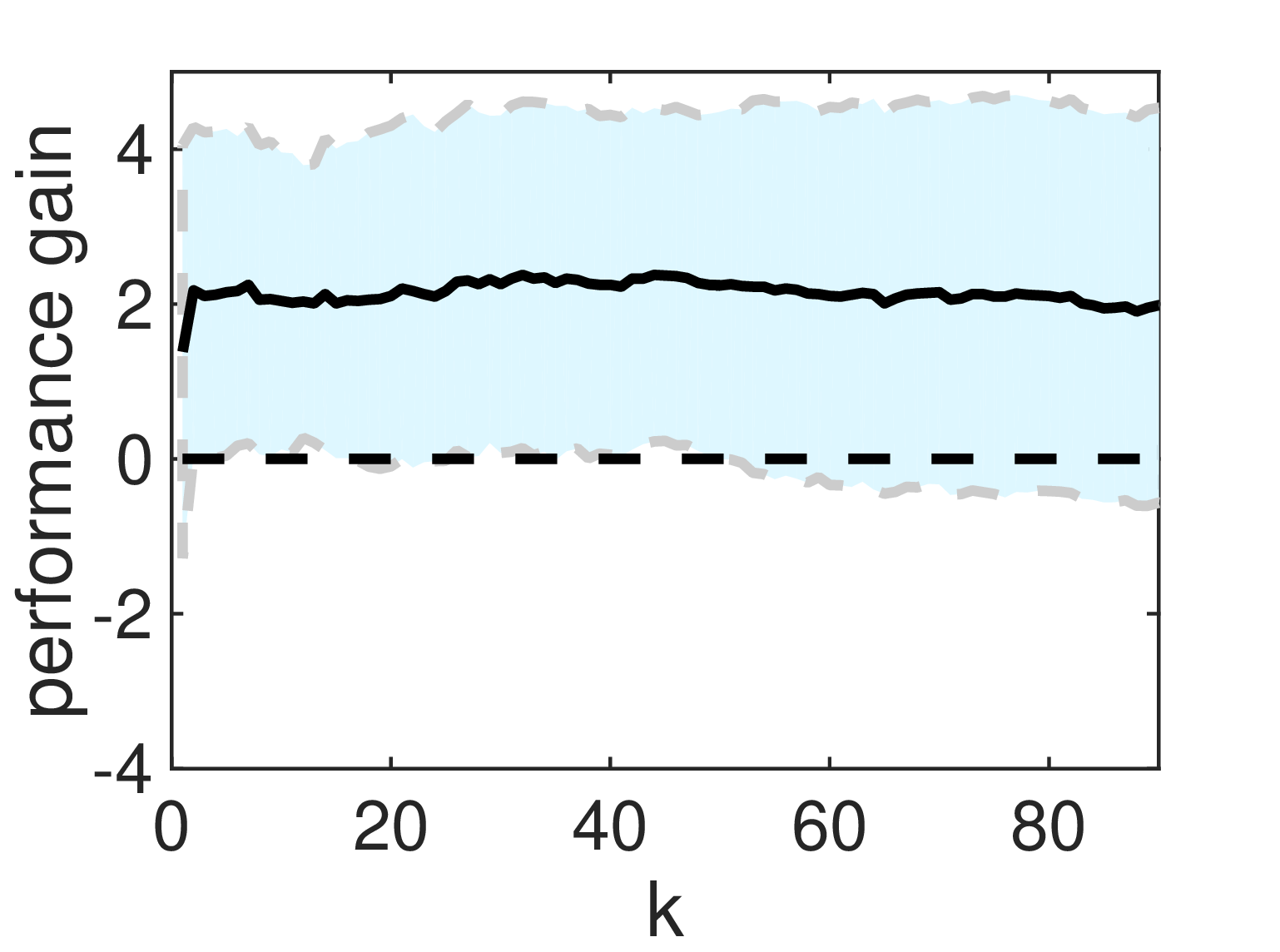}}
\subfigure[rank-20]{\includegraphics[clip=true,trim=12 5 15 0,width=0.245\columnwidth]{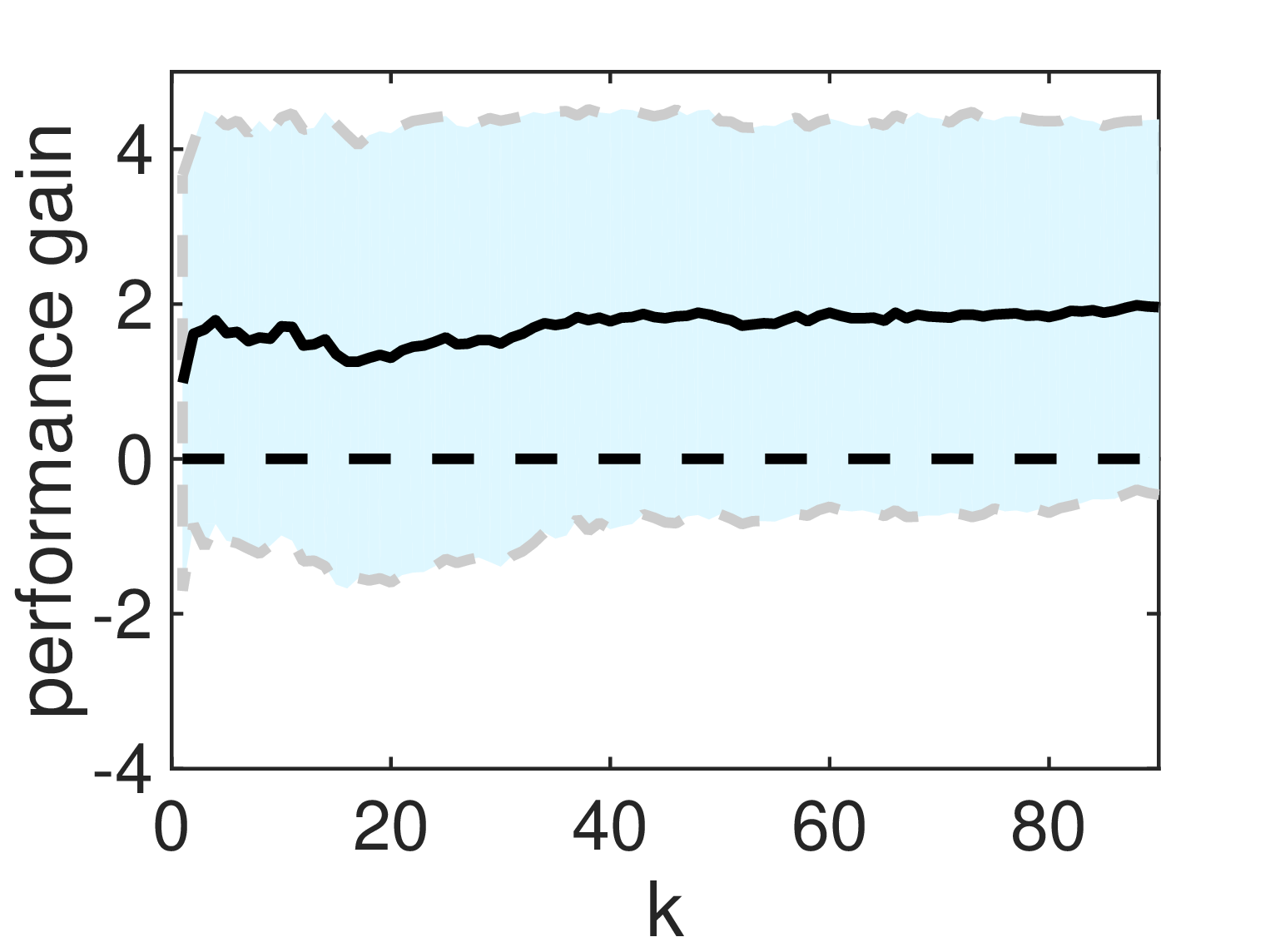}}
\caption{ Average performance gain 
of bi-DAKR+ as a function of $k$ in an imperfect one-to-one matching scenario.}
\label{fig:increment_bi_set_1}
\vspace{-1mm}
\end{figure*}

Moreover, in Table~{\ref{tab:PRID450s-VIPeR-CUHK03}}, from the LOMO features to the \emph{Fusion} features for all three datasets, we can observe the performance gain achieved when using the probe set. This is because the more discriminative and robust the features are, the better the feature space that it will build for the local context detecting method to work. Since GOG features are more robust than LOMO features and weaker than \emph{Fusion} features, which are obtained by combining with LOMO features, the performance gain comes from the robustness of the feature. Compared to VIPeR and PRID450s, it is more obvious that $k$-INN\cite{Korn:ASR2000}+ and $k$-RNN+ improve more notably 
in CUHK03.

\subsection{Experiments on a Dataset with Imperfect Single-Shot matching: GRID}
\label{sec:experiments-imperfect-single-shot}

\subsubsection{Dataset Description} GRID is a basic dataset for the person reidentification task that contains 250 pedestrian image pairs. Accordingly, 125 image pairs are used for training, and 125 image pairs are used for testing. In addition, there are 775 images in the gallery set; however, these 775 images do not match any person in the probe set. Therefore, we call this an \emph{imperfect single-shot matching} scenario because the gallery set does not contain one-to-one matches to the probe set. %

\subsubsection{Experimental Results and Discussion}
We list the experimental results in the last few columns of Table~\ref{tab:PRID450s-VIPeR-CUHK03} and in Table~\ref{tab:GRID}. Although our proposed inv-DAKR and bi-DAKR still show notable performance improvements compared to $k$-NN, there are two differences: a) inv-DAKR can compete and even outperform bi-DAKR, and b) the performance difference between using the probe set and not using the probe set is minor. 
Note that in addition to the 125 pairs of probe samples, the gallery set also includes 775 irrelevant images, which do not match any of the probe samples. When these 775 irrelevant images are added, they cannot provide any useful information to 
improve the local neighborhoods of the gallery samples. Thus, the performance improvements of $k$-INN+, $k$-RNN+, inv-DAKR+, and bi-DAKR+ over their counterpart methods are relatively minor.

To better understand the experimental results, we calculate the curves of the average performance gains with respect to the corresponding result of $k$-NN and show them as functions of the parameter $k$ 
in Fig.~\ref{fig:increment_inv_one_1}, \ref{fig:increment_inv_set_1}, \ref{fig:increment_bi_one_1} and \ref{fig:increment_bi_set_1}. The performance gain of using the probe set over not using the probe set is minor. This is because, while the 125 pairs of probe samples can provide some useful information, they represent only a small proportion (\textit{i.e.}, approximately $25\%$). The addition of the 775 irrelevant images will dilute the true distribution and thus eliminate the performance improvements. 
In addition, in GRID, there is an observation that the performance of $k$-INN \cite{Korn:ASR2000} and $k$-RNN is better than the $k$-NN baseline under certain settings and much higher than $k$-NN with the ELF6 features. This is because the testing set includes 1,025 samples, which is more than that of VIPeR, PRID450s and CUHK03; thus, the distribution sparsity is alleviated. In particular, the dimensionality of the ELF6 features is just 2786, which is much lower than the other features that we used. Therefore, $k$-INN \cite{Korn:ASR2000} and $k$-RNN show their advantages and achieve better results under the ELF6 features. Specifically, SCA \cite{Bai:TIP2016} produces amazing results in this situation, and we obtain this result by setting $k_1=1$, $k_2=1$. After sufficient experiments and careful analysis, 
we conclude that it is due to the effect of inverted indexing that SCA actually determines the similarity from the perspective of the gallery samples.
In GRID, 775 background images are used to expand the gallery set. If we compute the similarity of the probe from their perspective, it is very easy to obtain a low similarity score for the probe because their high ranks are occupied by other background images. In addition, the hard boundaries and context information of the probe set also facilitate this exclusion. Therefore, SCA achieves 
excellent results because it is well suited to the data structure of GRID.  

\subsection{Experiments on Datasets with Multiple-Shot matching: Market-1501 and Mars}
\label{sec:experiments-multiple-shot}

\begin{table}[htbp]
	\caption{Comparison on Market-1501 with ResNet-50-IDE features in terms of different metrics.}
	\label{tab:Market-1501 with ResNet-50-IDE}
	\vspace{4pt}
	\small
	\centering
	\begin{tabular}{l|l|l|l|l|l}
		\hline
		\textbf{Metric} &\textbf{Methods} &\textbf{r=1} &\textbf{r=5} &\textbf{r=10} &\textbf{mAP}\\\hline
		\hline
		\multirow{5}{*}{Euc}  &$k$-NN &69.51 &83.94 &88.69 &44.45\\
		&$k$-INN\cite{Korn:ASR2000} &56.95 &82.96 &88.66 &32.90\\
		&$k$-RNN &69.54 &84.32 &88.75 &36.59\\
		&SCA\cite{Bai:TIP2016} &71.05 &81.95 &85.54 &\textbf{56.93}\\
		&inv-DAKR &69.48 &84.41 &88.66 &45.27\\
		&bi-DAKR &69.66 &84.62 &89.52 &45.50\\
		&inv-DAKR+ &69.52 &84.44 &88.79 &45.37\\
		&bi-DAKR+ &69.69 &\textbf{84.73} &\textbf{89.59} &45.52\\
		&Zhong's\cite{Zhong:CVPR2017} &71.32 &83.43 &88.42 &49.01\\
		&Yu's\cite{Yu:BMVC2017} &\textbf{74.02} &81.92 &85.36 &54.59\\
		\hline
		\multirow{5}{*}{XQDA}  &$k$-NN &75.53 &88.63 &91.66 &53.03\\
		&$k$-INN\cite{Korn:ASR2000} &64.55 &88.66 &92.52 &32.90\\
		&$k$-RNN &75.74 &88.69 &91.75 &39.00\\
		&SCA\cite{Bai:TIP2016} &77.32 &86.46 &89.46 &\textbf{65.98}\\
		&inv-DAKR &76.87 &89.43 &92.67 &54.58\\
		&bi-DAKR &76.87 &89.34 &92.96 &54.68\\
		&inv-DAKR+ &77.02 &89.10 &92.61 &54.53\\
		&bi-DAKR+ &76.90 &\textbf{89.58} &\textbf{93.17} &54.88\\
		&Zhong's\cite{Zhong:CVPR2017} &77.58 &88.57 &91.51 &57.94\\
		&Yu's\cite{Yu:BMVC2017} &\textbf{78.03} &85.57 &88.33 &64.96\\
		\hline
		\multirow{5}{*}{KISSME}  &$k$-NN &77.52 &89.61 &93.05 &53.88\\
		&$k$-INN\cite{Korn:ASR2000} &66.09 &89.58 &93.29 &41.92\\
		&$k$-RNN &77.64 &89.79 &93.11 &43.14\\
		&SCA\cite{Bai:TIP2016} &80.61 &87.77 &90.71 &\textbf{68.73}\\
		&inv-DAKR &78.95 &90.26 &93.53 &55.95\\
		&bi-DAKR &78.92 &90.68 &\textbf{94.24} &55.88\\
		&inv-DAKR+ &78.80 &90.23 &93.68 &55.74\\
		&bi-DAKR+ &78.95 &\textbf{90.77} &94.12 &55.96\\
		&Zhong's\cite{Zhong:CVPR2017} &\textbf{79.90} &89.52 &93.14 &59.37\\
		&Yu's\cite{Yu:BMVC2017} &75.56 &83.17 &86.61 &57.20\\
		\hline
		\multirow{5}{*}{Mahal}  &$k$-NN &77.20 &89.82 &92.99 &52.99\\
		&$k$-INN\cite{Korn:ASR2000} &65.41 &89.34 &93.11 &42.03\\
		&$k$-RNN &77.35 &89.93 &92.93 &46.73\\
		&SCA\cite{Bai:TIP2016} &80.46 &87.65 &90.56 &\textbf{68.75}\\
		&inv-DAKR &78.74 &90.08 &93.71 &54.84\\
		&bi-DAKR &78.41 &90.80 &94.00 &55.09\\
		&inv-DAKR+ &78.71 &90.05 &93.71 &54.73\\
		&bi-DAKR+ &78.41 &\textbf{90.90} &\textbf{94.07} &55.21\\
		&Zhong's\cite{Zhong:CVPR2017} &\textbf{79.13} &89.82 &93.29 &57.73\\
		&Yu's\cite{Yu:BMVC2017} &75.83 &83.64 &86.88 &57.76\\
		\hline
		
	\end{tabular}
	\vspace{-5pt}
\end{table}

\begin{table}[htbp]
	\caption{ Comparison on Market-1501 with Caffe features in terms of different metrics.}
	\label{tab:Market-1501 with Caffe}
	\vspace{4pt}
	\small
	\centering
	\begin{tabular}{l|l|l|l|l|l}
		\hline
		\textbf{Metric} &\textbf{Methods} &\textbf{r=1} &\textbf{r=5} &\textbf{r=10} &\textbf{mAP}\\\hline
		\hline
		\multirow{5}{*}{Euc}  &$k$-NN &55.91 &76.84 &83.79 &31.66\\
		&$k$-INN\cite{Korn:ASR2000} &45.19 &75.03 &83.05 &19.84\\
		&$k$-RNN &56.15 &77.02 &83.85 &23.80\\
		&inv-DAKR &56.95 &77.26 &83.91 &32.41\\
		&bi-DAKR &56.95 &77.41 &84.53 &32.58\\
		&inv-DAKR+ &57.00 &77.42 &83.83 &32.47\\
		&bi-DAKR+ &57.00 &\textbf{77.44} &\textbf{84.63} &32.61\\
		&Zhong's\cite{Zhong:CVPR2017} &\textbf{58.17} &76.90 &83.52 &\textbf{35.22}\\
		\hline
		\multirow{5}{*}{XQDA}  &$k$-NN &61.73 &81.03 &87.29 &37.62\\
		&$k$-INN\cite{Korn:ASR2000} &50.86 &80.97 &87.65 &31.20\\
		&$k$-RNN &62.05 &81.50 &87.23 &26.53\\
		&inv-DAKR &63.30 &82.10 &87.95 &39.33\\
		&bi-DAKR &63.24 &82.19 &88.15 &39.19\\
		&inv-DAKR+ &63.63 &82.19 &88.30 &39.56\\
		&bi-DAKR+ &63.72 &\textbf{82.27} &\textbf{88.42} &39.26\\
		&Zhong's\cite{Zhong:CVPR2017} &\textbf{64.70} &81.18 &87.23 &\textbf{41.63}\\
		\hline
		\multirow{5}{*}{KISSME}  &$k$-NN &61.05 &81.00 &86.46 &36.75\\
		&$k$-INN\cite{Korn:ASR2000} &49.67 &80.40 &86.94 &29.02\\
		&$k$-RNN &61.31 &81.44 &86.97 &23.52\\
		&inv-DAKR &63.45 &81.80 &\textbf{87.74} &38.91\\
		&bi-DAKR &62.86 &82.10 &87.47 &38.51\\
		&inv-DAKR+ &63.72 &81.80 &87.59 &38.82\\
		&bi-DAKR+ &62.89 &\textbf{82.39} &87.35 &38.60\\
		&Zhong's\cite{Zhong:CVPR2017} &\textbf{63.27} &80.88 &86.43 &\textbf{40.50}\\
		\hline
		\multirow{5}{*}{Mahal}  &$k$-NN &60.45 &79.96 &86.16 &35.52\\
		&$k$-INN\cite{Korn:ASR2000} &49.79 &80.11 &86.52 &24.31\\
		&$k$-RNN &60.57 &80.29 &86.37 &27.39\\
		&inv-DAKR &62.20 &80.97 &87.11 &36.93\\
		&bi-DAKR &61.64 &81.15 &87.26 &36.80\\
		&inv-DAKR+ &62.27 &80.93 &87.21 &36.98\\
		&bi-DAKR+ &61.69 &\textbf{81.22} &\textbf{87.31} &36.70\\
		&Zhong's\cite{Zhong:CVPR2017} &\textbf{62.26} &80.29 &86.28 &\textbf{38.57}\\
		\hline
		
	\end{tabular}
	\vspace{-0pt}
\end{table}

\begin{table}[htbp]
	\caption{ Comparison on Market-1501 with LOMO features in terms of different metrics.}
	\label{tab:Market-1501 with LOMO}
	\vspace{4pt}
	\small
	\centering
	\begin{tabular}{l|l|l|l|l|l}
		\hline
		\textbf{Metric} &\textbf{Methods} &\textbf{r=1} &\textbf{r=5} &\textbf{r=10} &\textbf{mAP}\\\hline
		\hline
		\multirow{5}{*}{Euc}  &$k$-NN &15.11 &27.23 &33.70 &4.03\\
		&$k$-INN\cite{Korn:ASR2000} &10.68 &25.92 &32.81 &2.05\\
		&$k$-RNN &15.17 &27.55 &33.73 &3.09\\
		&inv-DAKR &15.65 &27.35 &33.67 &4.18\\
		&bi-DAKR &15.77 &27.67 &34.35 &4.24\\
		&inv-DAKR+ &15.66 &27.13 &33.57 &4.20\\
		&bi-DAKR+ &15.82 &\textbf{27.71} &\textbf{34.37} &4.28\\
		&Zhong's\cite{Zhong:CVPR2017} &\textbf{15.83} &27.02 &33.17 &\textbf{4.55}\\
		\hline
		\multirow{5}{*}{XQDA}  &$k$-NN &28.56 &51.60 &61.82 &13.70\\
		&$k$-INN\cite{Korn:ASR2000} &23.81 &50.50 &61.43 &8.46\\
		&$k$-RNN &28.92 &52.02 &62.29 &7.78\\
		&inv-DAKR &30.97 &52.49 &62.68 &15.10\\
		&bi-DAKR &30.52 &\textbf{52.97} &63.21 &14.86\\
		&inv-DAKR+ &31.53 &53.21 &63.48 &15.43\\
		&bi-DAKR+ &30.76 &52.76 &\textbf{63.87} &15.16\\
		&Zhong's\cite{Zhong:CVPR2017} &\textbf{31.12} &51.48 &61.37 &\textbf{15.86}\\
		\hline
		\multirow{5}{*}{KISSME}  &$k$-NN &41.60 &63.87 &73.43 &19.37\\
		&$k$-INN\cite{Korn:ASR2000} &31.80 &61.22 &69.36 &14.59\\
		&$k$-RNN &41.81 &62.89 &70.49 &14.90\\
		&inv-DAKR &43.41 &63.51 &72.33 &21.97\\
		&bi-DAKR &43.76 &65.56 &75.24 &21.92\\
		&inv-DAKR+ &43.08 &64.22 &73.22 &21.90\\
		&bi-DAKR+ &43.79 &\textbf{65.57} &\textbf{75.27} &21.83\\
		&Zhong's\cite{Zhong:CVPR2017} &\textbf{45.16} &64.01 &73.16 &\textbf{23.45}\\
		\hline
		\multirow{5}{*}{Mahal}  &$k$-NN &35.04 &55.73 &65.62 &13.78\\
		&$k$-INN\cite{Korn:ASR2000} &26.99 &52.08 &60.99 &11.47\\
		&$k$-RNN &35.51 &54.25 &61.13 &10.26\\
		&inv-DAKR &36.79 &55.46 &63.81 &16.93\\
		&bi-DAKR &37.77 &\textbf{58.19} &\textbf{67.58} &16.83\\
		&inv-DAKR+ &36.84 &55.38 &63.79 &16.95\\
		&bi-DAKR+ &37.92 &58.11 &67.49 &\textbf{16.85}\\
		&Zhong's\cite{Zhong:CVPR2017} &\textbf{37.86} &56.00 &65.50 &16.23\\
		\hline
	\end{tabular}
	\vspace{-5pt}
\end{table}

\begin{table}[htbp]
	\caption{ Comparison on Mars with IDE features in terms of different metrics.}
	\label{tab:Mars-IDE}
	\vspace{4pt}
	\small
	\centering
	\begin{tabular}{l|l|l|l|l|l}
		\hline
		\textbf{Metric} &\textbf{Methods} &\textbf{r=1} &\textbf{r=5} &\textbf{r=20} &\textbf{mAP}\\\hline
		\hline
		\multirow{5}{*}{Euc}  &$k$-NN &60.81 &77.93 &87.88 &41.24\\
		&$k$-INN\cite{Korn:ASR2000} &49.34 &76.52 &86.21 &34.34\\
		&$k$-RNN &61.16 &77.73 &85.91 &34.57\\
		&SCA\cite{Bai:TIP2016} &61.92 &74.75 &84.29 &\textbf{50.82}\\
		&inv-DAKR &61.77 &78.08 &87.83 &42.72\\
		&bi-DAKR &61.67 &\textbf{79.39} &88.54 &42.77\\
		&inv-DAKR+ &61.72 &78.18 &87.98 &42.64\\
		&bi-DAKR+ &61.77 &78.99 &\textbf{88.84} &42.72\\
		&Zhong's\cite{Zhong:CVPR2017} &\textbf{62.53} &78.23 &87.58 &44.09\\
		&Yu's\cite{Yu:BMVC2017} &61.11 &72.78 &81.92 &46.69\\
		\hline
		\multirow{5}{*}{XQDA}  &$k$-NN &65.51 &81.72 &90.10 &46.85\\
		&$k$-INN\cite{Korn:ASR2000} &52.93 &80.81 &89.70 &39.20\\
		&$k$-RNN &65.76 &82.02 &89.04 &42.63\\
		&SCA\cite{Bai:TIP2016} &66.82 &79.75 &87.93 &\textbf{57.31}\\
		&inv-DAKR &66.46 &82.63 &90.91 &48.58\\
		&bi-DAKR &66.46 &82.93 &91.21 &48.76\\
		&inv-DAKR+ &67.02 &82.42 &91.11 &48.63\\
		&bi-DAKR+ &66.67 &\textbf{83.13} &\textbf{91.52} &48.98\\
		&Zhong's\cite{Zhong:CVPR2017} &\textbf{67.07} &81.82 &90.10 &50.61\\
		&Yu's\cite{Yu:BMVC2017} &67.02 &77.68 &86.01 &56.63\\
		\hline
		\multirow{5}{*}{KISSME}  &$k$-NN &64.95 &81.01 &89.90 &44.20\\
		&$k$-INN\cite{Korn:ASR2000} &52.68 &79.55 &87.73 &38.81\\
		&$k$-RNN &65.15 &81.01 &87.17 &40.48\\
		&SCA\cite{Bai:TIP2016} &66.16 &78.59 &87.68 &\textbf{55.91}\\
		&inv-DAKR &64.90 &82.37 &90.05 &47.12\\
		&bi-DAKR &66.16 &82.68 &91.57 &47.55\\
		&inv-DAKR+ &64.55 &82.32 &90.66 &46.72\\
		&bi-DAKR+ &66.31 &\textbf{82.83} &\textbf{91.67} &47.89\\
		&Zhong's\cite{Zhong:CVPR2017} &\textbf{66.46} &81.16 &90.10 &48.22\\
		&Yu's\cite{Yu:BMVC2017} &64.39 &75.35 &83.89 &49.38\\
		\hline
		\multirow{5}{*}{Mahal}  &$k$-NN &63.33 &80.51 &87.74 &42.12\\
		&$k$-INN\cite{Korn:ASR2000} &51.97 &77.58 &86.11 &37.76\\
		&$k$-RNN &63.59 &79.80 &85.40 &38.22\\
		&SCA\cite{Bai:TIP2016} &64.09 &77.68 &86.97 &\textbf{54.25}\\
		&inv-DAKR &63.54 &80.51 &89.70 &45.50\\
		&bi-DAKR &65.15 &81.57 &91.16 &45.78\\
		&inv-DAKR+ &63.99 &80.56 &89.19 &45.84\\
		&bi-DAKR+ &\textbf{65.40} &\textbf{81.62} &\textbf{90.86} &46.49\\
		&Zhong's\cite{Zhong:CVPR2017} &\textbf{65.40} &80.71 &89.29 &45.94\\
		&Yu's\cite{Yu:BMVC2017} &63.54 &75.40 &84.49 &49.10\\
		\hline
		
	\end{tabular}
	\vspace{-5pt}
\end{table}

\subsubsection{Dataset Descriptions} Market-1501 is a large dataset for the person ReID tasks generated in an open environment. There are 32,668 boxes of 1,501 people walking captured from six surveillance camera in a network on a campus, and 2,793 distracters are included. In the experiment, 12,936 images of 751 identities are used for training, 19,732 images are used for testing (gallery set), and 3,368 images of 750 identities are randomly selected as probes. Therefore, there are approximately 26.3 ground truths on average for each identity in the gallery set.
Mars is a video extension of Market-1501 and contains approximately 20,000 images of 1,261 identities. In the experiment, 8,298 images of 631 identities are used for training, and the reminders are used for testing. Similarly, 1,980 images of 630 identities are selected as the probe set from a total of 12,180 images, and 3,248 distracters are included. There are approximately 16.2 ground truths on average for each identity in the gallery set.

Since there are multiple ground truths in the both probe and gallery sets for these two datasets, we refer to this as a \emph{multiple-shot matching} scenario. 
In this situation, we also consider the effect of adding the probe samples to the gallery set to research how the addition of these extra probe samples will affect the data local distribution and the performance of our proposals.
In addition, since MRank-$L_{n}$ \cite{Chen:ICIP13} requires the construction of a large-scale affinity matrix and matrix inversion calculation, which are quite time consuming, we do not present results for this method on these two large datasets.

\subsubsection{Experimental Results and Discussion}

We show the experimental results on Market-1501 and Mars in Tables~\ref{tab:Market-1501 with ResNet-50-IDE}, \ref{tab:Market-1501 with Caffe}, \ref{tab:Market-1501 with LOMO} and~\ref{tab:Mars-IDE}. Again, we can observe the performance improvements over the results of the baseline $k$-NN method.
In contrast to $k$-INN \cite{Korn:ASR2000} and $k$-RNN, our proposed inv-DAKR and bi-DAKR still provide performance improvements.
Compared to the reranking method proposed in Zhong et al. \cite{Zhong:CVPR2017}, while the rank-1 accuracy and mAP of inv-DAKR and bi-DAKR are inferior, the results at rank-5, rank-10 and rank-20 are competitive or even superior.

Specifically, based on SCA \cite{Bai:TIP2016}, the superiority of Zhong et al. \cite{Zhong:CVPR2017} at rank-1 
comes from the detection of 
sample unbalance via query expansion. In both Market-1501 and Mars, some identities are associated with only two samples, whereas some others have more than 50 samples; thus, the samples 
of different identities 
are rather unbalanced.
By contrast, in the \emph{perfect single-shot matching} and \emph{imperfect single-shot matching} scenarios, each identity is associated with only 1 or 2 samples, meaning that the samples are 
balanced. 
We observe that the query expansion process for $k$-RNN sets 
in \cite{Zhong:CVPR2017} is very suitable for detecting such sample imbalance. 
Therefore, the rank-1 and mAP of Zhong et al. \cite{Zhong:CVPR2017} are greatly improved on both Market-1501 and Mars, whereas this technique loses its discriminative ability in balanced scenario.
We also observe that SCA \cite{Bai:TIP2016} produces the highest mAP on both datasets. This is also 
due to the inverted index and hard boundary of SCA \cite{Bai:TIP2016}.
However, 
we argue that mAP 
is unfair for the unbalanced dataset because
the precision of a sample with many samples of the same class can have a higher precision, whereas a class with 
few samples will have a very low precision. Therefore, by breaking the hard boundary limitation and expansion of the $k$-RNN set, Zhong et al. \cite{Zhong:CVPR2017} obtained a lower mAP score while enforcing its reranking ability as well as our proposals.
In addition, it should be noted that both local query expansion and sophisticated weighting strategies involving the Jaccard distance are used to refine the ranking list in Zhong et al. \cite{Zhong:CVPR2017}, whereas our approaches are completely built on a smooth kernel function with local density-adaptive parameters. 
When the neighborhood grows, it is difficult for the local query expansion and Jaccard distance to find good matches. Moreover, the reranking methods of SCA \cite{Bai:TIP2016} and Zhong et al. \cite{Zhong:CVPR2017} are sensitive to the values used for various parameters, \textit{e.g.}, $k_1$, $k_2$, $\lambda$ and the implicit parameter in the $k$-RNN set expansion process, whereas our inv-DAKR and bi-DAKR are not strongly sensitive to the parameter $k$, as will be shown in subsection \ref{sec:sensitivity-to-k}.

\begin{figure*}[htbp]
	\vspace{-2pt}
	\centering
	\small
	\subfigure[rank-1]{\includegraphics[clip=true,trim=12 5 15 0,width=0.245\columnwidth]{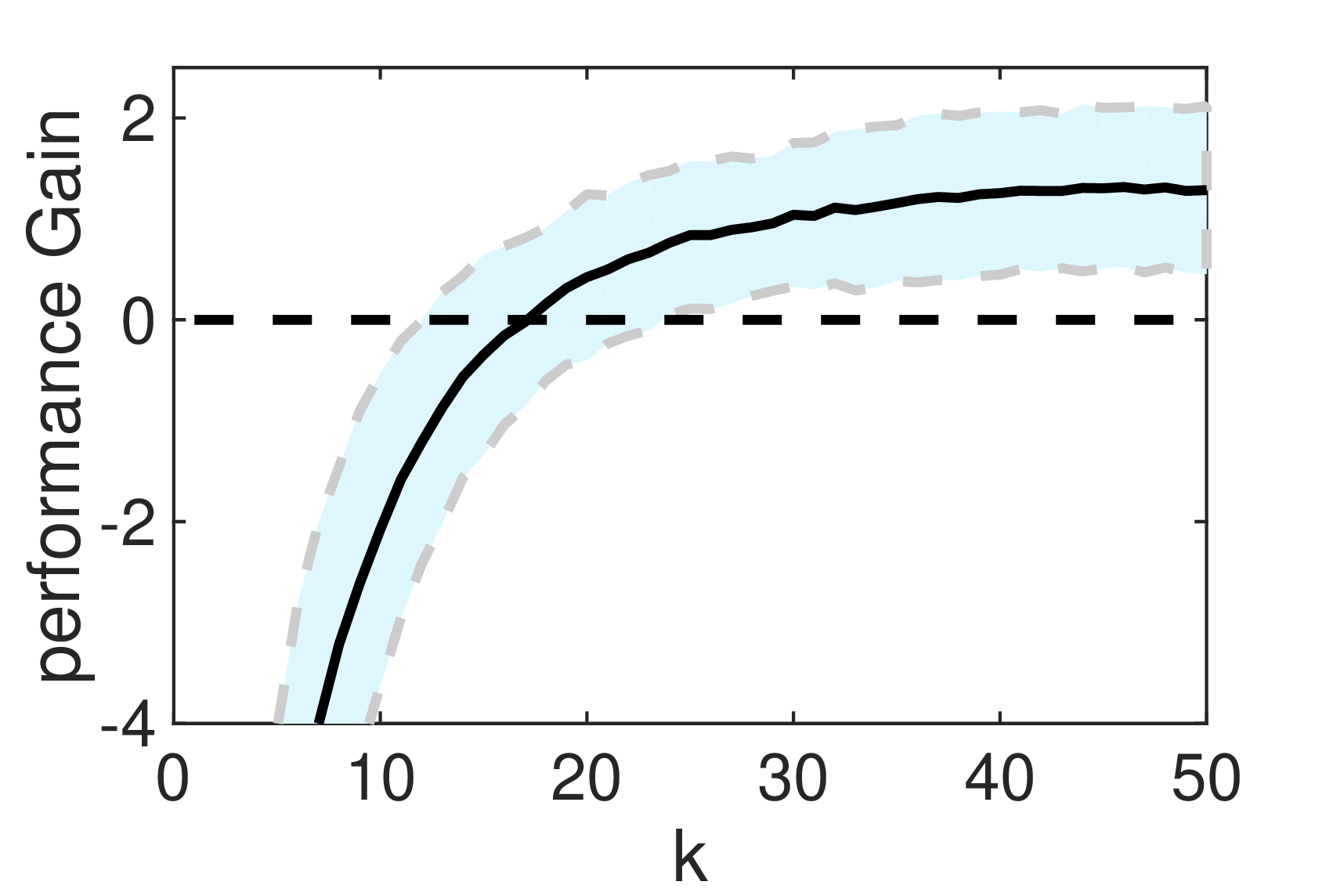}}
	\subfigure[rank-5]{\includegraphics[clip=true,trim=12 5 15 0,width=0.245\columnwidth]{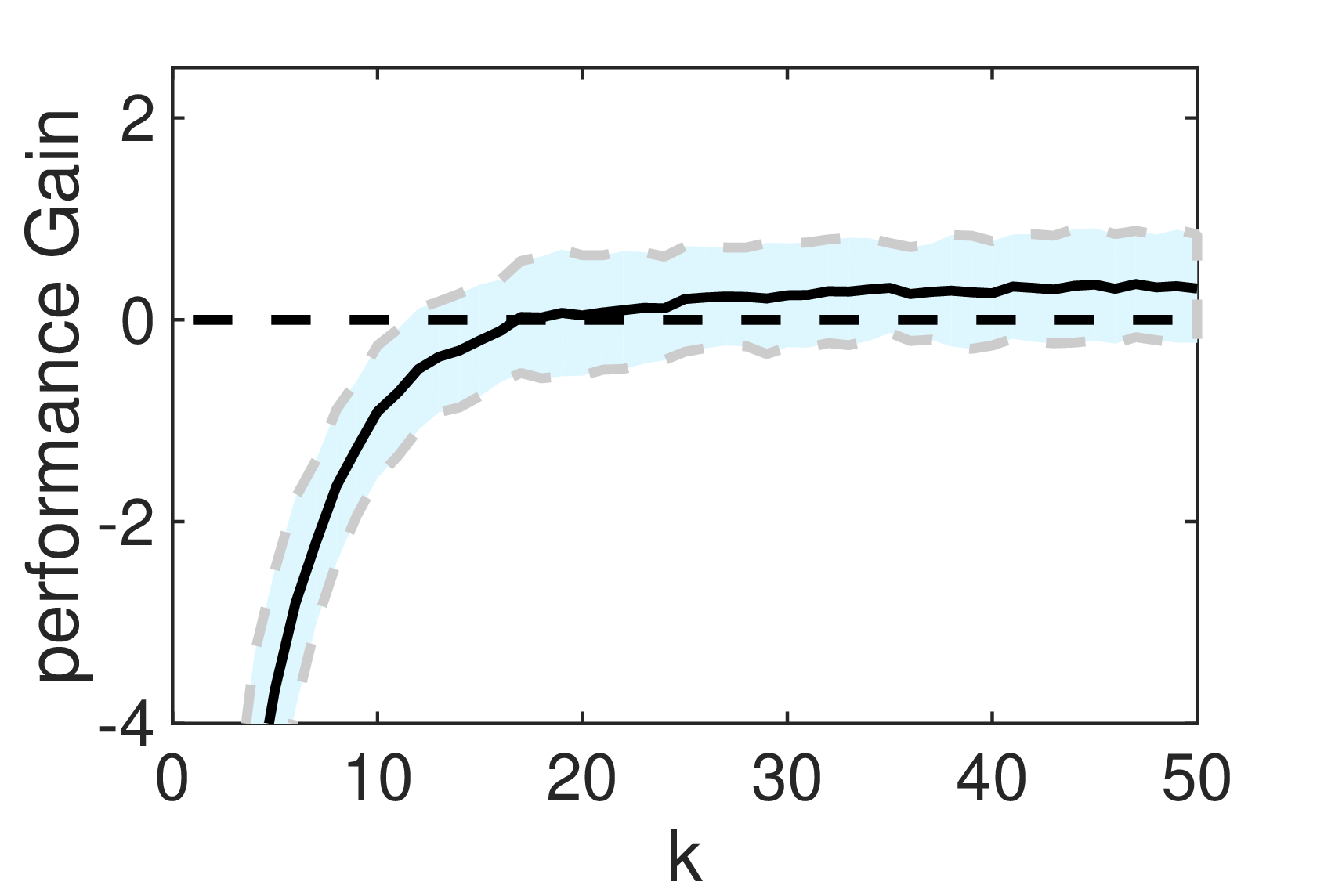}}
	\subfigure[rank-10]{\includegraphics[clip=true,trim=12 5 15 0,width=0.245\columnwidth]{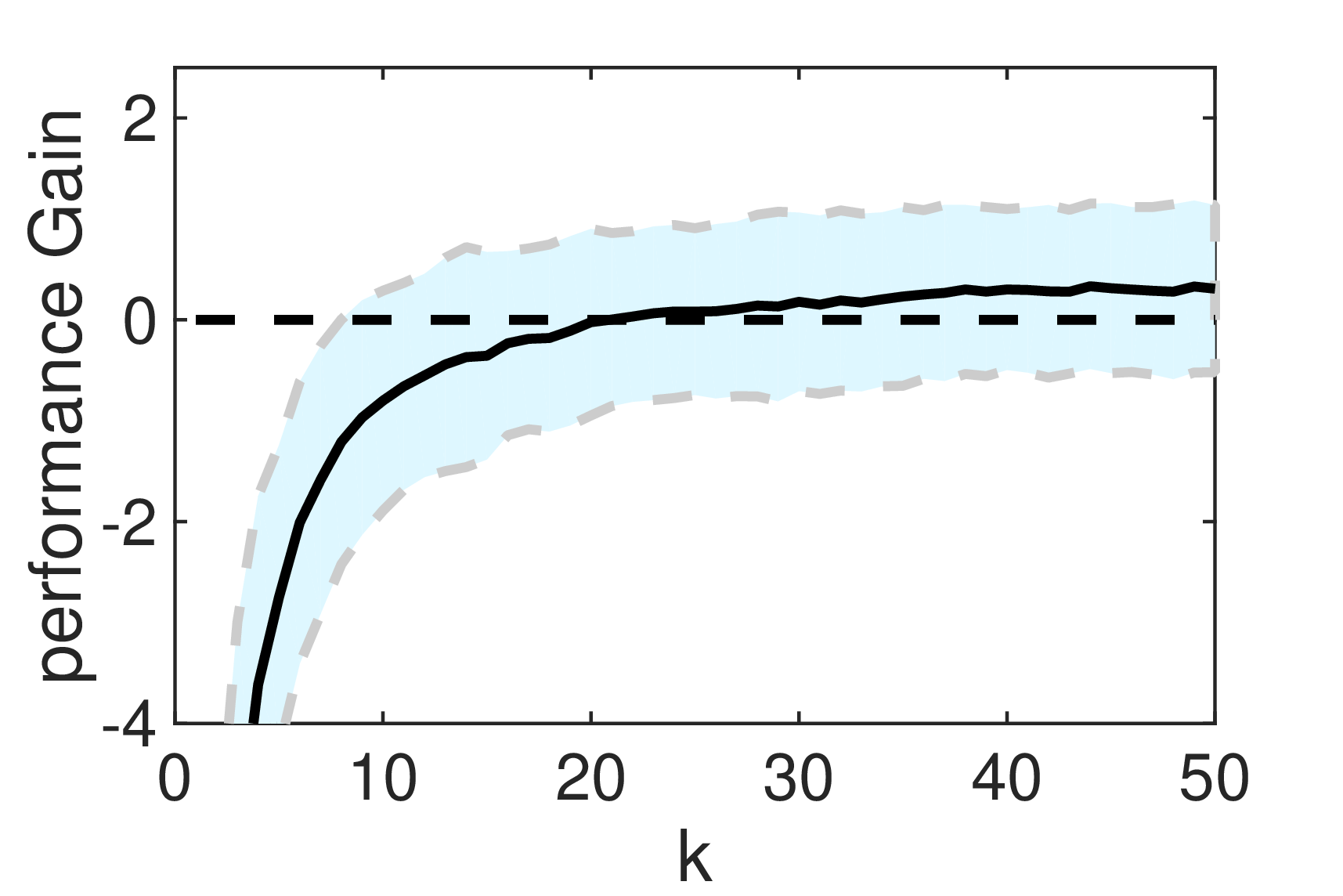}}
	\subfigure[rank-20]{\includegraphics[clip=true,trim=12 5 15 0,width=0.245\columnwidth]{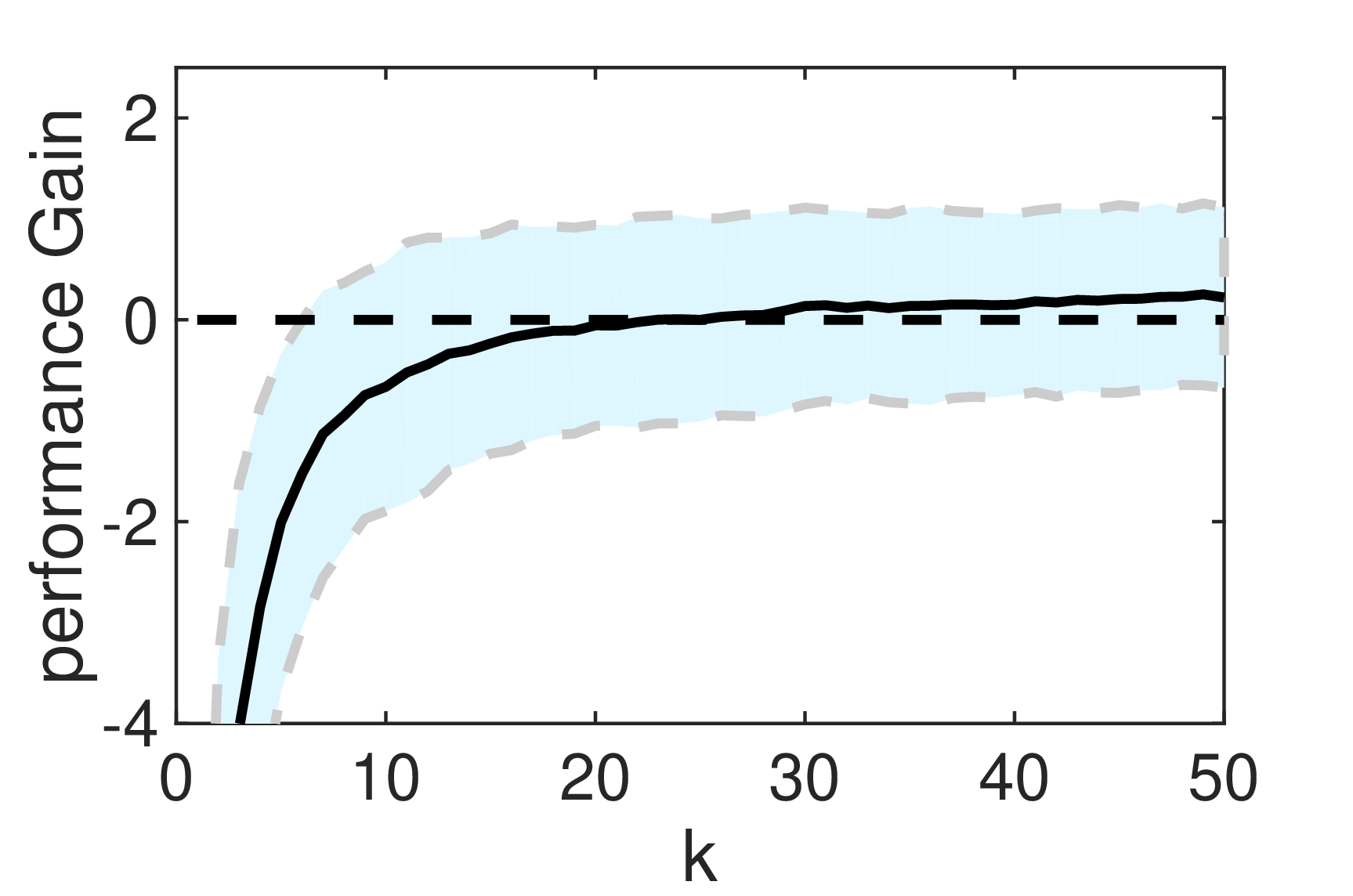}}
	\caption{ Average performance gain of inv-DAKR as a function of $k$ in a multiple-to-multiple matching scenario.} 
	\label{fig:increment_inv_set_3}
	\vspace{-1mm}
\end{figure*}

\begin{figure*}[htbp]
	\vspace{-2pt}
	\centering
	\small
	\subfigure[rank-1]{\includegraphics[clip=true,trim=12 5 15 0,width=0.245\columnwidth]{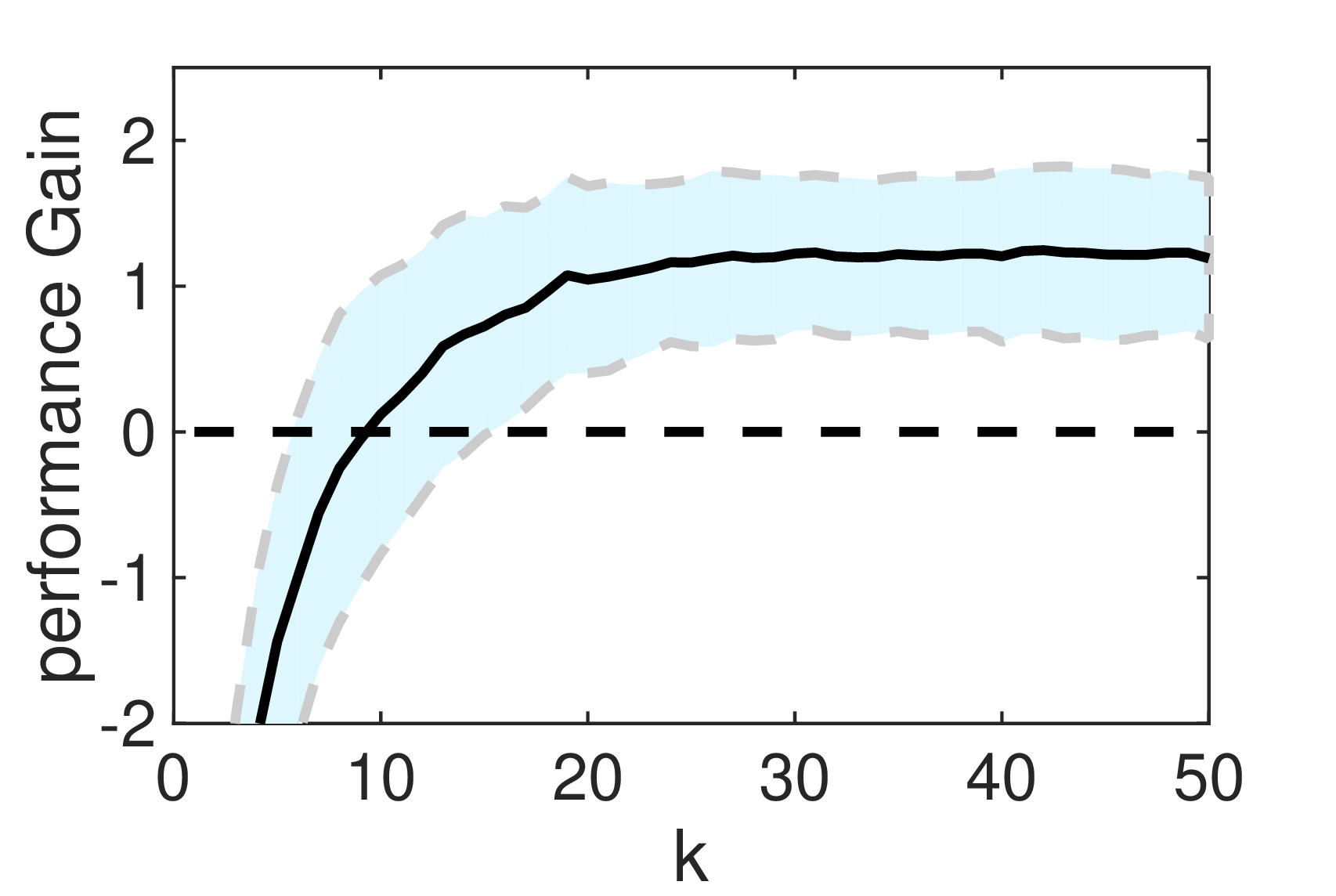}}
	\subfigure[rank-5]{\includegraphics[clip=true,trim=12 5 15 0,width=0.245\columnwidth]{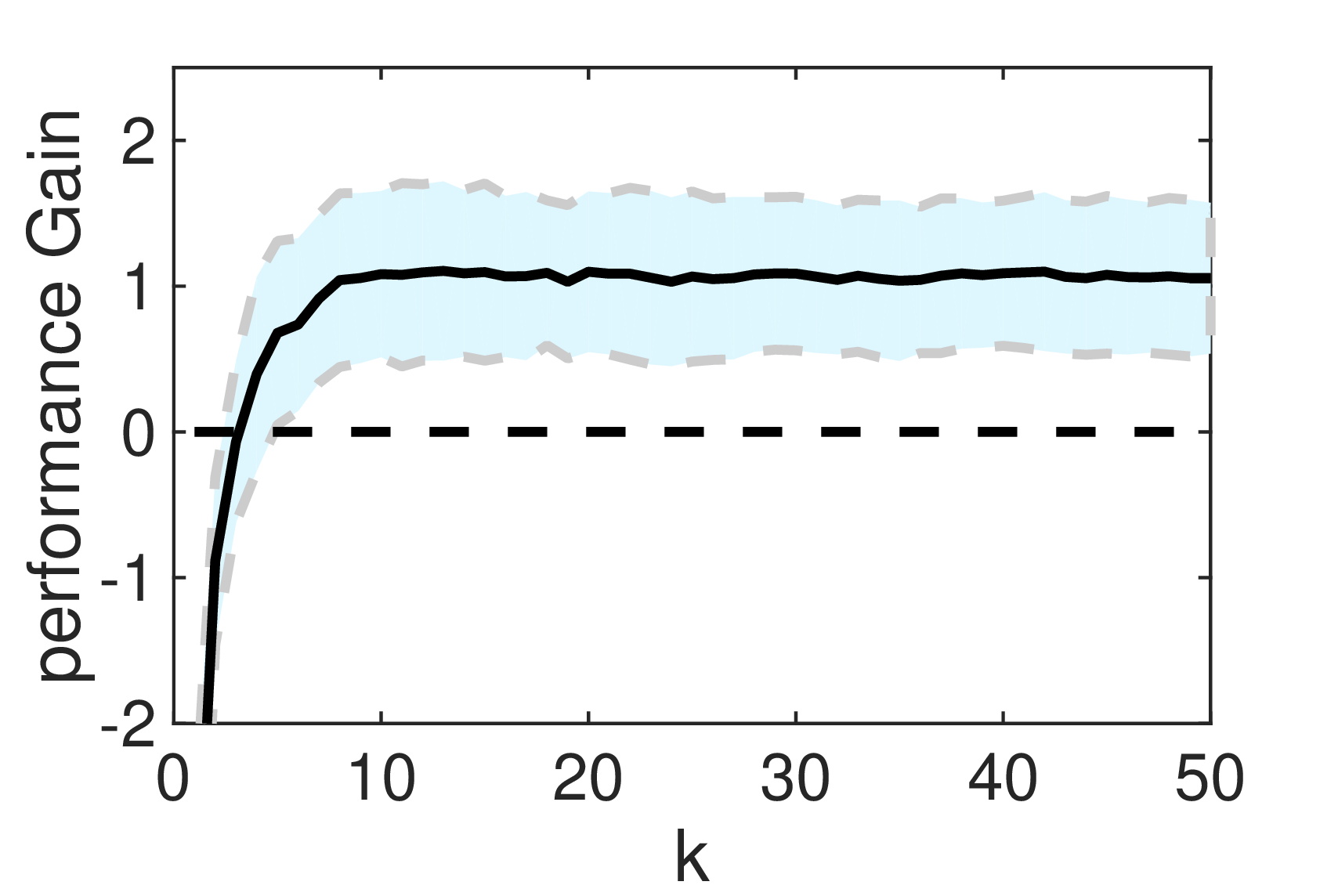}}
	\subfigure[rank-10]{\includegraphics[clip=true,trim=12 5 15 0,width=0.245\columnwidth]{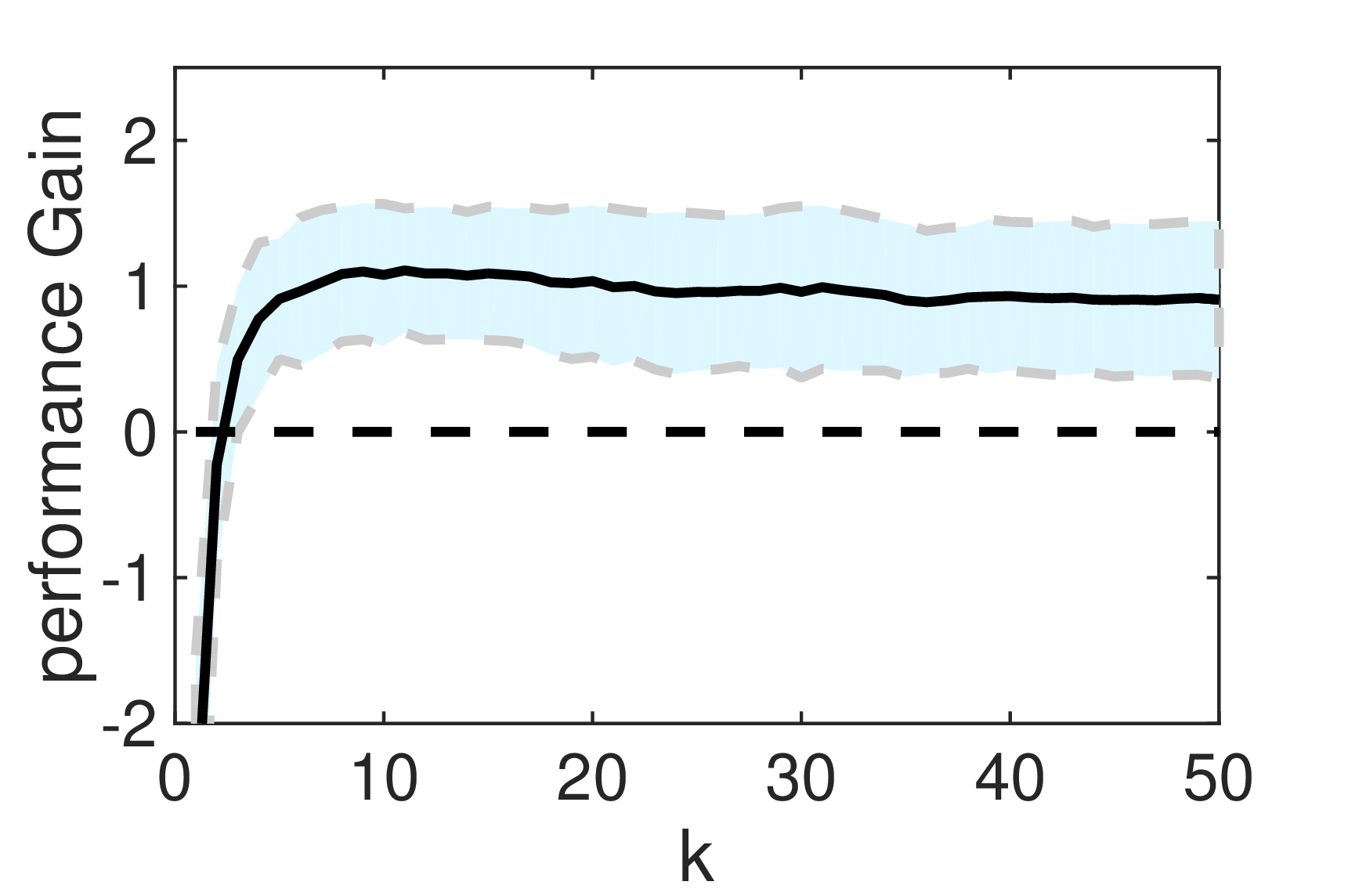}}
	\subfigure[rank-20]{\includegraphics[clip=true,trim=12 5 15 0,width=0.245\columnwidth]{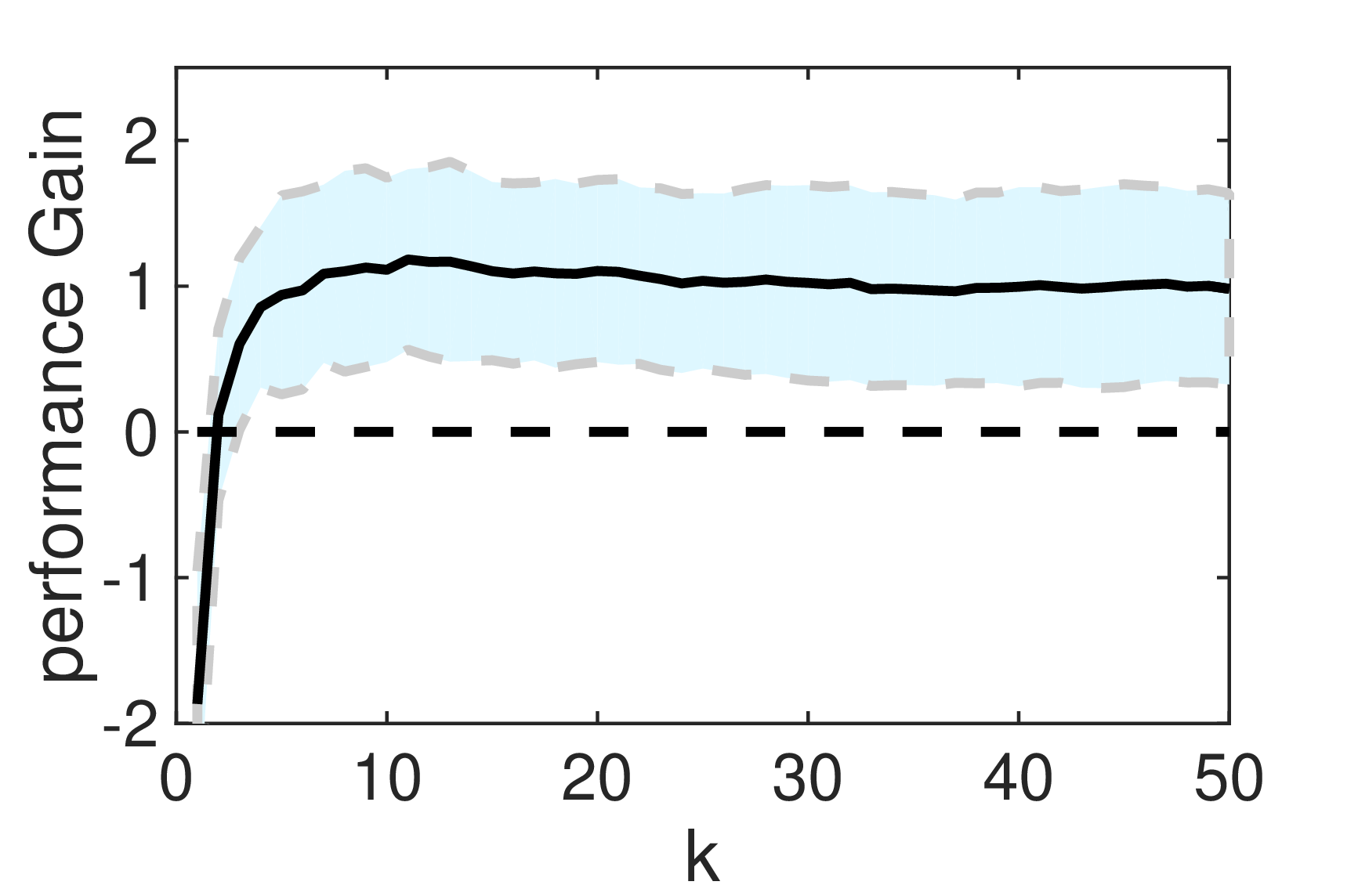}}
	\caption{ Average performance gain of bi-DAKR as a function of $k$ in a multiple-to-multiple matching scenario.} 
	\label{fig:increment_bi_set_3}
	\vspace{-1mm}
\end{figure*}

\begin{figure*}[htbp]
	\vspace{-2pt}
	\centering
	\small
	\subfigure[rank-1]{\includegraphics[clip=true,trim=12 5 15 0,width=0.245\columnwidth]{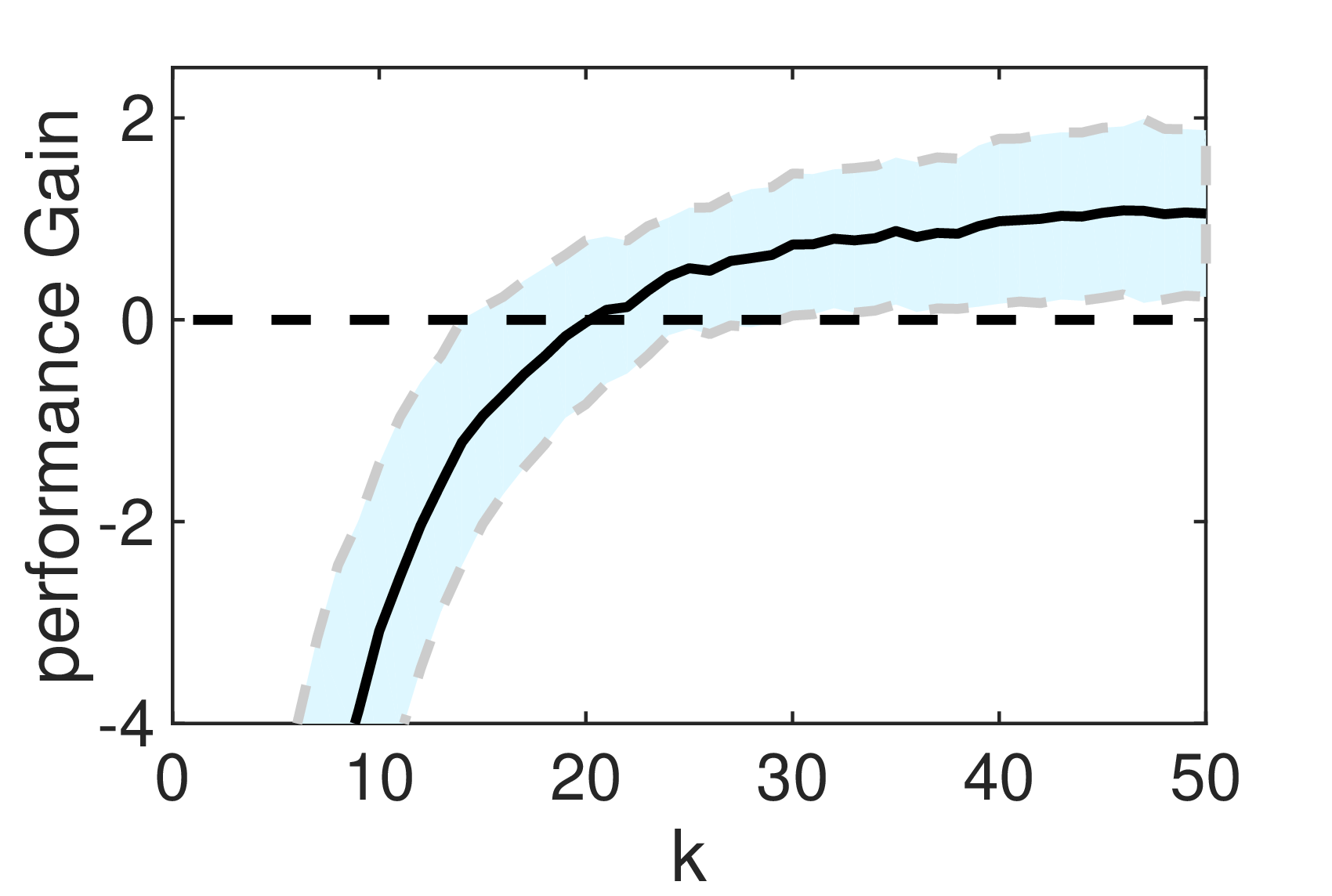}}
	\subfigure[rank-5]{\includegraphics[clip=true,trim=12 5 15 0,width=0.245\columnwidth]{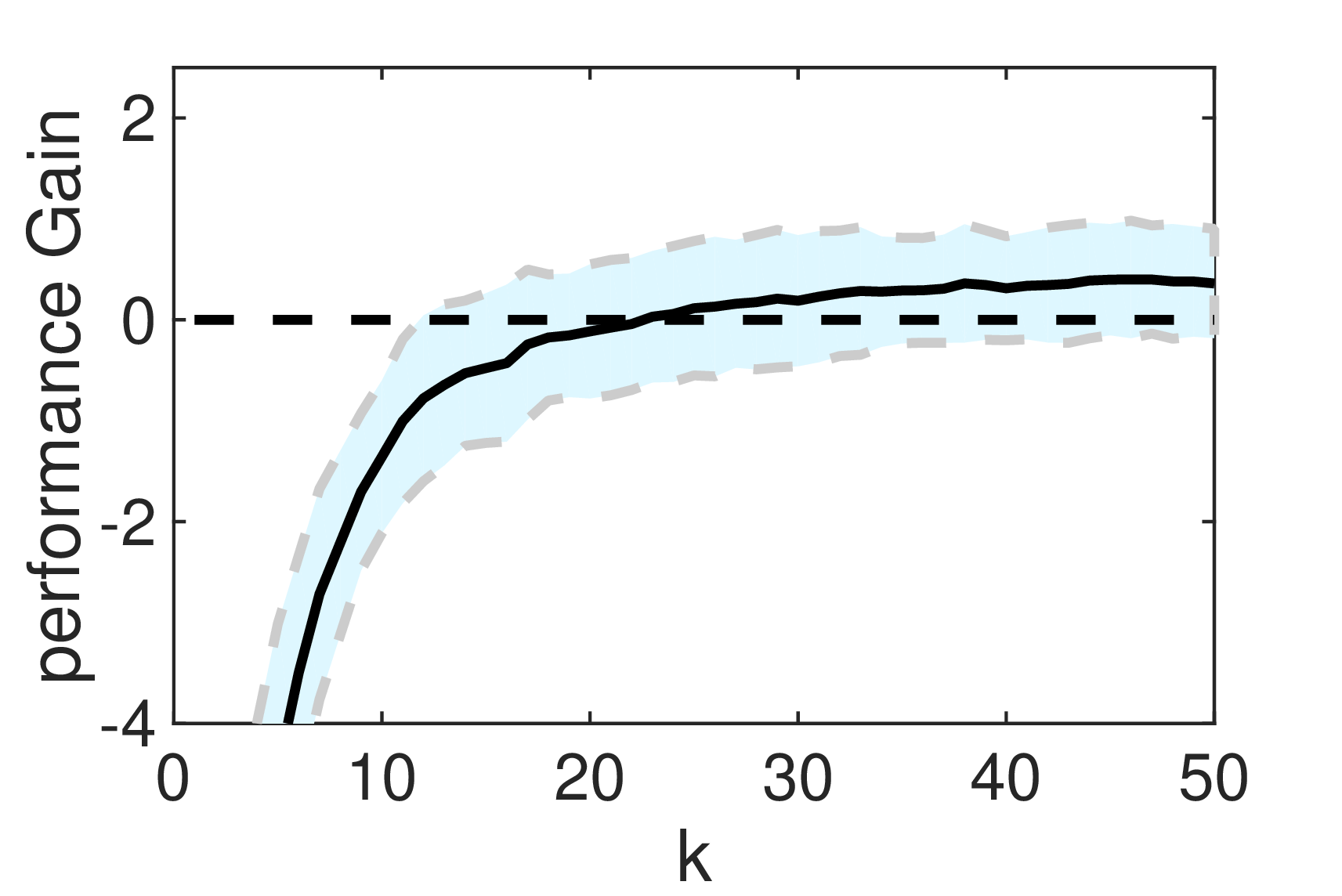}}
	\subfigure[rank-10]{\includegraphics[clip=true,trim=12 5 15 0,width=0.245\columnwidth]{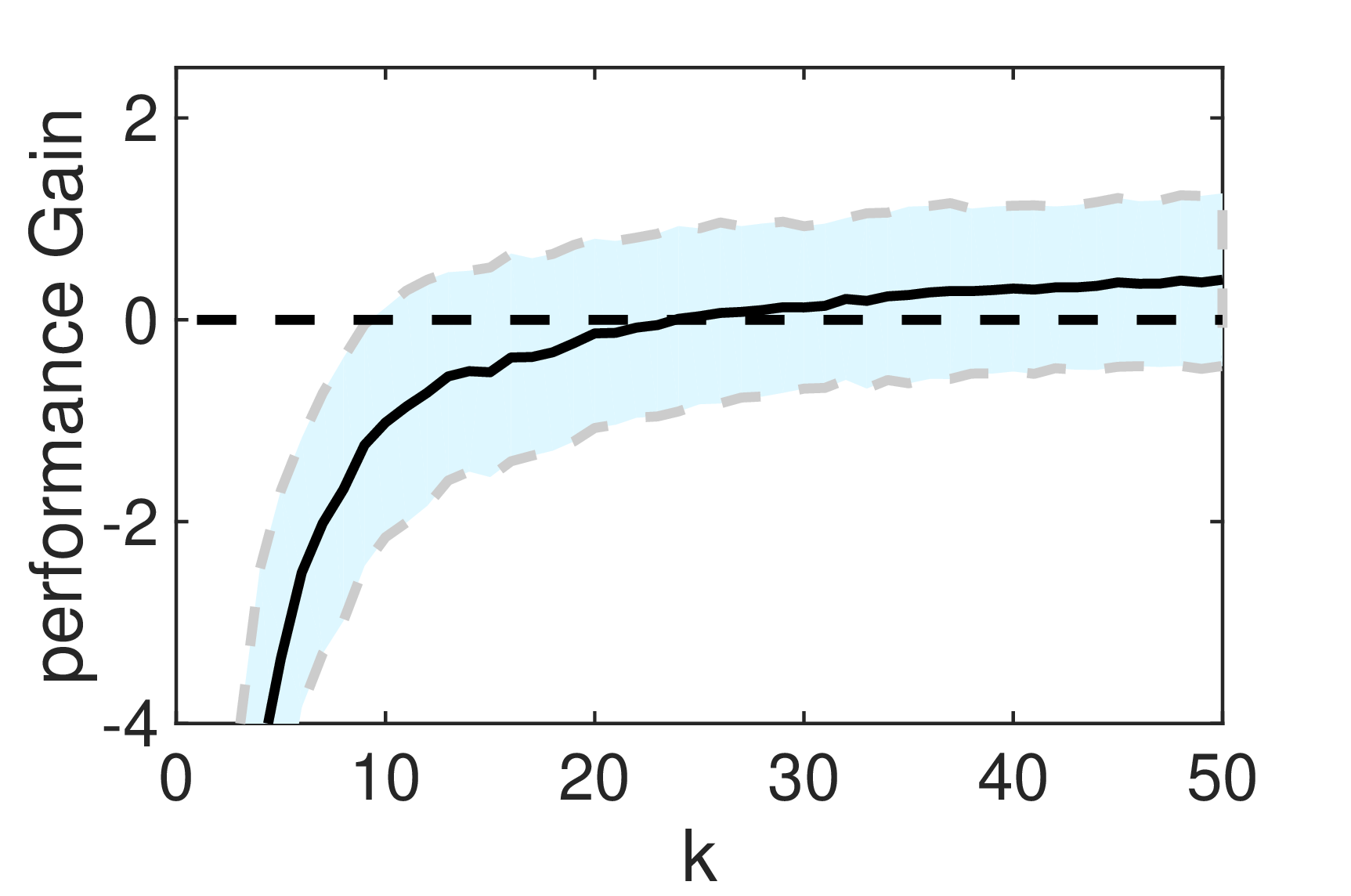}}
	\subfigure[rank-20]{\includegraphics[clip=true,trim=12 5 15 0,width=0.245\columnwidth]{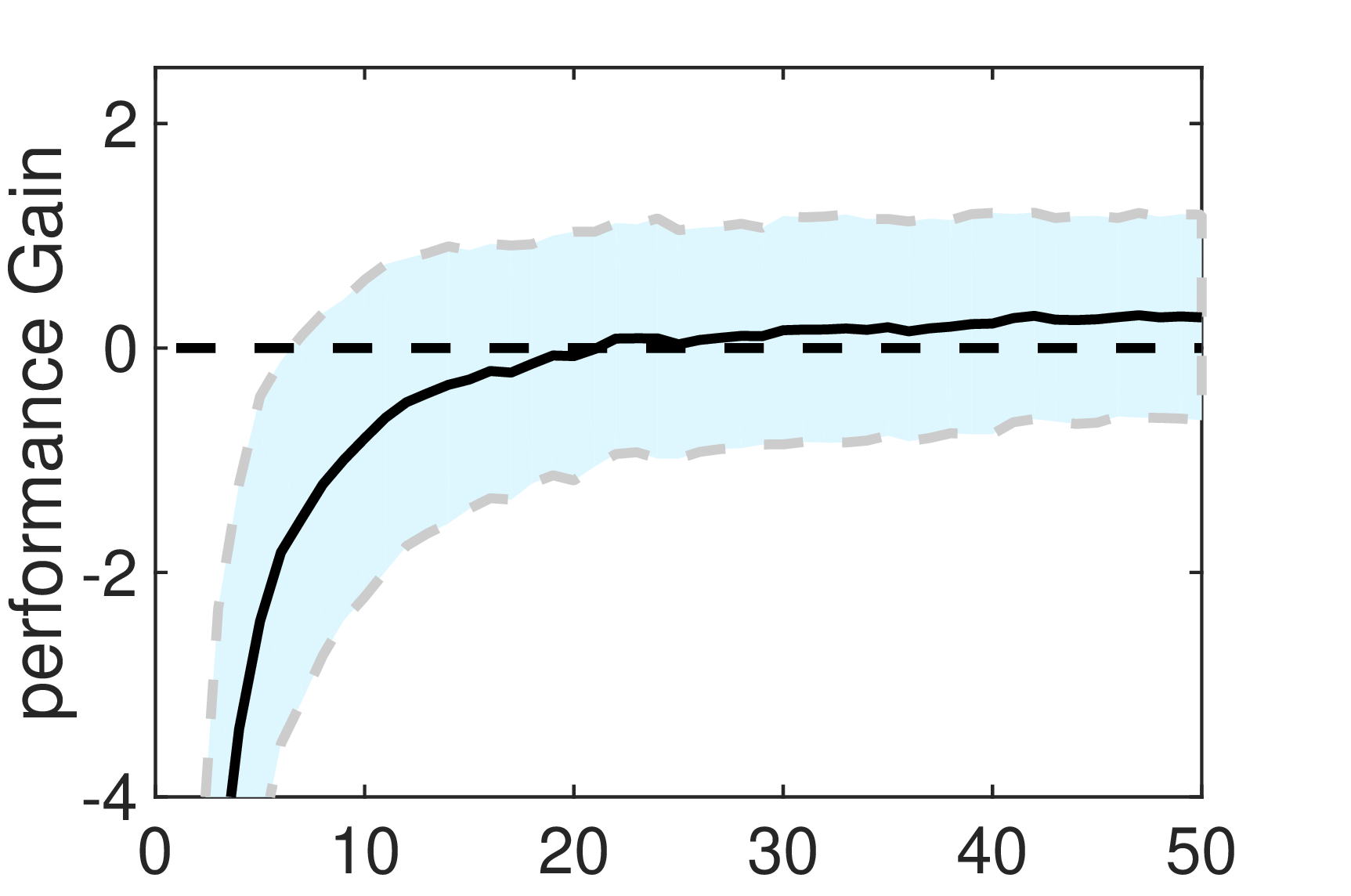}}
	\caption{  Average performance gain of inv-DAKR+ as a function of $k$ in a multiple-to-multiple matching scenario.} 
	\label{fig:increment_inv_new}
	\vspace{-1mm}
\end{figure*}

\begin{figure*}[htbp]
	\vspace{-2pt}
	\centering
	\small
	\subfigure[rank-1]{\includegraphics[clip=true,trim=12 5 15 0,width=0.245\columnwidth]{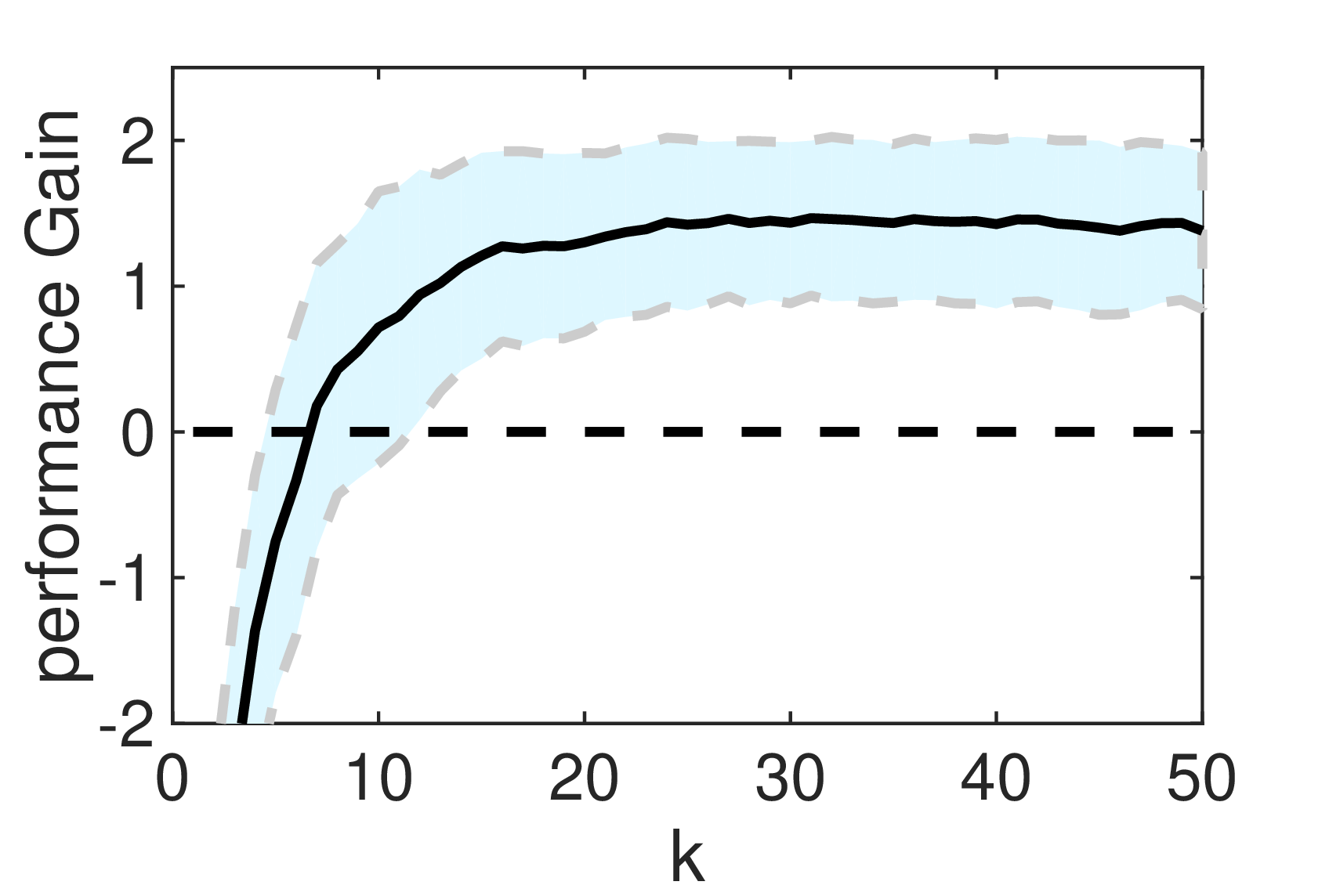}}
	\subfigure[rank-5]{\includegraphics[clip=true,trim=12 5 15 0,width=0.245\columnwidth]{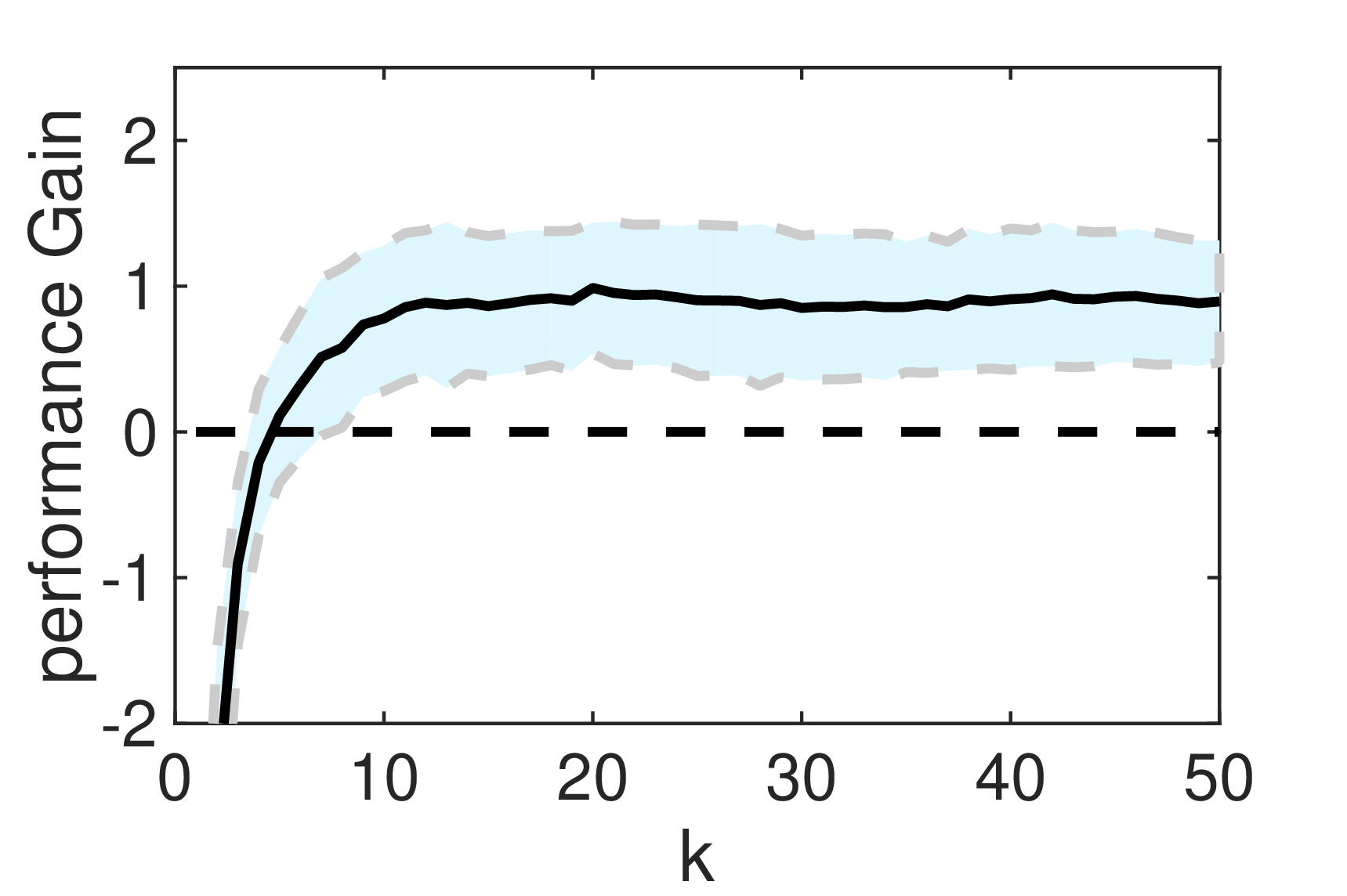}}
	\subfigure[rank-10]{\includegraphics[clip=true,trim=12 5 15 0,width=0.245\columnwidth]{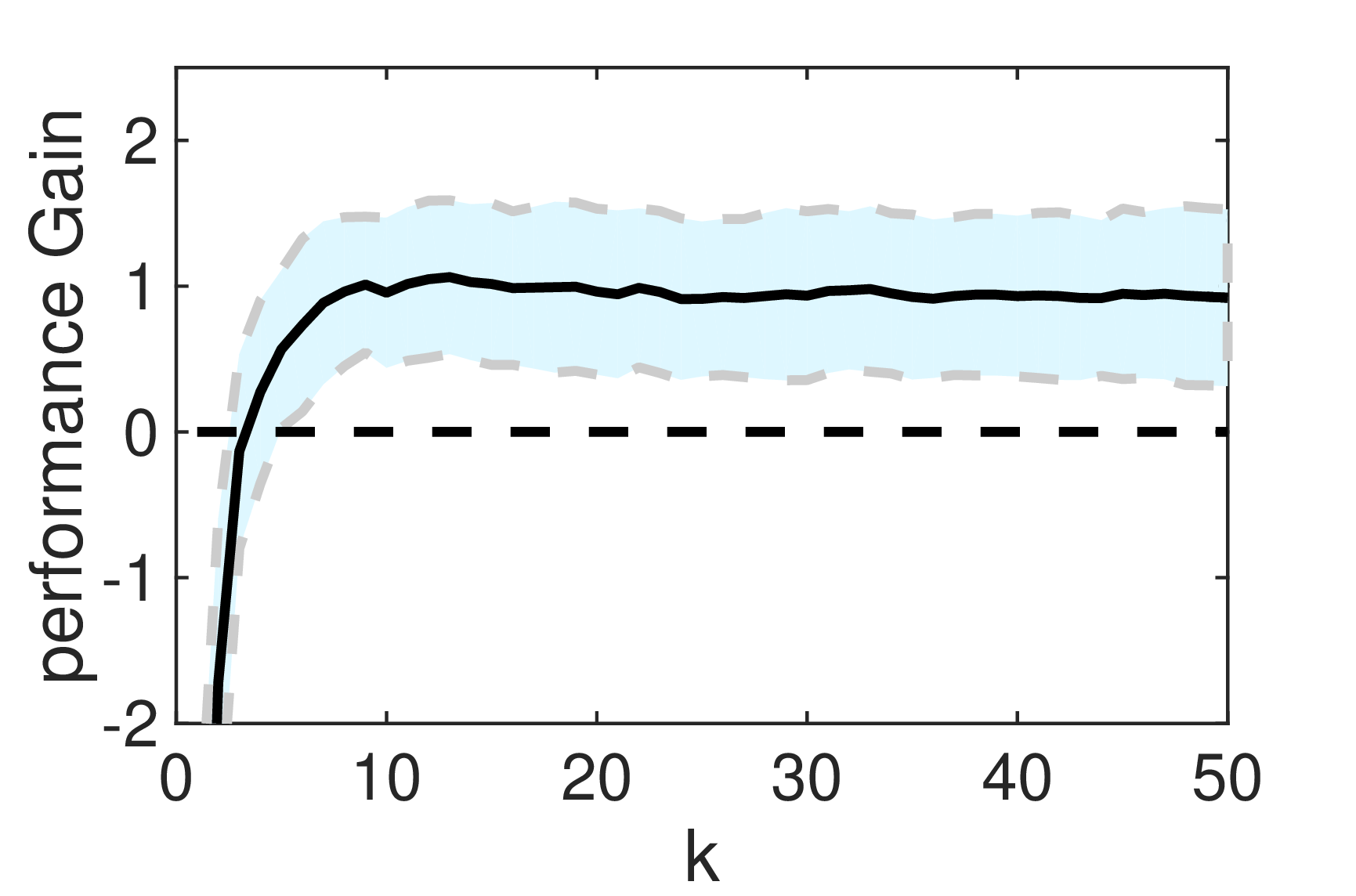}}
	\subfigure[rank-20]{\includegraphics[clip=true,trim=12 5 15 0,width=0.245\columnwidth]{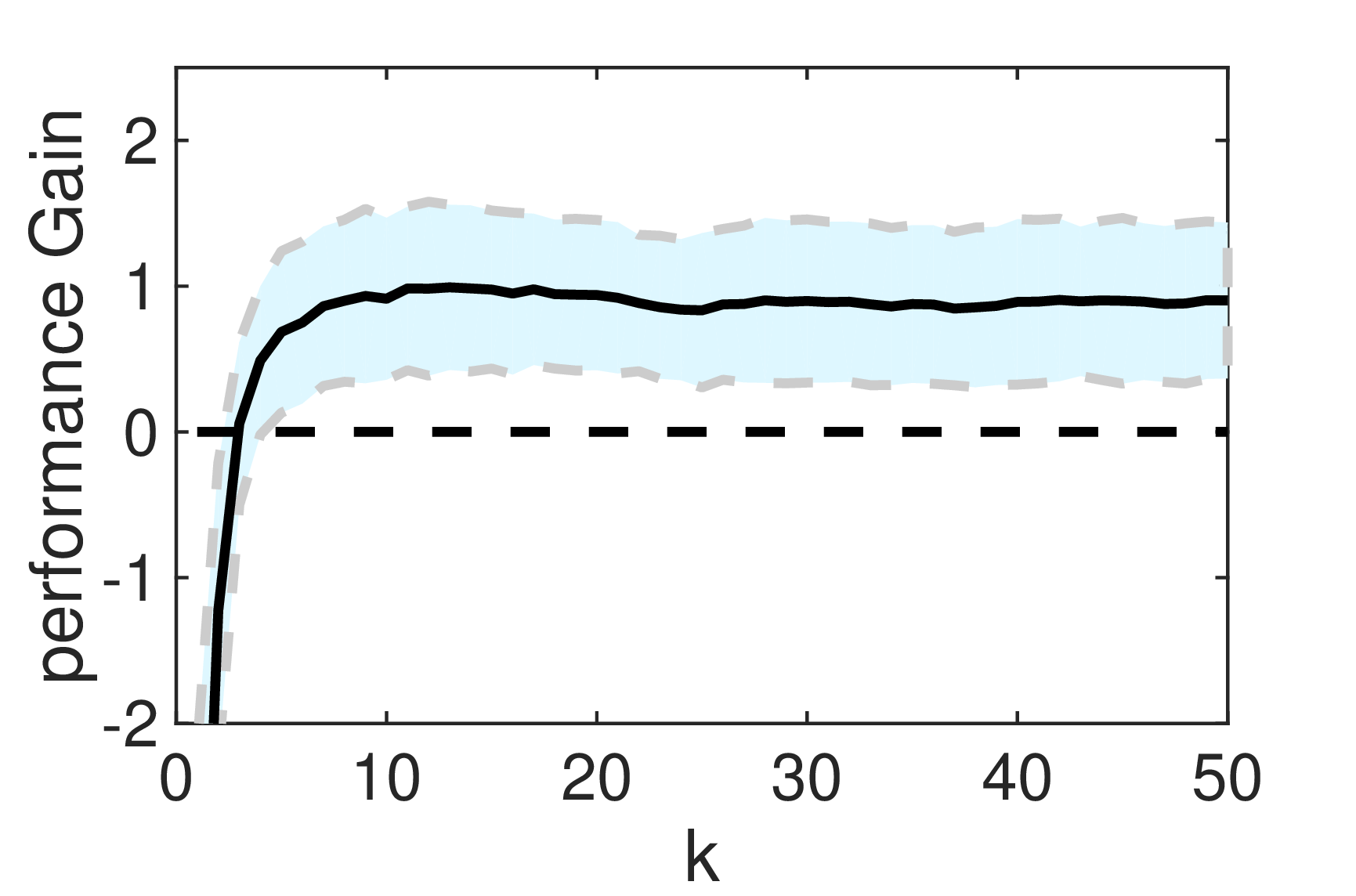}}
	\caption{ Average performance gain of bi-DAKR+ as a function of $k$ in a multiple-to-multiple matching scenario.} 
	\label{fig:increment_bi_new}
	\vspace{-1mm}
\end{figure*}

To gain an understanding of the experimental results, we again show the average performance gains of inv-DAKR and bi-DAKR with respect to the results of the $k$-NN method as functions of the parameter $k$ 
in Figs.~\ref{fig:increment_inv_set_3}, \ref{fig:increment_bi_set_3}, \ref{fig:increment_inv_new} and \ref{fig:increment_bi_new}. From these curves, we observe that, compared to inv-DAKR, bi-DAKR is more promising.
The experimental results show that the performance difference between using the probe set and not using the probe set is again minor in this scenario. 
Note that the ratio of the probe set to the gallery set is approximately $17\%$ for Market-1501 and approximately $19\%$ for Mars. 
Consequently, the extra probe samples cannot significantly improve the data distribution. 

\begin{figure*}[htbp]
	\vspace{-2pt}
	\centering
	\small
	\subfigure[rank-1]{\includegraphics[clip=true,trim=12 5 15 0,width=0.245\columnwidth]{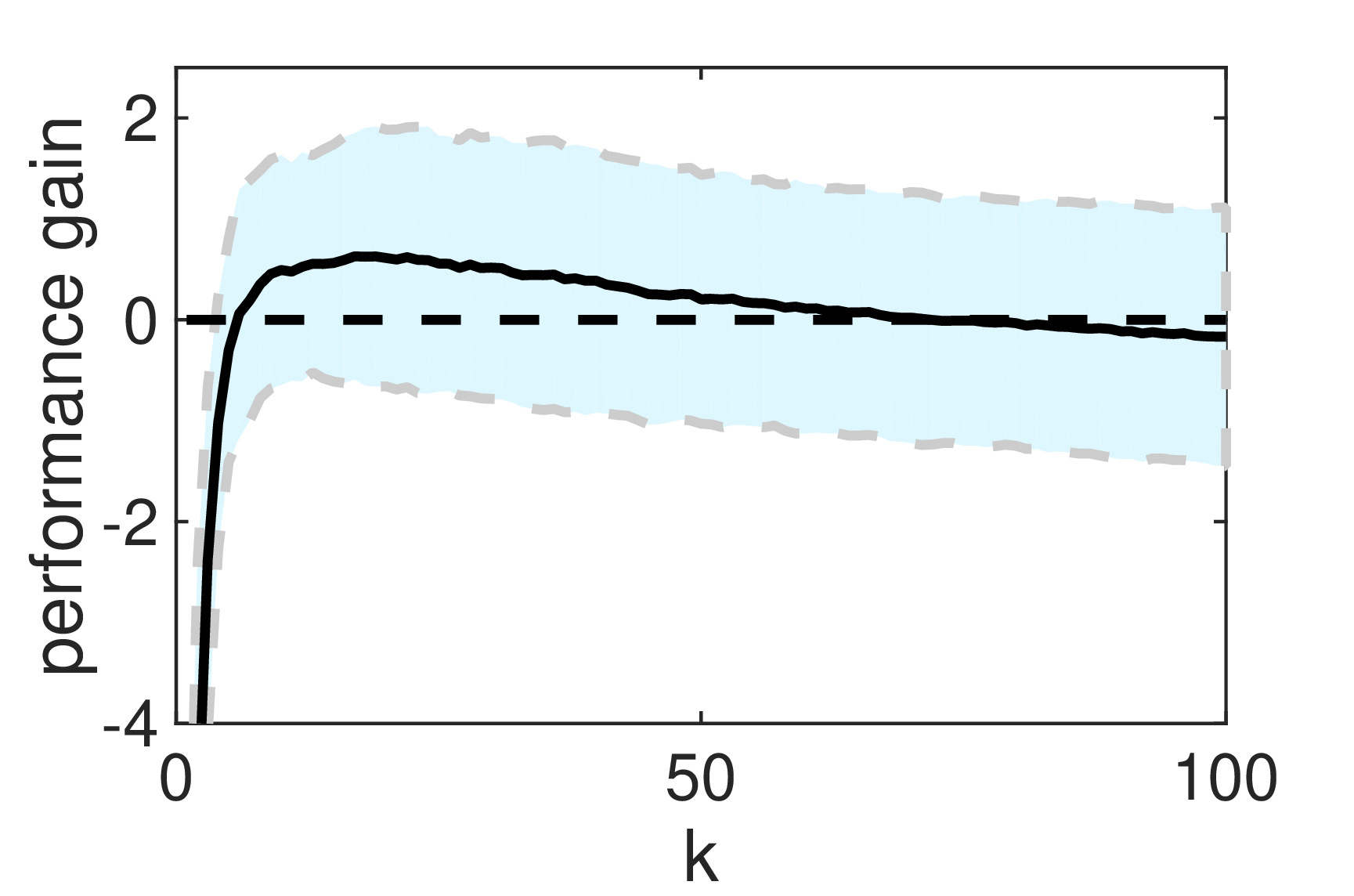}}
	\subfigure[rank-5]{\includegraphics[clip=true,trim=12 5 15 0,width=0.245\columnwidth]{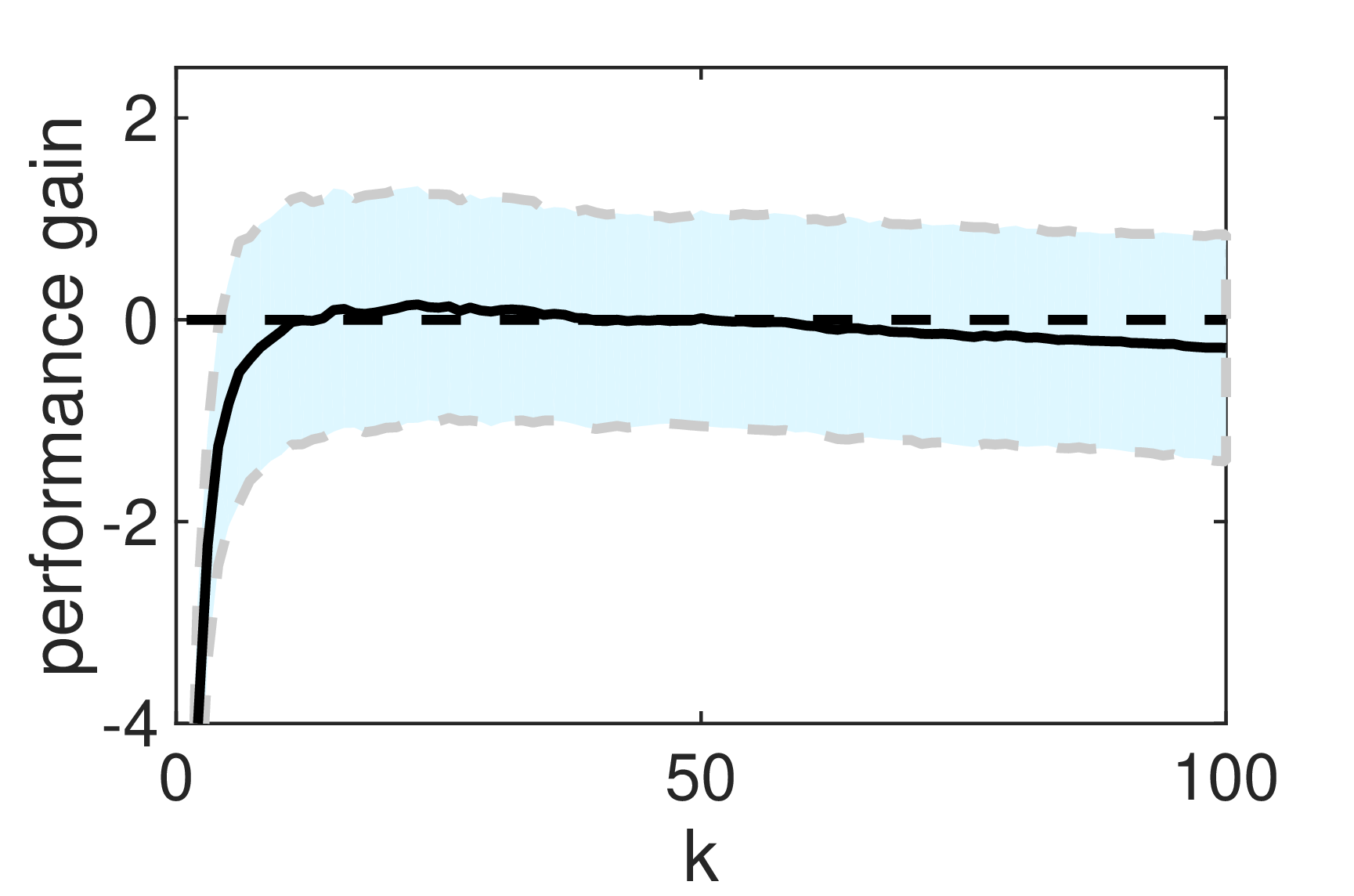}}
	\subfigure[rank-10]{\includegraphics[clip=true,trim=12 5 15 0,width=0.245\columnwidth]{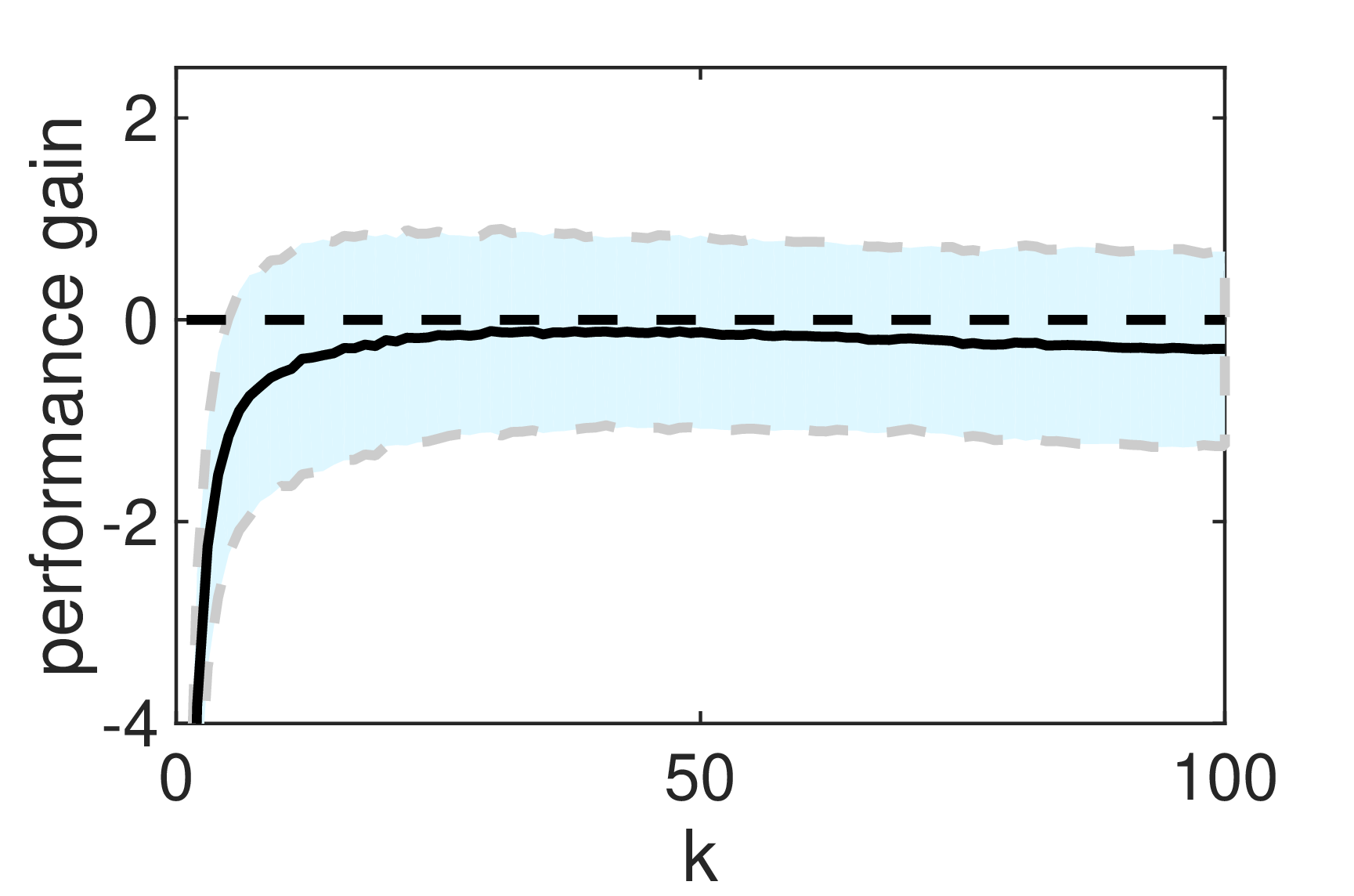}}
	\subfigure[rank-20]{\includegraphics[clip=true,trim=12 5 15 0,width=0.245\columnwidth]{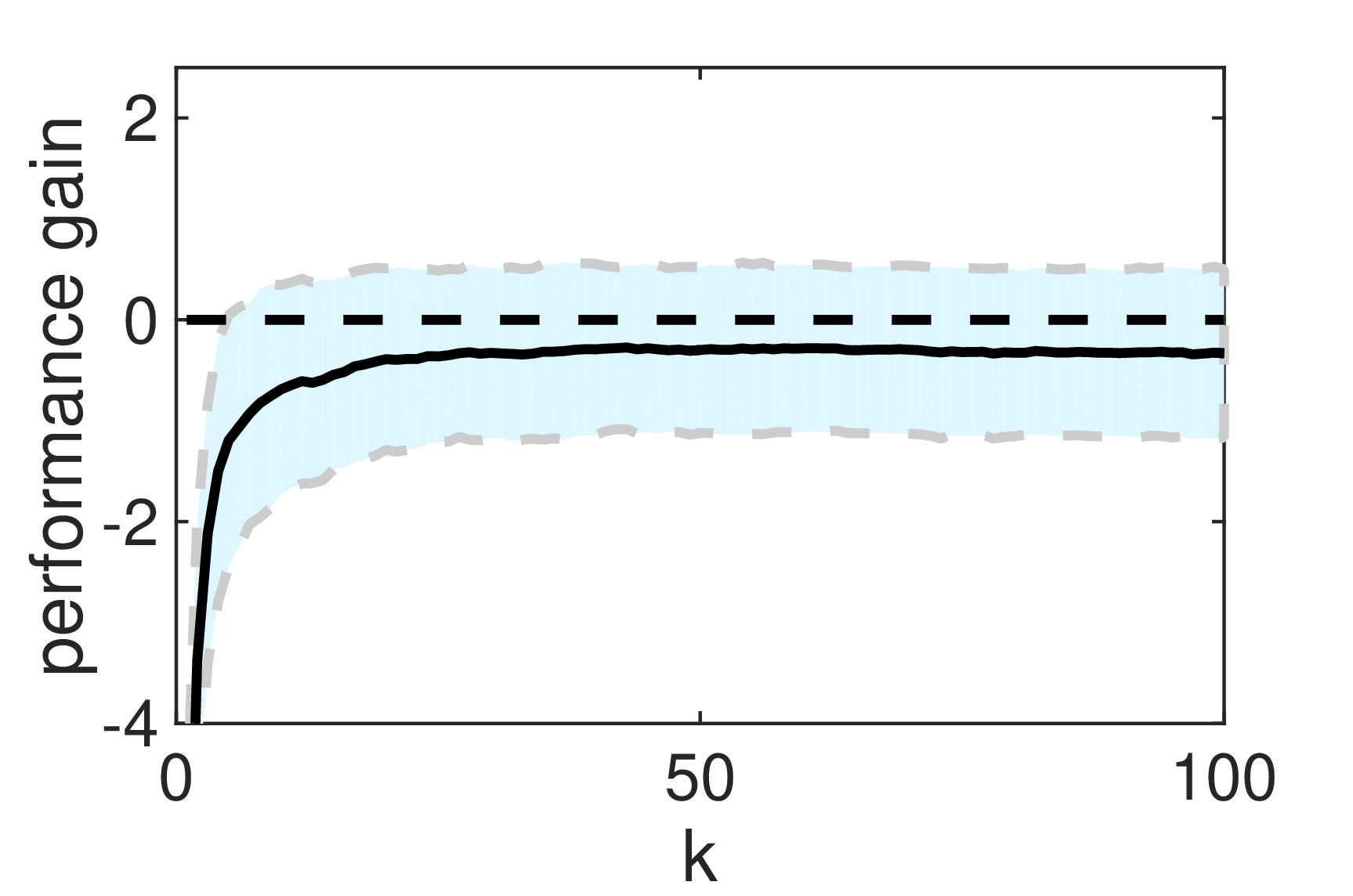}}
	\caption{Average performance gain of inv-DAKR as a function of $k$ in a multiple-to-multiple matching scenario after the addition of a large number of extra samples.} 
	\label{fig:increment_inv_set_3_large}
	\vspace{-1mm}
\end{figure*}

\begin{figure*}[htbp]
	\vspace{-2pt}
	\centering
	\small
	\subfigure[rank-1]{\includegraphics[clip=true,trim=12 5 15 0,width=0.245\columnwidth]{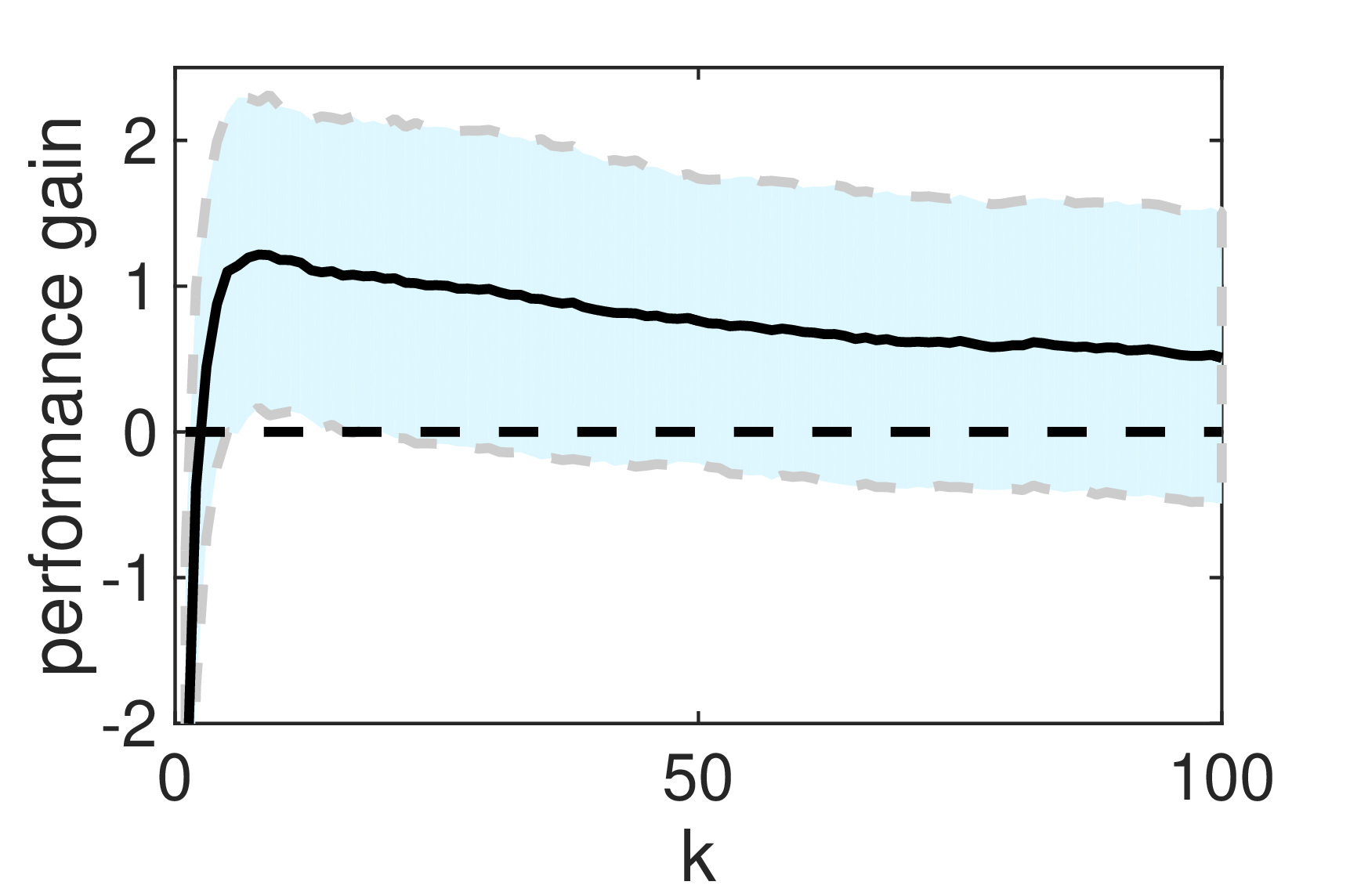}}
	\subfigure[rank-5]{\includegraphics[clip=true,trim=12 5 15 0,width=0.245\columnwidth]{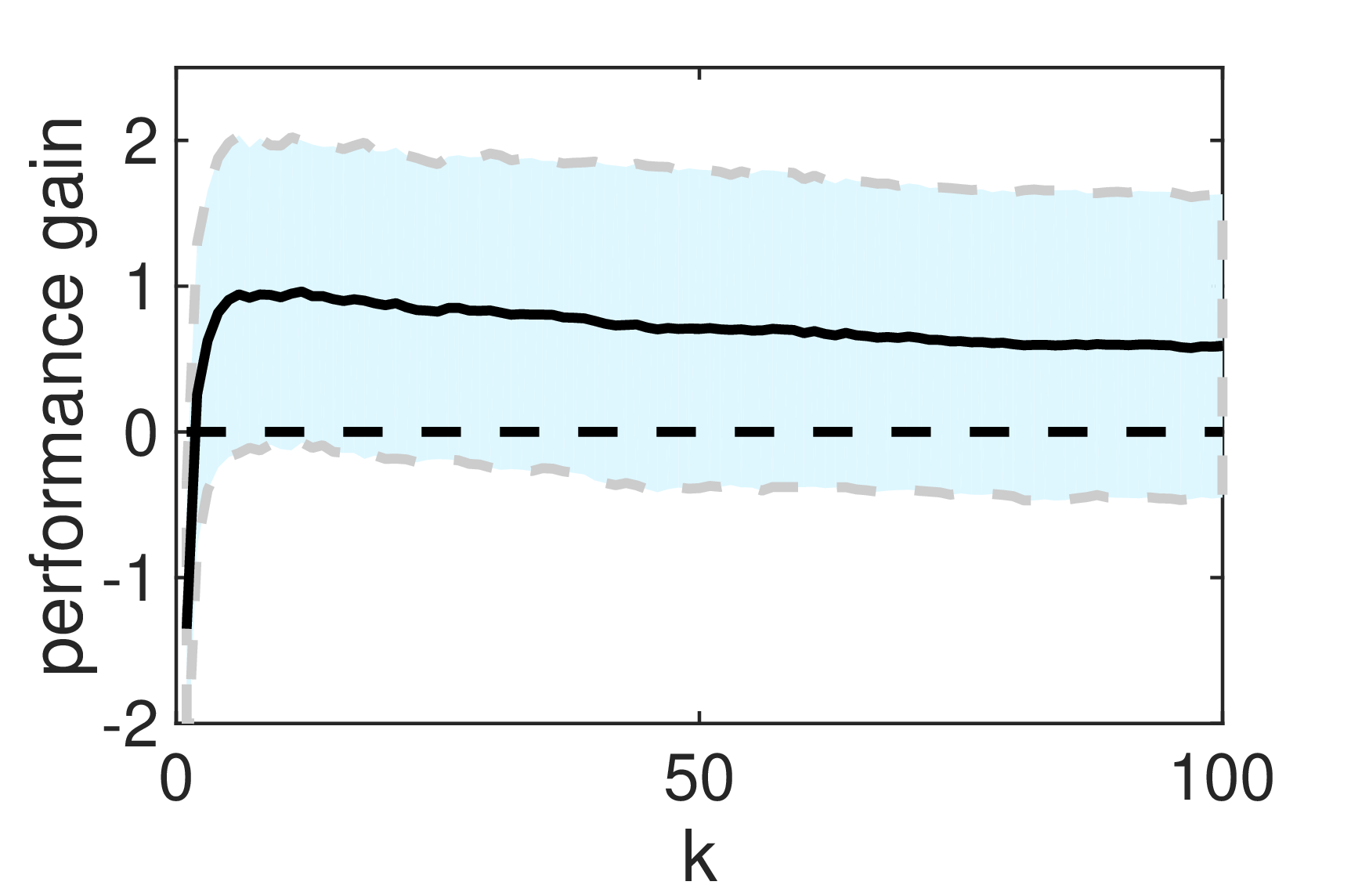}}
	\subfigure[rank-10]{\includegraphics[clip=true,trim=12 5 15 0,width=0.245\columnwidth]{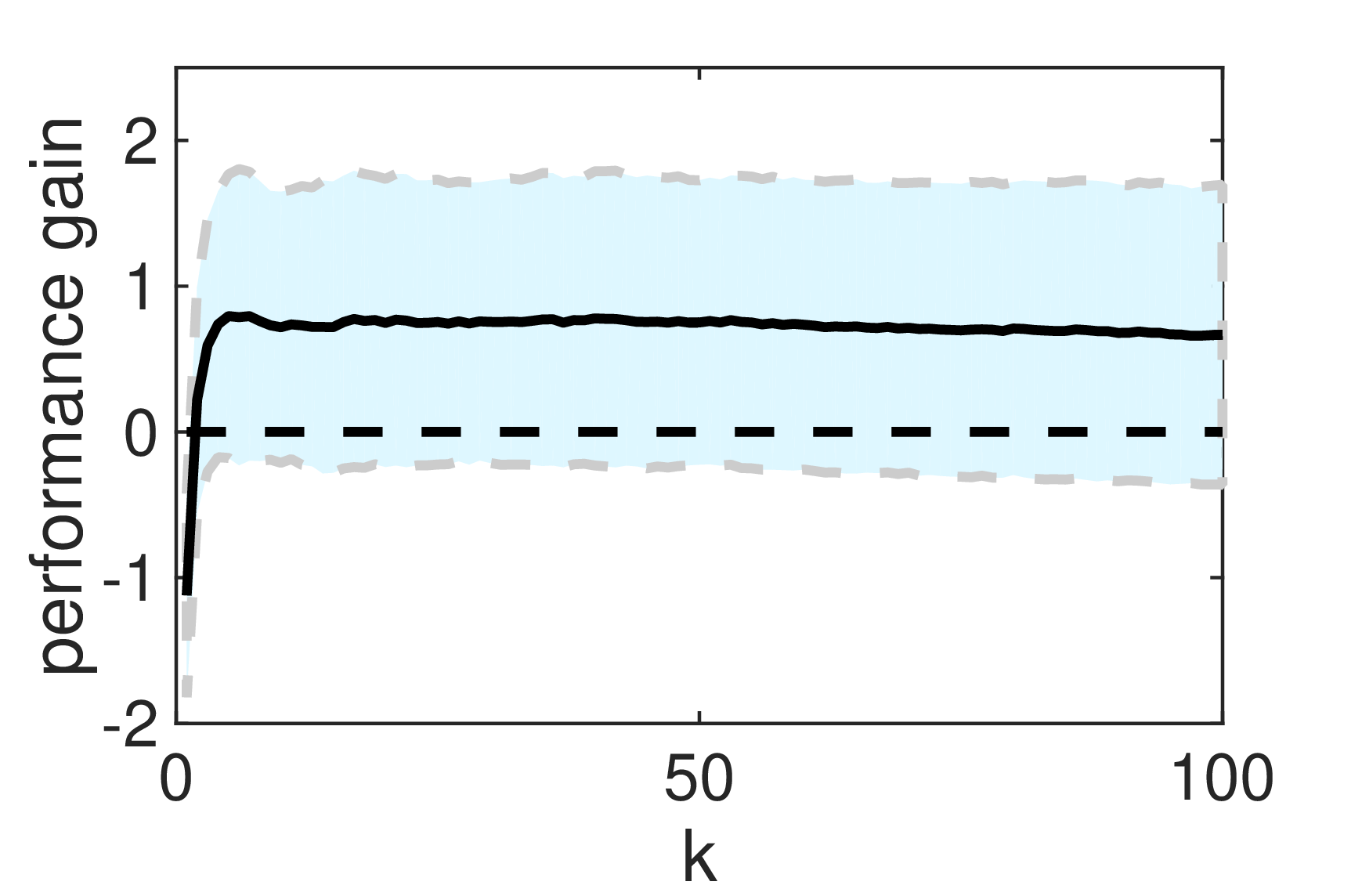}}
	\subfigure[rank-20]{\includegraphics[clip=true,trim=12 5 15 0,width=0.245\columnwidth]{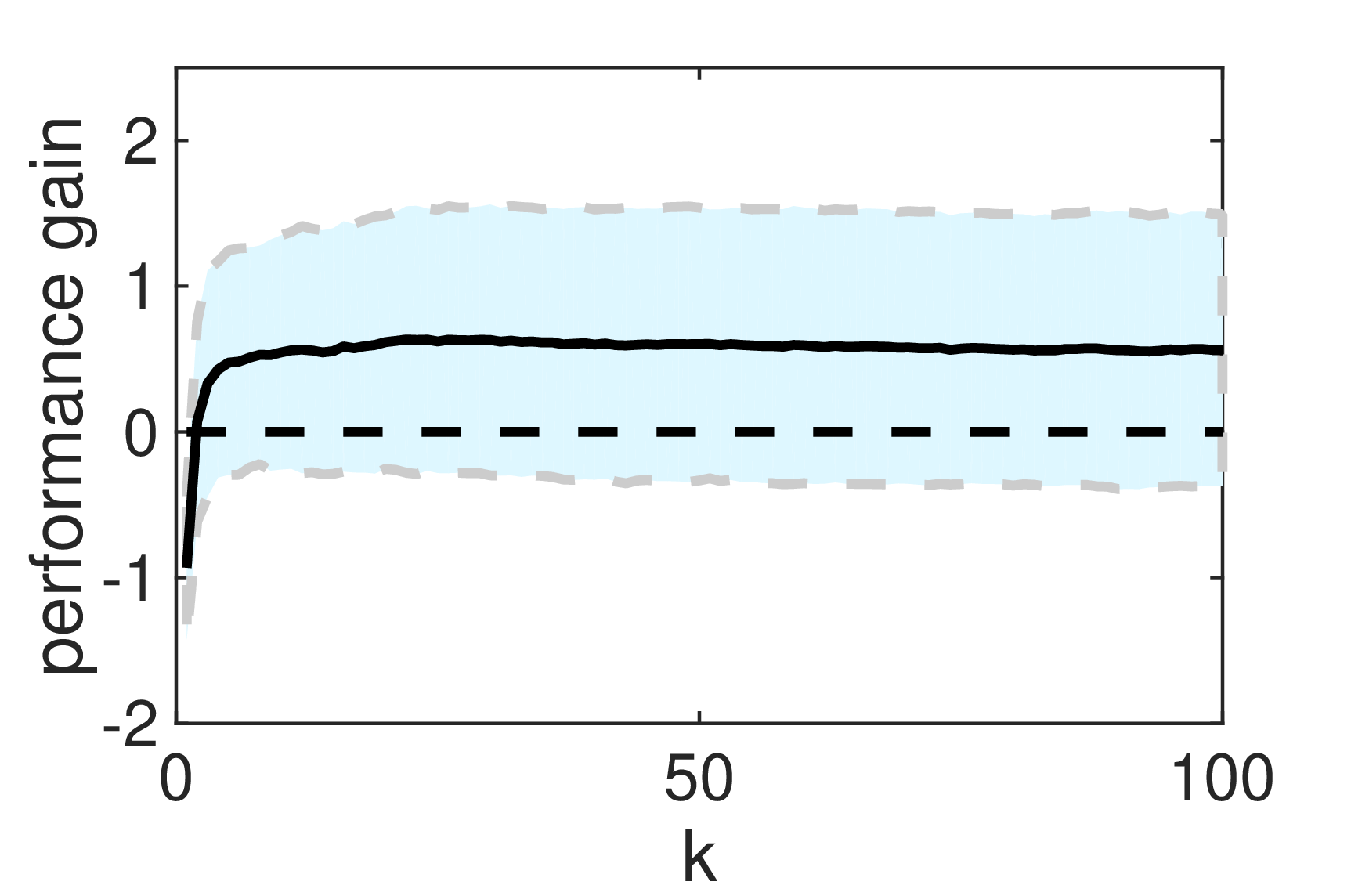}}
	\caption{Average performance gain of bi-DAKR as a function of $k$ in a multiple-to-multiple matching scenario after the addition of a large number of extra samples.} 
	\label{fig:increment_bi_set_3_large}
	\vspace{-1mm}
\end{figure*}

\begin{figure*}[htbp]
	\vspace{-2pt}
	\centering
	\small
	\subfigure[rank-1]{\includegraphics[clip=true,trim=12 5 15 0,width=0.245\columnwidth]{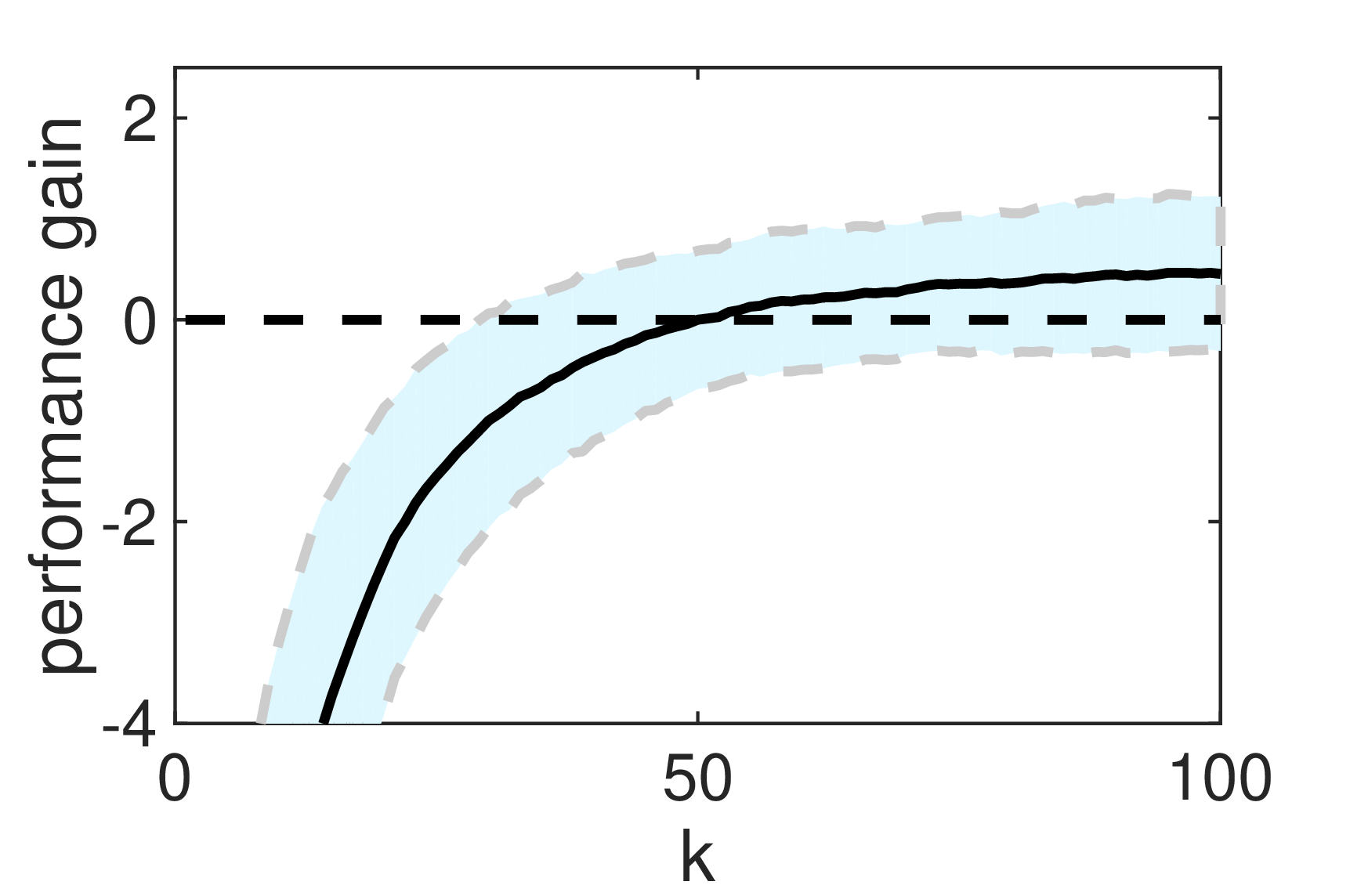}}
	\subfigure[rank-5]{\includegraphics[clip=true,trim=12 5 15 0,width=0.245\columnwidth]{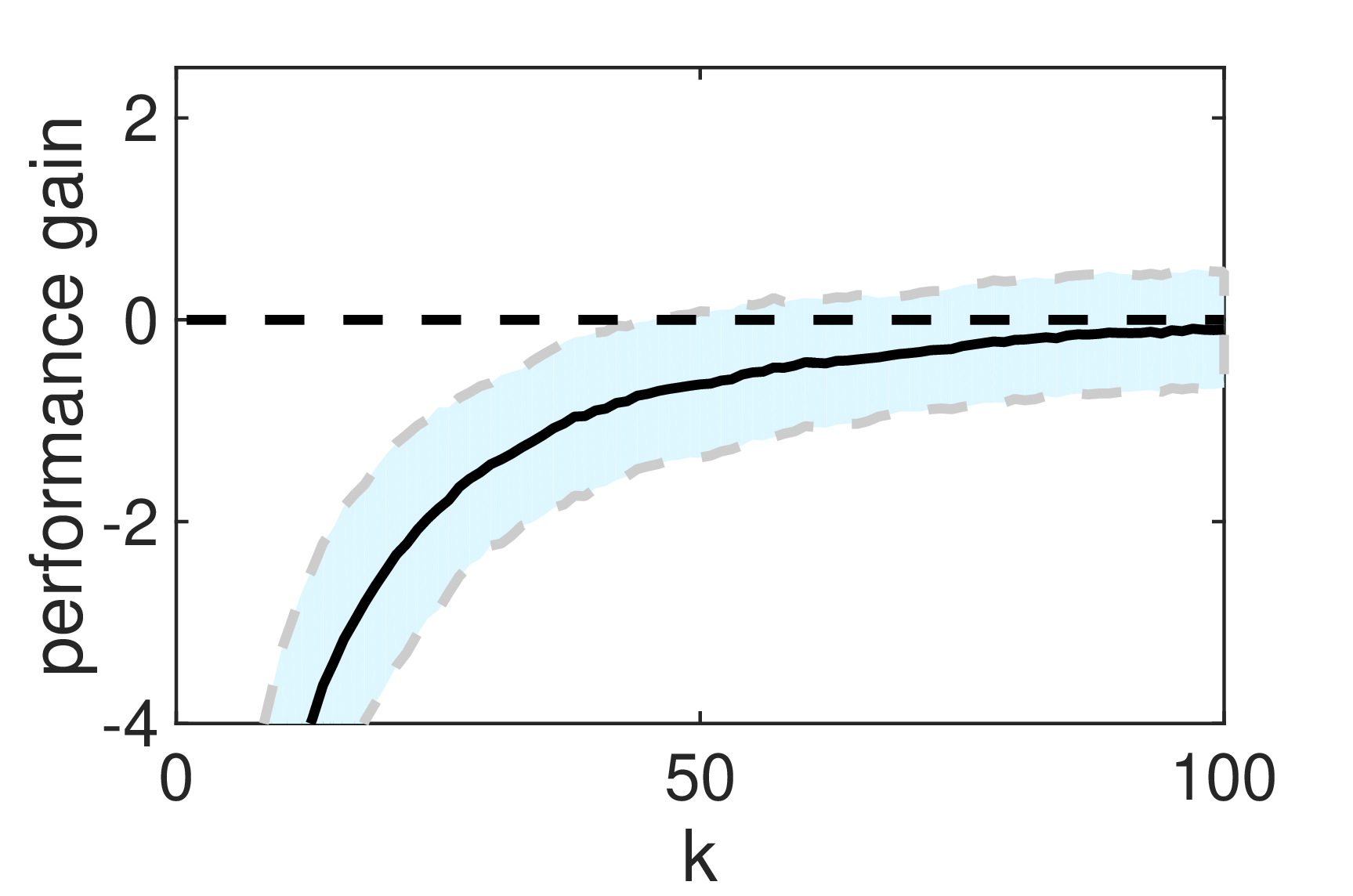}}
	\subfigure[rank-10]{\includegraphics[clip=true,trim=12 5 15 0,width=0.245\columnwidth]{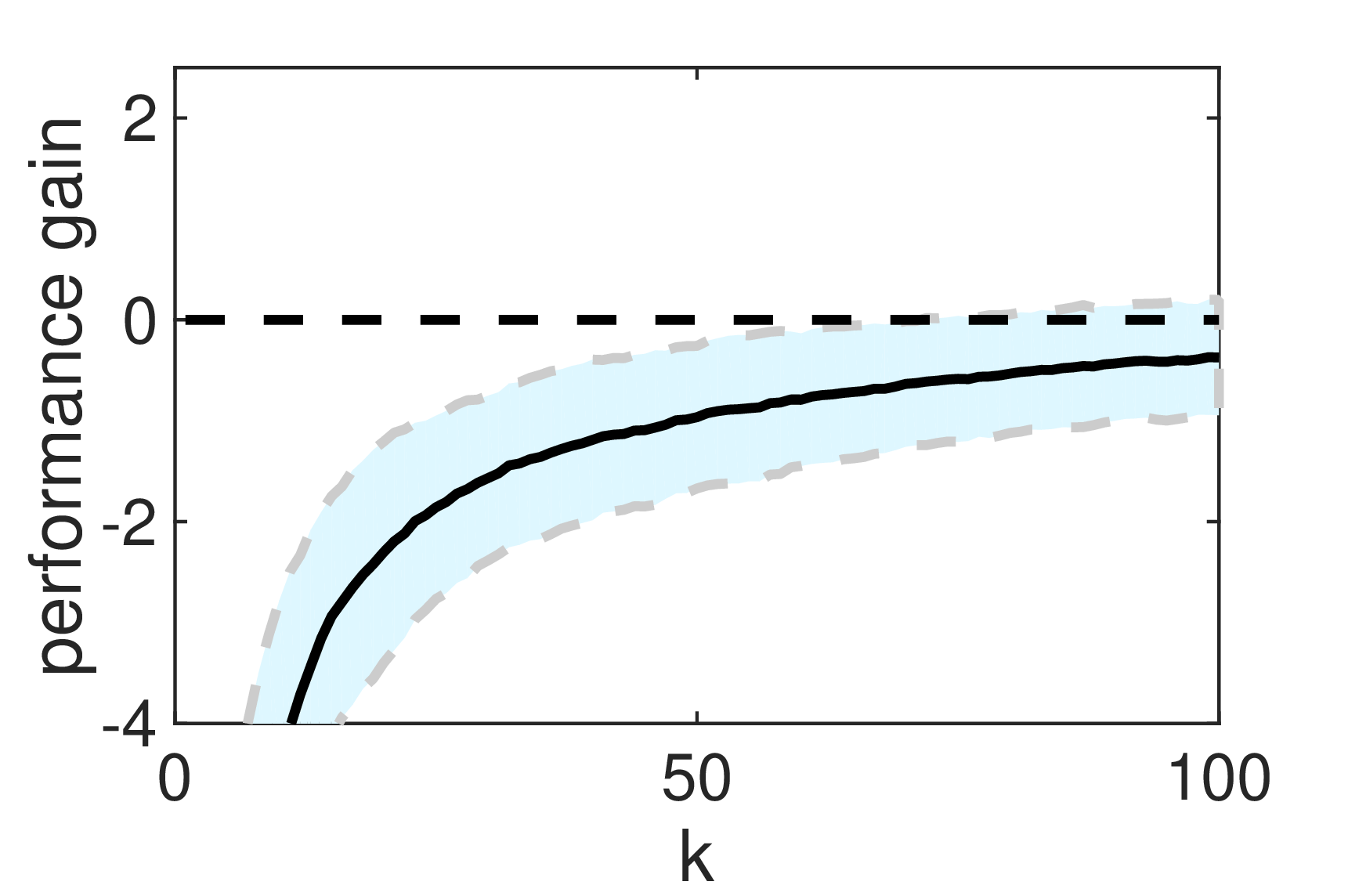}}
	\subfigure[rank-20]{\includegraphics[clip=true,trim=12 5 15 0,width=0.245\columnwidth]{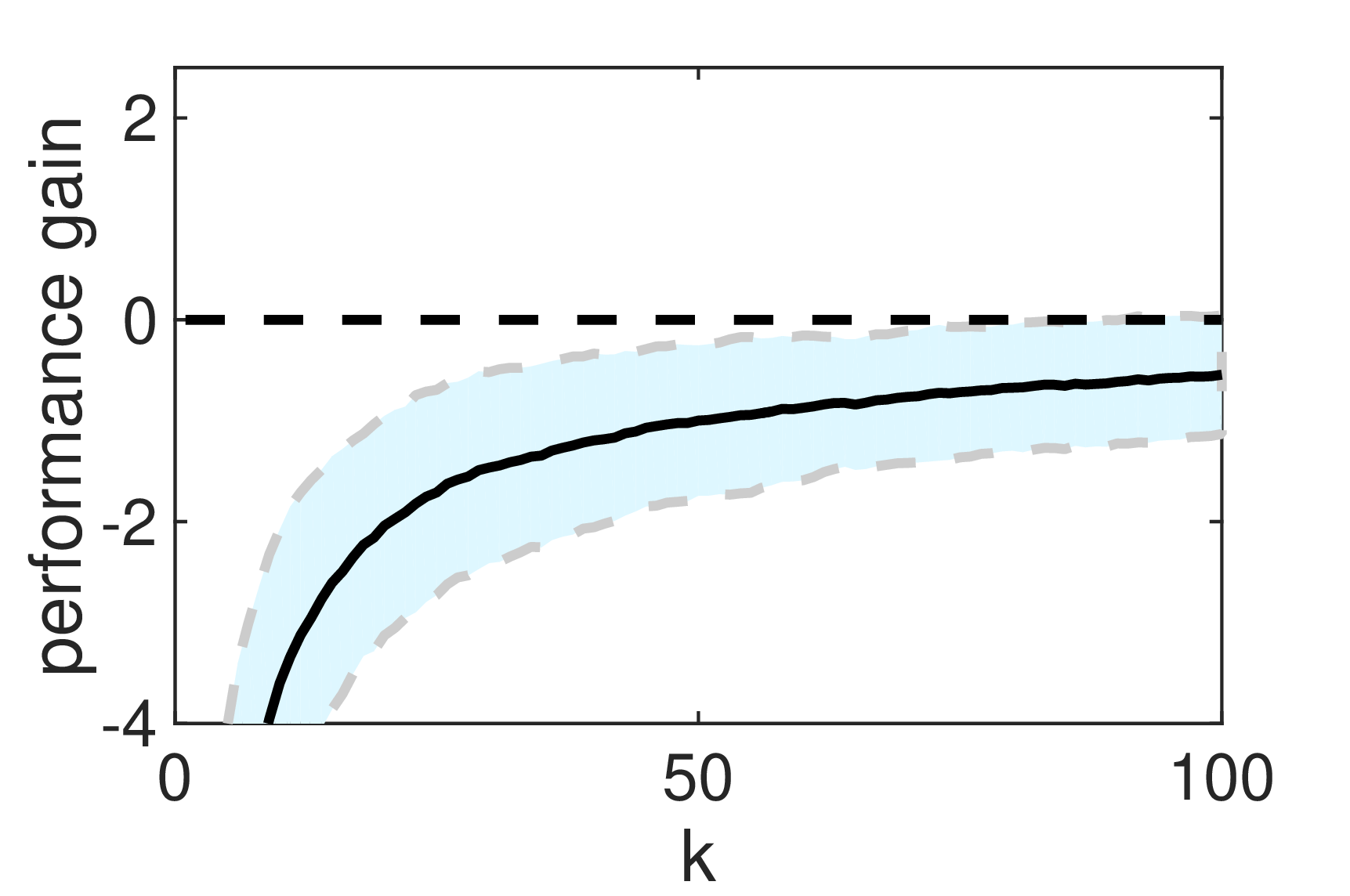}}
	\caption{Average performance gain of inv-DAKR+ as a function of $k$ in a multiple-to-multiple matching scenario after the addition of a large number of extra samples.} 
	\label{fig:increment_inv_new_large}
	\vspace{-1mm}
\end{figure*}

\begin{figure*}[htbp]
	\vspace{-2pt}
	\centering
	\small
	\subfigure[rank-1]{\includegraphics[clip=true,trim=12 5 15 0,width=0.245\columnwidth]{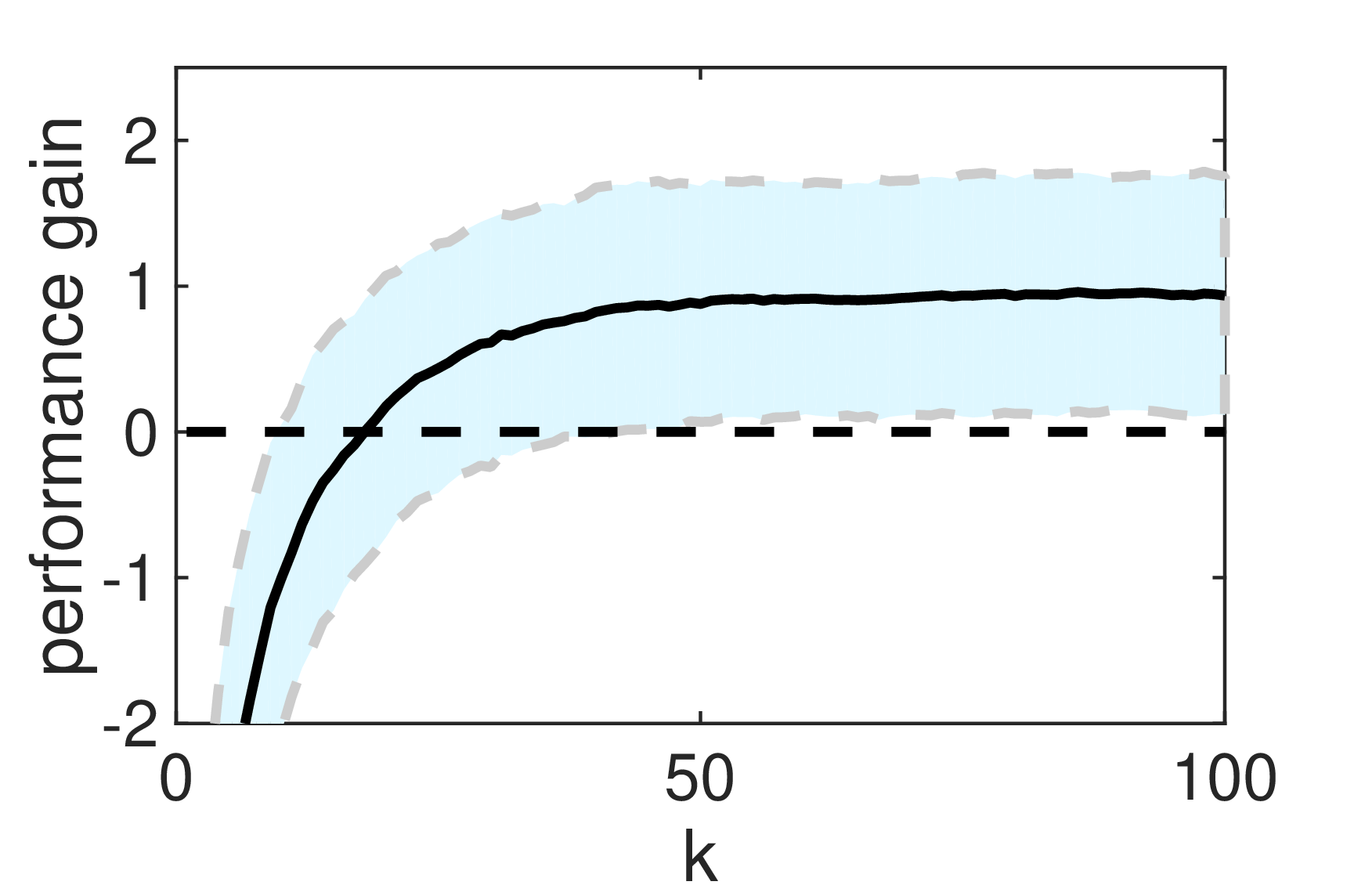}}
	\subfigure[rank-5]{\includegraphics[clip=true,trim=12 5 15 0,width=0.245\columnwidth]{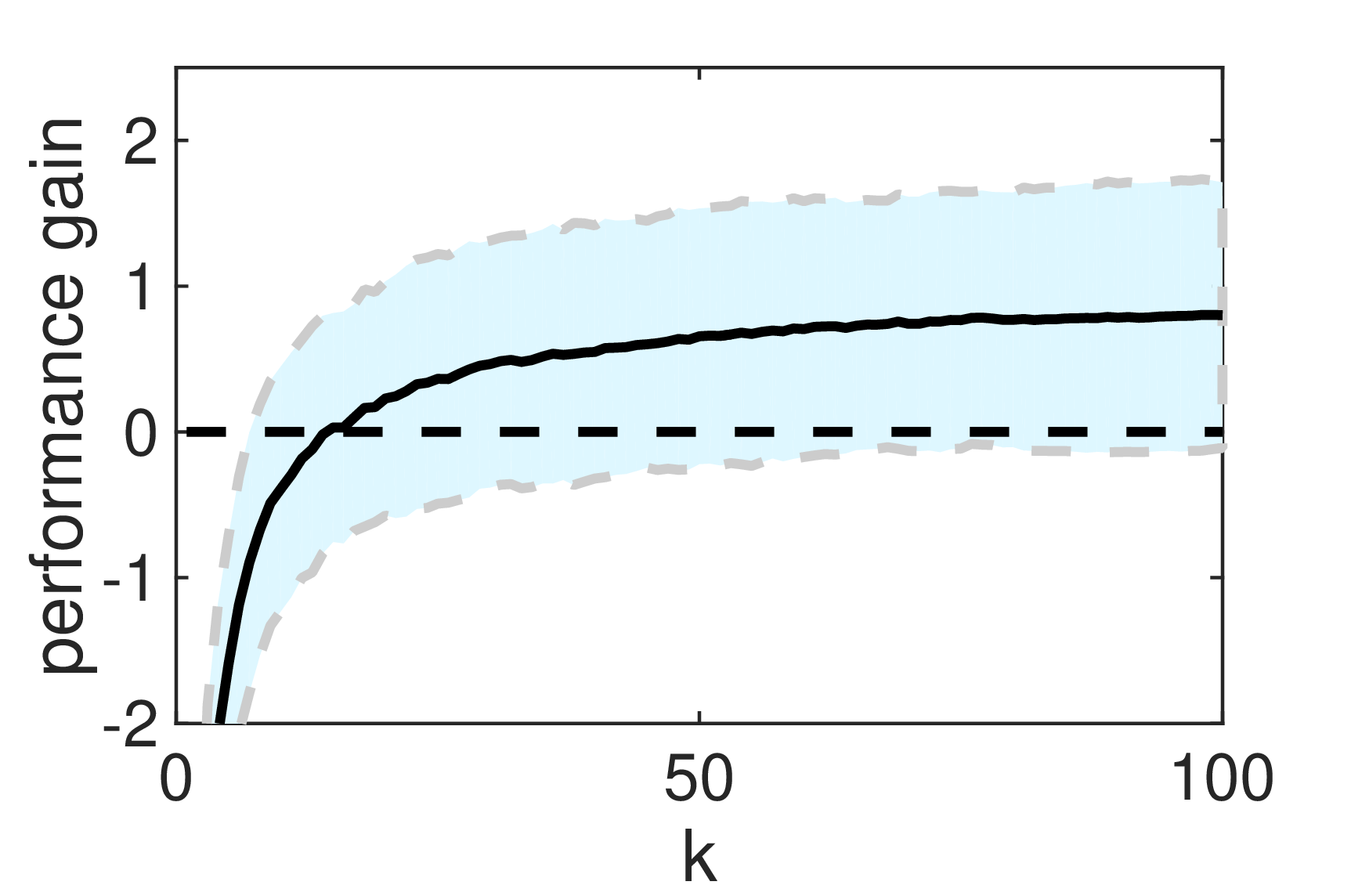}}
	\subfigure[rank-10]{\includegraphics[clip=true,trim=12 5 15 0,width=0.245\columnwidth]{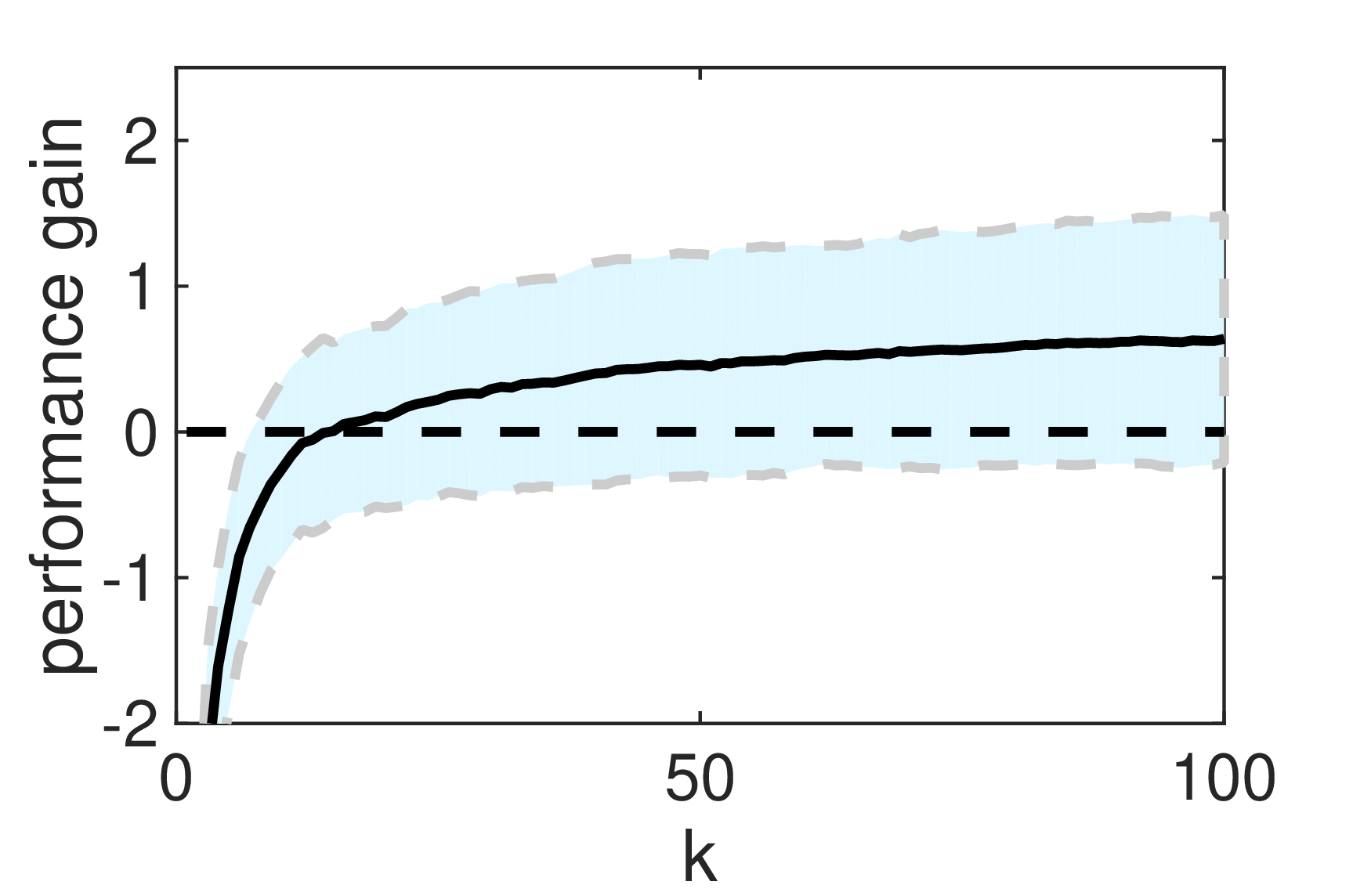}}
	\subfigure[rank-20]{\includegraphics[clip=true,trim=12 5 15 0,width=0.245\columnwidth]{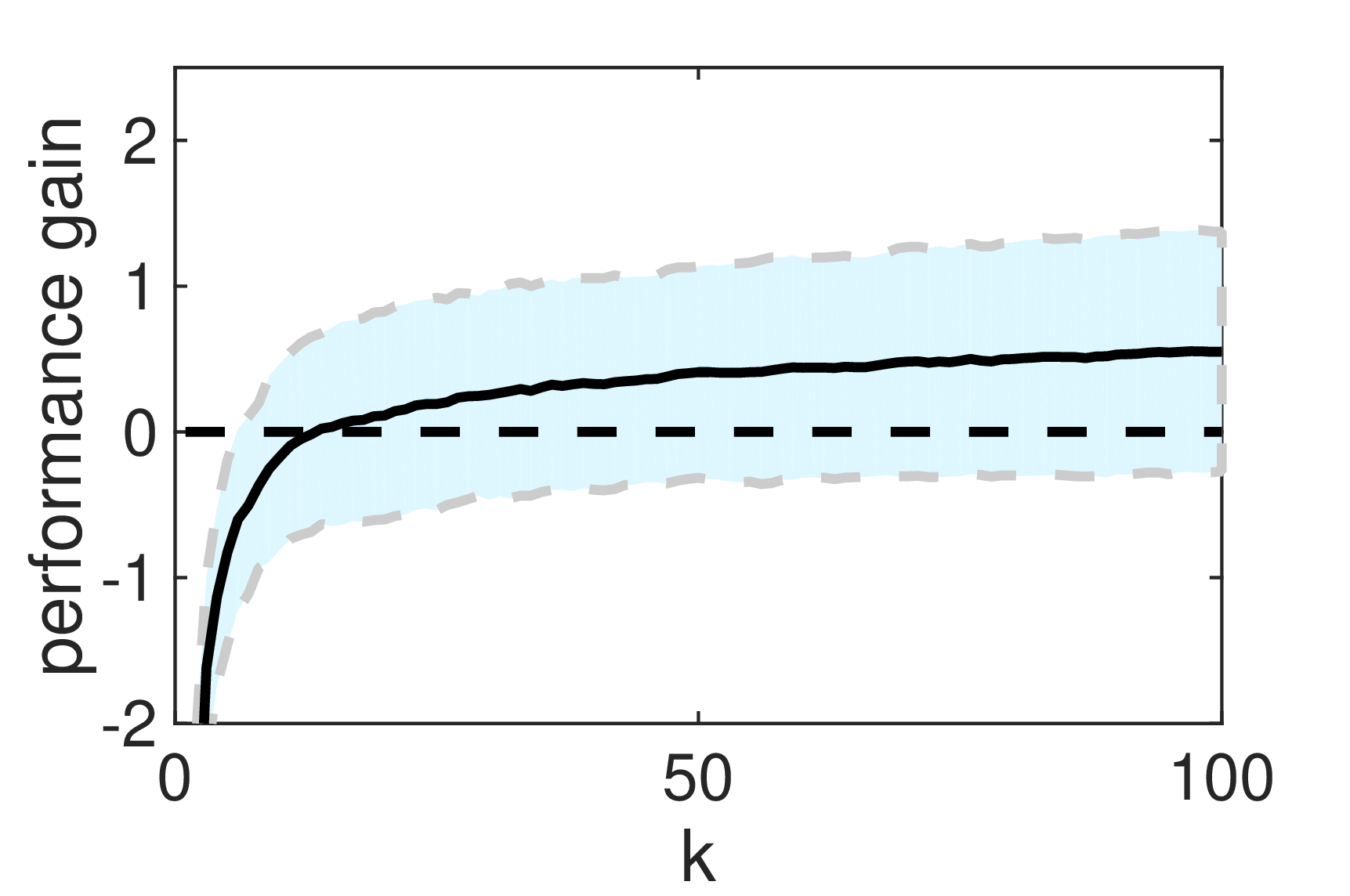}}
	\caption{Average performance gain of bi-DAKR+ as a function of $k$ in a multiple-to-multiple matching scenario after the addition of a large number of extra samples.} 
	\label{fig:increment_bi_new_large}
	\vspace{-1mm}
\end{figure*}

Therefore, to further verify the effect of adding extra \textit{dummy} samples in the multiple-shot setting, we exchange the probe set and the gallery set and repeat experiments on Market-1501 and Mars.
Specifically, in Market-1501, 13,115 images of 750 identities compose the probe set, and 3,368 images of the same 750 identities compose the gallery set. In Mars, the probe set consists of 6,082 images sampled from 630 identities, and gallery set consists of 1,980 images sampled from the same 630 identities.
Therefore, the average number of ground truths per identity is 4.5 in Market-1501 and 3.1 in Mars. The experimental results are presented in Tables~\ref{tab:Market-1501 with ResNet-50-IDE for large extra samples}, \ref{tab:Market-1501 with Caffe for large extra samples}, \ref{tab:Market-1501 with LOMO for large extra samples} and \ref{tab:Mars-IDE for large extra samples}. In this case, 
we can read that the performance of the inv-DAKR+ and bi-DAKR+ even degenerate a little bit after adding a large number of extra ``probe samples''. 
Although the samples in the ``probe set'' become three times of the ``gallery set'', the added extra probe samples did not substantially improve the performance in the datasets under the multiple-shot setting.
In addition, we note that after adding a large number of extra samples, the number of average ground-truths has dramatically changed (\textit{e.g.}, now it is 22.0 in Market-1501 and 12.8 in Mars) and thus a larger value of $k$ is needed to make the performance of our proposals stable. This 
is illustrated again in Figs.~\ref{fig:increment_inv_set_3_large}, \ref{fig:increment_bi_set_3_large}, \ref{fig:increment_inv_new_large} and \ref{fig:increment_bi_new_large}.

\begin{table}[htbp]
	\caption{{ Comparison on Market-1501 with ResNet-50-IDE features in terms of different metrics after the addition of a large number of extra samples.}}
	\label{tab:Market-1501 with ResNet-50-IDE for large extra samples}
	\vspace{4pt}
	\small
	\centering
	\begin{tabular}{l|l|l|l|l|l}
		\hline
		\textbf{Metric} &\textbf{Methods} &\textbf{r=1} &\textbf{r=5} &\textbf{r=10} &\textbf{mAP}\\\hline
		\hline
		\multirow{5}{*}{Euc}  &$k$-NN &62.01 &83.05 &89.16 &47.71\\
		&inv-DAKR &62.42 &82.63 &88.75 &47.96\\
		&bi-DAKR &62.66 &83.19 &\textbf{89.42} &\textbf{48.48}\\
		&inv-DAKR+ &62.48 &82.88 &88.81 &48.13\\
		&bi-DAKR+ &\textbf{62.74} &\textbf{83.26} &89.21 &48.44\\
		\hline
		\multirow{5}{*}{XQDA}  &$k$-NN &70.64 &88.60 &93.21 &56.43\\
		&inv-DAKR &70.52 &88.43 &93.02 &56.66\\
		&bi-DAKR &\textbf{71.24} &\textbf{89.13} &\textbf{93.52} &\textbf{57.51}\\
		&inv-DAKR+ &70.55 &88.39 &92.91 &56.72\\
		&bi-DAKR+ &71.00 &89.03 &93.37 &57.38\\
		\hline
		\multirow{5}{*}{KISSME}  &$k$-NN &72.45 &89.11 &93.67 &57.06\\
		&inv-DAKR &72.66 &89.23 &93.53 &57.48\\
		&bi-DAKR &\textbf{73.35} &\textbf{89.83} &\textbf{94.06} &\textbf{58.44}\\
		&inv-DAKR+ &72.60 &89.24 &93.59 &57.65\\
		&bi-DAKR+ &73.22 &89.69 &93.97 &55.96\\
		\hline
		\multirow{5}{*}{Mahal}  &$k$-NN &71.99 &88.88 &93.38 &55.94\\
		&inv-DAKR &71.82 &88.75 &93.25 &55.97\\
		&bi-DAKR &\textbf{72.79} &\textbf{89.39} &93.76 &\textbf{57.33}\\
		&inv-DAKR+ &71.89 &88.83 &93.22 &56.58\\
		&bi-DAKR+ &72.64 &89.25 &\textbf{93.77} &57.20\\
		\hline
		
	\end{tabular}
	\vspace{-5pt}
\end{table}

\begin{table}[htbp]
	\caption{{ Comparison on Market-1501 with Caffe features in terms of different metrics after the addition of a large number of extra samples.}}
	\label{tab:Market-1501 with Caffe for large extra samples}
	\vspace{4pt}
	\small
	\centering
	\begin{tabular}{l|l|l|l|l|l}
		\hline
		\textbf{Metric} &\textbf{Methods} &\textbf{r=1} &\textbf{r=5} &\textbf{r=10} &\textbf{mAP}\\\hline
		\hline
		\multirow{5}{*}{Euc}  &$k$-NN &48.33 &73.75 &82.26 &35.28\\
		&inv-DAKR &48.84 &73.72 &82.11 &35.66\\
		&bi-DAKR &49.19 &\textbf{73.95} &82.51 &35.91\\
		&inv-DAKR+ &49.31 &73.89 &82.13 &35.93\\
		&bi-DAKR+ &\textbf{49.33} &73.44 &\textbf{82.65} &\textbf{36.02}\\
		\hline
		\multirow{5}{*}{XQDA}  &$k$-NN &55.52 &79.37 &86.63 &41.89\\
		&inv-DAKR &55.78 &79.35 &86.37 &42.20\\
		&bi-DAKR &\textbf{56.11} &\textbf{80.01} &\textbf{87.00} &42.66\\
		&inv-DAKR+ &55.89 &79.33 &86.60 &42.32\\
		&bi-DAKR+ &56.05 &79.78 &86.88 &\textbf{42.67}\\
		\hline
		\multirow{5}{*}{KISSME}  &$k$-NN &54.94 &78.86 &86.05 &41.06\\
		&inv-DAKR &54.61 &78.33 &85.84 &41.03\\
		&bi-DAKR &55.55 &\textbf{79.38} &\textbf{86.63} &41.81\\
		&inv-DAKR+ &54.94 &78.47 &85.83 &41.37\\
		&bi-DAKR+ &\textbf{55.65} &79.22 &86.43 &\textbf{41.85}\\
		\hline
		\multirow{5}{*}{Mahal}  &$k$-NN &53.63 &78.15 &85.66 &39.58\\
		&inv-DAKR &52.99 &77.31 &84.86 &39.20\\
		&bi-DAKR &\textbf{54.11} &\textbf{78.25} &\textbf{85.76} &40.07\\
		&inv-DAKR+ &53.47 &77.41 &84.96 &39.67\\
		&bi-DAKR+ &54.14 &78.46 &85.69 &\textbf{40.16}\\
		\hline
		
	\end{tabular}
	\vspace{-0pt}
\end{table}

\begin{table}[htbp]
	\caption{{ Comparison on Market-1501 with LOMO features in terms of different metrics after the addition of a large number of extra samples.}}
	\label{tab:Market-1501 with LOMO for large extra samples}
	\vspace{4pt}
	\small
	\centering
	\begin{tabular}{l|l|l|l|l|l}
		\hline
		\textbf{Metric} &\textbf{Methods} &\textbf{r=1} &\textbf{r=5} &\textbf{r=10} &\textbf{mAP}\\\hline
		\hline
		\multirow{5}{*}{Euc}  &$k$-NN &10.24 &21.11 &27.66 &4.03\\
		&inv-DAKR &10.25 &20.92 &27.29 &5.29\\
		&bi-DAKR &10.46 &\textbf{21.54} &27.82 &\textbf{5.38}\\
		&inv-DAKR+ &10.50 &21.06 &27.30 &5.37\\
		&bi-DAKR+ &\textbf{10.51} &\textbf{21.54} &\textbf{27.84} &5.37\\
		\hline
		\multirow{5}{*}{XQDA}  &$k$-NN &25.62 &51.04 &63.51 &17.80\\
		&inv-DAKR &\textbf{26.99} &51.48 &63.53 &18.87\\
		&bi-DAKR &26.70 &\textbf{51.76} &\textbf{64.32} &18.75\\
		&inv-DAKR+ &26.85 &51.18 &63.18 &\textbf{18.92}\\
		&bi-DAKR+ &26.62 &51.74 &64.17 &18.83\\
		\hline
		\multirow{5}{*}{KISSME}  &$k$-NN &35.55 &61.57 &72.41 &23.18\\
		&inv-DAKR &36.09 &61.03 &71.19 &25.04\\
		&bi-DAKR &\textbf{37.38} &\textbf{62.79} &\textbf{73.57} &\textbf{25.50}\\
		&inv-DAKR+ &36.33 &60.87 &71.04 &25.28\\
		&bi-DAKR+ &37.35 &62.82 &73.26 &25.49\\
		\hline
		\multirow{5}{*}{Mahal}  &$k$-NN &29.28 &52.16 &63.19 &17.06\\
		&inv-DAKR &30.45 &52.09 &62.24 &19.79\\
		&bi-DAKR &31.55 &54.70 &\textbf{65.98} &19.95\\
		&inv-DAKR+ &30.68 &52.22 &62.33 &20.02\\
		&bi-DAKR+ &\textbf{31.66} &\textbf{54.82} &65.73 &\textbf{20.04}\\
		\hline
	\end{tabular}
	\vspace{-5pt}
\end{table}

\begin{table}[htbp]
	\caption{{ Comparison on Mars with IDE features and in terms of different metrics for after the addition of a large number of extra samples.}}
	\label{tab:Mars-IDE for large extra samples}
	\vspace{4pt}
	\small
	\centering
	\begin{tabular}{l|l|l|l|l|l}
		\hline
		\textbf{Metric} &\textbf{Methods} &\textbf{r=1} &\textbf{r=5} &\textbf{r=20} &\textbf{mAP}\\\hline
		\hline
		\multirow{5}{*}{Euc}  &$k$-NN &55.10 &76.83 &90.27 &43.70\\
		&inv-DAKR &55.80 &77.66 &89.87 &44.99\\
		&bi-DAKR &\textbf{56.22} &\textbf{78.08} &\textbf{90.46} &\textbf{45.37}\\
		&inv-DAKR+ &54.42 &76.13 &88.54 &43.69\\
		&bi-DAKR+ &55.28 &77.26 &89.92 &44.24\\
		\hline
		\multirow{5}{*}{XQDA}  &$k$-NN &59.95 &81.54 &92.04 &49.59\\
		&inv-DAKR &61.90 &82.70 &92.67 &51.34\\
		&bi-DAKR &\textbf{62.00} &\textbf{82.93} &\textbf{93.06} &\textbf{51.70}\\
		&inv-DAKR+ &60.42 &81.21 &91.22 &49.73\\
		&bi-DAKR+ &61.23 &81.85 &92.55 &50.48\\
		\hline
		\multirow{5}{*}{KISSME}  &$k$-NN &57.99 &79.05 &90.79 &46.55\\
		&inv-DAKR &\textbf{61.34} &81.70 &92.14 &50.41\\
		&bi-DAKR &61.31 &\textbf{82.46} &\textbf{92.85} &\textbf{50.49}\\
		&inv-DAKR+ &59.83 &80.07 &90.79 &48.74\\
		&bi-DAKR+ &60.46 &81.67 &92.42 &49.38\\
		\hline
		\multirow{5}{*}{Mahal}  &$k$-NN &56.15 &77.38 &89.59 &44.16\\
		&inv-DAKR &59.88 &80.43 &91.61 &48.85\\
		&bi-DAKR &\textbf{60.16} &\textbf{81.11} &\textbf{92.27} &\textbf{49.03}\\
		&inv-DAKR+ &58.07 &78.92 &90.02 &47.10\\
		&bi-DAKR+ &58.88 &80.11 &91.89 &47.47\\
		\hline		
	\end{tabular}
	\vspace{-5pt}
\end{table}

\subsection{Evaluation of Performance Sensitivity to the Parameter $k$ and Discussion}
\label{sec:sensitivity-to-k}

In the proposed inv-DAKR and bi-DAKR, the smooth kernel function adopts a local-density adaptive parameter $\sigma_j$, which is based on the distance from $\x_j$ to its $k$-th nearest neighbor $\x_j^{(k)}$.
Thus, it is interesting to evaluate the sensitivity of the performance of inv-DAKR and bi-DAKR to the parameter $k$.
To this end, we compute the average performance gains of inv-DAKR and bi-DAKR over the baseline algorithm ($k$-NN) in terms of accuracies at rank-1, rank-5 rank-10 and rank-20 in three different scenarios, namely, a) \emph{perfect single-shot matching}, b) \emph{imperfect single-shot matching} and c) \emph{multiple-shot matching}, and show each of the results as a function of the parameter $k$ in Figs.~\ref{fig:increment_inv_one_2} to \ref{fig:increment_bi_new_large}.

From all results in Figs.~\ref{fig:increment_inv_one_2} to \ref{fig:increment_bi_new_large}, we observe that while the proposed inv-DAKR and bi-DAKR both yield promising improvements, bi-DAKR yields consistent improvements on all datasets.

When extra probe samples are available and can 
\textit{discriminately} improve the local distribution of the gallery samples, the performance of our proposals (inv-DAKR+ and bi-DAKR+) can be further enhanced.
For example, in the \emph{perfect single-shot matching} scenario, each identity has only one sample in the original gallery set, and its local distribution cannot be accurately described. However, when extra probe samples are added, there are now two samples for each identity, thus forming good local distributions. Therefore, 
our proposals show notably improved performance without requiring any additional supervision information.
This is confirmed 
by the experiments presented in Section~\ref{sec:experiments-perfect-single-shot}.
%
%
By contrast, when extra probe samples are considered in the \emph{imperfect single-shot matching} scenario, the corresponding improvement is negligible.
The reason is that the extra probe samples 
cannot \textit{dramatically and discriminately} change the local distribution. 
Especially, the sample for each identity is surrounded by strongly related background samples, and these extra probe samples are also dominated by these background samples so that local distributions can hardly be improved. This is verified in Section~\ref{sec:experiments-imperfect-single-shot}.
%
%
%
%
%
%
In the \emph{multiple-shot matching} scenario, we consider two situations to demonstrate the effects of adding extra probe samples. One is that the probe set is much smaller than the gallery set. 
The other is that the probe set is much larger than the gallery set. 
From the experimental results of adding extra probe samples in Section~\ref{sec:experiments-multiple-shot}, we observe merely minor improvements or even minor degenerations. Through above the experiments on using the extra probe samples, we suggest that: if extra probe samples are not too many compared to the gallery set, we should add the extra samples and perform inv-DAKR+ or bi-DAKR+. When the extra probe samples can \textit{discriminately} find their gallery samples, the added extra probe samples will lead to notably improvement to the reranking result.

By comparing the optimal value of $k$ with the average number of ground-truth samples for each identity, 
we can make some interesting observations. 
In the \emph{perfect single-shot matching} situation, it is clearly shown in Figs.~\ref{fig:increment_inv_one_2} to \ref{fig:increment_bi_one_2} that bi-DAKR+ will exhibit the best performance when $k$ is approximately 2. This is because when extra probe samples are available in this situation, they can help to generate a more balanced distribution, in which there are two samples for each identity on the manifold, and the performance of 
bi-DAKR+ is simply related to this balance.
%
The stability of inv-DAKR+ is weaker; thus, the optimal value of $k$ needs to be slightly larger than 2. The same phenomena can also be found in the \emph{multiple-shot matching} situation. The results show that the proper value of the parameter $k$ is also related to the average number of ground-truth matches in the datasets. For example, there are, on average, 26.3 ground-truth matches in Market-1501; thus, the performance of bi-DAKR becomes stable when $k \ge 27$. Compared to bi-DAKR, which performs reidentification from two directions, inv-DAKR needs a larger value of $k$ to achieve relatively stable performance. 
Nevertheless, the observations are slightly different in the \emph{imperfect single-shot matching} situation. Here, there are 7.2  ground-truth samples for each identity on average, 
but the number of incorrect distracters 
is too large compared to the number of real samples of identities (775:125). As a result, the average number of ground-truth samples per identity cannot be suitably estimated. In this case, 
we set a larger value of $k$ to obtain more local information.
In summary, we suggest that if the average number of ground-truth matches can be suitably estimated as prior information, as in the case of \emph{perfect single-shot matching} and \emph{multiple-shot matching}, then $k$ should be set to a value close to this number for bi-DAKR and slightly larger for inv-DAKR; otherwise, $k$ should be empirically set to $30$ or larger.

\subsection{Comparison of Time Costs}
\label{sec:time cost}

\begin{table}[htbp]
\vspace{-10pt}
  \caption{Comparison of Computational Time Costs.}
  \label{tab:complexity with dataset}
  \scriptsize
  \centering
  \begin{tabular}{c|c|c|c}
    \hline
    \textbf{Dataset} &\textbf{Size} & \textbf{Methods} & \textbf{Time} \\
    \hline
    \hline
    \multirow{5}{*}{CUHK03} &\multirow{5}{*}{200} &inv-DAKR &0.0356$s$ \\
     & &bi-DAKR  &0.0353$s$ \\
     & &inv-DAKR+ &0.0716$s$ \\
     & &bi-DAKR+ &0.0704$s$ \\
     & &SCA\cite{Bai:TIP2016}   &0.1841$s$\\
     & &$k$-INN\cite{Korn:ASR2000}   &0.1070$s$\\
     & &$k$-RNN   &0.1130$s$\\
     & &Zhong's\cite{Zhong:CVPR2017}  &0.2292$s$ \\
     & &Yu's\cite{Yu:BMVC2017}  &1.2273$s$ \\
     & &MRank$-L_{n}$\cite{Chen:ICIP13} &0.24$s$ \\
    \hline
    \multirow{5}{*}{PRID450s} &\multirow{5}{*}{450} &inv-DAKR &0.0824$s$ \\
     & &bi-DAKR  &0.0804$s$ \\
     & &inv-DAKR+ &0.1683$s$ \\
     & &bi-DAKR+ &0.1648$s$ \\
     & &SCA\cite{Bai:TIP2016}   &0.3652$s$\\
     & &$k$-INN\cite{Korn:ASR2000}   &0.3982$s$\\
     & &$k$-RNN   &0.4245$s$\\
     & &Zhong's\cite{Zhong:CVPR2017}  &0.5682$s$ \\
     & &Yu's\cite{Yu:BMVC2017}  &2.5876$s$ \\
     & &MRank$-L_{n}$\cite{Chen:ICIP13} &1.01$s$ \\
    \hline
    \multirow{5}{*}{VIPeR} &\multirow{5}{*}{632} &inv-DAKR &0.1563$s$ \\
     & &bi-DAKR  &0.1582$s$ \\
     & &inv-DAKR+ &0.2786$s$ \\
     & &bi-DAKR+ &0.2809$s$ \\
     & &SCA\cite{Bai:TIP2016}   &0.5622$s$\\
     & &$k$-INN\cite{Korn:ASR2000}   &0.8642$s$\\
     & &$k$-RNN   &0.8984$s$\\
     & &Zhong's\cite{Zhong:CVPR2017}  &0.8657$s$ \\
     & &Yu's\cite{Yu:BMVC2017}  &4.1541$s$ \\
     & &MRank$-L_{n}$\cite{Chen:ICIP13} &3.05$s$ \\
    \hline
    \multirow{5}{*}{GRID} &\multirow{5}{*}{1025} &inv-DAKR &0.1302$s$ \\
     & &bi-DAKR  &0.1282$s$ \\
     & &inv-DAKR+ &0.3144$s$ \\
     & &bi-DAKR+ &0.3169$s$ \\
     & &SCA\cite{Bai:TIP2016}   &0.6196$s$\\
     & &$k$-INN\cite{Korn:ASR2000}   &2.6937$s$\\
     & &$k$-RNN   &2.7288$s$\\
     & &Zhong's\cite{Zhong:CVPR2017}  &1.2032$s$ \\
     & &Yu's\cite{Yu:BMVC2017}  &7.1974$s$ \\
     & &MRank$-L_{n}$\cite{Chen:ICIP13} &15.38$s$ \\
    \hline
    \multirow{5}{*}{Mars} &\multirow{5}{*}{12180} &inv-DAKR &3.2643$s$ \\ 
    &  &bi-DAKR  &3.3172$s$ \\  
    & &inv-DAKR+ &3.5496$s$ \\
    & &bi-DAKR+ &4.1559$s$ \\
    &  &SCA\cite{Bai:TIP2016}   &31.01$s$\\
    &  &$k$-INN\cite{Korn:ASR2000}   &9966.2$s$\\
    &  &$k$-RNN   &9973.7$s$\\
    &  &Zhong's\cite{Zhong:CVPR2017}  &33.4623$s$ \\
    &  &Yu's\cite{Yu:BMVC2017}  &2923.2$s$ \\
    \hline
    \multirow{5}{*}{Market-1501} &\multirow{5}{*}{19732} &inv-DAKR  &4.6836$s$ \\
    &  &bi-DAKR  &5.6088$s$ \\ 
    &  &inv-DAKR+  &5.8147$s$ \\
    &  &bi-DAKR+  &7.5005$s$ \\
    &  &SCA\cite{Bai:TIP2016}   &66.2138$s$\\
    &  &$k$-INN\cite{Korn:ASR2000}   &48826.4$s$\\
    &  &$k$-RNN   &48610.1$s$\\
    &  &Zhong's\cite{Zhong:CVPR2017}  &88.3883$s$ \\
    &  &Yu's\cite{Yu:BMVC2017}  &8785.0$s$ \\
    \hline
  \end{tabular}
\end{table}

For a fair comparison, we list the time costs of $k$-INN~\cite{Korn:ASR2000}, $k$-RNN, Zhong et al.~\cite{Zhong:CVPR2017}, MRank-$L_{n}$~\cite{Chen:ICIP13} and our proposals in Table~\ref{tab:complexity with dataset}. For CHUK03, we report the time costs for the labeled dataset. The sizes listed in Table~\ref{tab:complexity with dataset} are the numbers of samples in the test datasets. For fairness, we set Zhong's method~\cite{Zhong:CVPR2017} to its default parameters, \textit{i.e.}, $k_1=20$, $k_2=6$, and $\lambda=0.95$, and uniformly set $k=20$ for $k$-INN~\cite{Korn:ASR2000}, $k$-RNN, SCA~\cite{Bai:TIP2016}, Yu's~\cite{Yu:BMVC2017}, MRank-$L_{n}$~\cite{Chen:ICIP13} and our proposals.

The results show that the time costs of our proposals are much lower than those of $k$-INN~\cite{Korn:ASR2000}, $k$-RNN, Zhong et al.~\cite{Zhong:CVPR2017}, and MRank-$L_{n}$~\cite{Chen:ICIP13}, especially when the dataset is large and with the addition of extra probe samples.
Specifically, as shown in Tables~\ref{tab:complexity with dataset}, \ref{tab:GRID} and \ref{tab:PRID450s-VIPeR-CUHK03}, the operation of matrix inversion in MRank-$L_{n}$\cite{Chen:ICIP13} limits both its efficiency and effectiveness such that it is difficult to deploy it on a large dataset.
In addition, Table~\ref{tab:complexity with dataset} also reveals that the time cost of $k$-INN~\cite{Korn:ASR2000} and $k$-RNN increases rapidly %
when the size of the testing set increases. 
The time costs are even larger than that of Zhong et al. \cite{Zhong:CVPR2017} when the size of the testing set reaches 1,025 samples in GRID and becomes excessive on both Market-1501 and Mars. In addition, SCA~\cite{Bai:TIP2016} is faster than $k$-INN~\cite{Korn:ASR2000} and $k$-RNN because of the assistance of the inverted index, and the efficiency of Zhong et al. \cite{Zhong:CVPR2017} is also beneficial. However, the method of Zhong et al. \cite{Zhong:CVPR2017} requires finding and expanding the $k$-RNN sets, expanding the local query sequence, and computing Jaccard distance, whereas in our inv-DAKR and bi-DAKR, only a density-adaptive kernel function needs to be evaluated. Moreover, based on Zhong et al.'s method \cite{Zhong:CVPR2017}, Yu et al. \cite{Yu:BMVC2017} used three additional hyperparameters for feature division, iterative encoding and fuzzy fusion. Although Yu et al. \cite{Yu:BMVC2017} produces promising 
reranking results, it is difficult to be used in reality because of too many hyperparameters.

Though slightly outperformed by the reranking methods in \cite{Bai:TIP2016}, \cite{Zhong:CVPR2017} and Yu et al. \cite{Yu:BMVC2017} with respect to the rank-1 accuracy and mAP, our inv-DAKR and bi-DAKR methods are much simpler and faster, and achieve improved results at rank-5, rank-10 and rank-20.

\section{Conclusion}
\label{sec:conclusion}

We have addressed the reranking problem for person ReID. Specifically, we have proposed two density-adaptive kernel based reranking approaches, named inv-DAKR and bi-DAKR, in which density-adaptive parameters are adopted to capture the local density information in the gallery set. 
The inv-DAKR and bi-DAKR can be viewed as a smoothed version of the $k$-INN and $k$-RNN reranking methods, respectively.
Moreover, 
we have extended inv-DAKR and bi-DAKR to the setting of reranking with extra probe samples and have demonstrated when and why the incorporation of extra samples can 
remarkably improve 
the reranking performance. 
Extensive experiments on six benchmark datasets have validated the efficiency and effectiveness of our proposals. Owing to the simplicity of the implementation and the low computational cost, we hope that our proposals can be widely applied in real-world person ReID systems.

\section*{Acknowledgments}
R. Guo, J. Lin, and Y. Li are supported by the National Natural Science Foundation of China
under Grant No. 61771066. 
C.-G. Li and J. Guo are supported by the National Natural Science Foundation of China
under Grant No. 61876022.

\section*{References}
\small
\bibliography{IEEEabrv,personReID,cgli} 
\end{document}